\documentclass[10pt,a4paper]{article}

\usepackage[table]{xcolor}
\usepackage{enumitem}
\usepackage{tabularx}
\usepackage{ragged2e}

\usepackage{arxiv}
\usepackage{tikz}
\usetikzlibrary{arrows.meta}
\usetikzlibrary{tikzmark, positioning, fit, shapes.misc}
\usetikzlibrary{arrows}
\usetikzlibrary{decorations.pathreplacing, calc}

\tikzset{brace/.style={decorate, decoration={brace}},
  brace mirrored/.style={decorate, decoration={brace,mirror}},
}

\usepackage[utf8]{inputenc} 
\usepackage[T1]{fontenc}    
\usepackage{hyperref}       
\usepackage{url}            
\usepackage{booktabs}       
\usepackage{amsfonts}       
\usepackage{nicefrac}       
\usepackage{microtype}      

\usepackage{amssymb}
\usepackage{lineno,hyperref}
\usepackage{amsmath}
\usepackage{bm}
\usepackage{subcaption}
\usepackage{graphicx}
\usepackage{svg}
\usepackage{rotating}
\usepackage{lscape}
\usepackage[title]{appendix}
\usepackage{float}
\usepackage{stmaryrd}
\usepackage{textcomp}

\usepackage{amsthm}
\usepackage{listings}
\usepackage{xcolor}
\usepackage{siunitx}

\usepackage{pifont}
\usepackage{lineno}

\usepackage{import}
\usepackage{physics}
\usepackage{comment}
\usetikzlibrary{positioning}

    \DeclareFontFamily{OT1}{pzc}{}
\DeclareFontShape{OT1}{pzc}{m}{it}{<-> s * [1.10] pzcmi7t}{}
\DeclareMathAlphabet{\mathpzc}{OT1}{pzc}{m}{it}
\usepackage[numbers]{natbib}
\usepackage{algpseudocode,algorithm,algorithmicx}

\makeatletter
\def\ALG@special@indent{%
    \ifdim\ALG@thistlm=0pt\relax
        \hskip-\leftmargin
    \else
        \hskip\ALG@thistlm
    \fi
}
\newcommand{\Begin}[1]{\item[]\noindent\ALG@special@indent \textbf{Begin:}\ #1}
\newcommand{\Establish}[1]{\item[]\noindent\ALG@special@indent \textbf{Establish:}\ #1}
\newcommand{\End}{\item[]\noindent\ALG@special@indent \textbf{End}}
\makeatother
\algblock[Name]{Start}{End}
\algblockdefx[NAME]{START}{END}%
    [2][Unknown]{Start #1(#2)}%
    {Ending}

\usepackage{nccmath}

\title{

Modular machine learning-based elastoplasticity: generalization in the context of limited data 
}

\author{
  Jan Niklas Fuhg$^{\star}$ \\
  Sibley School of Mechanical and Aerospace Engineering \\
  Cornell University, 
   NY 14850, USA \\
  \texttt{jf853@cornell.edu} \\
     \And
  Craig M. Hamel \\
 Sandia National Laboratories\\ Albuquerque, 87185, NM, USA
 \And
   Kyle Johnson \\
 Sandia National Laboratories\\ Albuquerque, 87185, NM, USA
  \And
   Reese Jones \\
 Sandia National Laboratories\\ Livermore, CA 94551, USA \\
   \And
 Nikolaos Bouklas \\
  Sibley School of Mechanical and Aerospace Engineering\\
  Center for Applied Mathematics\\
  Cornell University,
   NY 14850, USA 
  }

\usetikzlibrary{arrows}

\makeatletter
\newcommand*{\compress}{\@minipagetrue}
\makeatother

\newtheorem{corollary}{Corollary}

\begin{document}
\maketitle

\begin{abstract}
The development of highly accurate constitutive models for materials that undergo path-dependent processes continues to be a complex challenge in computational solid mechanics. Challenges arise both in considering the appropriate model assumptions and from the viewpoint of data availability, verification, and validation. 
Recently, data-driven modeling approaches have been proposed that aim to establish stress-evolution laws that avoid user-chosen functional forms by relying on machine learning representations and algorithms.
However, these approaches not only require a significant amount of data but also need data that probes the full stress space with a variety of complex loading paths.  
Furthermore, they rarely enforce all necessary thermodynamic principles as hard constraints.
Hence, they are in particular not suitable for low-data or limited-data regimes, where the first arises from the cost of obtaining the data and the latter from the experimental limitations of obtaining labeled data, which is commonly the case in engineering applications.
In this work, we discuss a hybrid framework that can work on a variable amount of data by relying on the modularity of the elastoplasticity formulation where each component of the model can be chosen to be either a classical phenomenological or a data-driven model depending on the amount of available information and the complexity of the response.
The method is tested on synthetic uniaxial data coming from simulations as well as cyclic experimental data for structural materials.
The discovered material models are found to not only interpolate well but also allow for accurate extrapolation in a thermodynamically consistent manner far outside the domain of the training data. This ability to extrapolate from limited data was the main reason for the early and continued success of phenomenological models and the main shortcoming in machine learning-enabled constitutive modeling approaches.
Training aspects and details of the implementation of these models into Finite Element simulations are discussed and analyzed.
\end{abstract}

\keywords{Physics-informed machine learning \and Solid mechanics \and Nonlinear kinematic hardening \and Non-associative plasticity \and  Data-driven constitutive models}

\section{Introduction}
In recent years machine learning tools have often emerged as preferred alternatives to established approaches, pushing the frontiers in the computational sciences. 
This is due to two main reasons: They allow for direct utilization of available data without the need to establish analytical models, and they have the potential to speed up computations in comparison to traditional numerical techniques.
Hence, in the last few years, machine learning-based schemes have, for example, been:
(i) explored as direct
solvers for forward problems involving partial differential equations\cite{raissi2019physics, fuhg2022interval, fuhg2022mixed},
(ii) deployed to construct active learning algorithms for physical systems \cite{gubaev2018machine,podryabinkin2019accelerating}, and
(iii) used to construct surrogates of parameterized partial differential solutions that can be utilized for topology optimization \cite{hamel2019machine, sosnovik2019neural}, uncertainty quantification \cite{wang2019deep,liu2020data} or inverse problems \cite{jagtap2020conservative,hamel2022calibrating}. 
Data-driven techniques have also been applied in modeling material constitutive behaviors \cite{wu1990representation,  ghaboussi1991knowledge, hashash2004numerical,bock2019review, rocha2021fly} in the context of solid mechanics.
Data-driven modeling of elastic material behavior was initially dominated by "black-box" models which directly map strain to stress 
\cite{javadi2003neural, lefik2003artificial, le2015computational, xu2021learning, chung2021neural, fuhg2022local}.
Lately, these models have been extended by including physical principles and mechanistic assumptions into the modeling process by, for example,  
employing the representation theorem of tensor functions
\cite{jones2018machine,frankel2020tensor,fuhg2022physics, FUHG2022105022} or by enforcing polyconvexity of the corresponding free energy \cite{klein2022polyconvex, FUHG2022105022}.
Furthermore, hybrid modeling frameworks have been explored where a data-driven model locally corrects the output of a traditional phenomenological approach \cite{gonzalez2019learning, fuhg2021model,frankel2022machine}.

Similar to the developments in elastic modeling, data-driven techniques have also been proposed to model path-dependent material behavior. 
These models commonly employ an incremental approach where strains and stresses from previous time steps are employed to predict the stress of the current time step.
In principle, as remarked in \cite{vlassis2021sobolev}, this resembles hypoelastic constitutive modeling where machine learning formulations are used to model the evolution equations. This includes approaches based on recurrent neural networks that can implicitly find and determine the evolution of internal variables \cite{koeppe2019efficient, mozaffar2019deep, gorji2020potential, wu2020recurrent,chen2021deep, abueidda2021deep} by relying on the internal state parameters of these networks. Jones et al. \cite{jones2021neural} use neural ordinary differential equation solvers for the same purpose.
Other models used feedforward neural networks to update the stress and make explicit use of internal variables that are inaccessible from an experiment (for example from simulations of microstructure representative volumes), which restricts the use of such approaches \cite{huang2020machine, masi2021thermodynamics}.
As an alternative, \cite{vlassis2021sobolev,vlassis2022component,vlassis2022geometric} recently proposed a modular approach for elastoplastic modeling where the elastic law and the yield function evolution are treated as separately trainable data-driven models, inspired by traditional modeling of elastoplasticity. In their approach, beyond training the elastic law which we discussed previously, a neural network-based yield function is trained using a level-set hardening framework that is dependent on the internal variables. 
However, most of the currently proposed data-driven constitutive models for path-dependent materials rely on large quantities of data, in terms of loading paths and stress states, which can only be obtained from lower-scale numerical models. This means that data from classical experiments that provide labeled data for stress-strain pairs do not offer enough information to obtain reliable data-driven models. 
One exception is the work by \cite{tang2020map123}, which proposes a model-free (c.f. \cite{kirchdoerfer2016data, eggersmann2019model}) elastoplastic framework that only uses one-dimensional experimental data. The authors project a three-dimensional stress/strain state onto a point on the plane of a uniaxial stress state. This plane can then be searched for the closest data point in an available material data set, which is finally taken to characterize the constitutive relationship.
The framework avoids explicit mathematical forms of the flow rule, hardening law, and the yielding surface but due to the lack of constraints still requires a large quantity of one-dimensional loading/unloading curves each involving different loading/unloading designs. Furthermore, it is not immediately clear how to reliably achieve extrapolation outside of the training domain with this technique since it relies on nearest-neighbor projections.

The recent drive towards big data has shifted much attention to data-driven approaches that assume direct access to sufficient labeled pairs to train data-hungry models, and at the other end when limited data are available the attention has focused more on parameter estimation, resorting back to analytical models where the coefficients are retrieved through machine learning approaches. These trends point to an obvious gap in the literature, where expressive machine learning--enabled constitutive models for elastoplasticity have not been explored in the low or limited data regime. The recent rise of physics-informed machine learning points to a possible solution, where physical principles and mechanistic assumptions formulated as constraints can be utilized to reduce the data burden and to obtain expressive models without resorting to simple parameter estimation. 

Figure \ref{fig:Categorization} shows a qualitative analysis of recent data-driven modeling frameworks for elastoplastic material modeling depending on the amount of available data and the degree of required constraints to obtain reliable models. 
A large amount of data is necessary to train a trustworthy model if i) the stress evolution laws are directly learned from data without being guided by thermodynamic principles and ii) the internal variables of the evolution are only implicitly represented by microstructure or latent space information, see  e.g. \cite{mozaffar2019deep,abueidda2021deep}.
The amount of required data can be reduced by assuming explicit forms of the internal variables, such as the accumulated plastic strain, which is commonly utilized in phenomenological models. Identifying the physical meaning of internal variables allows one to build efficient datasets that span the complete range of interest of the internal variables. An example of this approach is given in \cite{huang2020machine}.
Thermodynamic restrictions that constrain the stress evolution but without directly specifying a flow direction can be used to further reduce the amount of necessary data, such as employed in \cite{jones2021neural}.
More model constraints are used in \cite{vlassis2022geometric} and \cite{vlassis2021sobolev} where the flow direction is fixed to be coaxial to the plastic potential; however, neither the specific hardening types nor the functional forms of the yield function and its evolution are assumed to be known.  
This leads to a very expressive but still data-hungry framework, i.e. full explorations of the stress space, as well as the space of internal variables, are necessary to obtain reliable models. This amount of data can generally only be obtained by relying on computational microscale simulations. This also means that so far, standard uniaxial tension/compression tests do not offer enough information for building expressive and at the same time reliable models.
This problem is avoided by \cite{tang2020map123} where a three-dimensional stress/strain state is projected onto a point on the plane of a uniaxial stress state. This allows the method to only rely on uniaxial data but still necessitates a considerable amount of data in this limited regime (i.e. different loading/unloading scenarios). 

On the other end of the spectrum, we have classical phenomenological models that for example rely on functional forms such as non-linear kinematic (NLK) hardening rules (see \cite{chaboche1986time}). These models are subject to restrictive simplifying constraints but have proven to be versatile and reliable enough to be the most common user choice for modeling path-dependent material behavior. 
These models generally only require fitting a limited amount of model parameters but can do this with a low amount of data coming, for example, only from uniaxial stress-strain curves. This characteristic, along with their ability to extrapolate beyond experimental stress states/loading paths, has been the key to the lasting prevalence of phenomenological and micromechanical models.
One weakness of these models however is that the functional forms for all components of the model have to be chosen by the user. Hence, their expressiveness may be significantly restricted by user knowledge.

\begin{figure}
    \centering
    \tikzstyle{stuff_fill}=[rectangle,,draw=none,fill=blue!10,text width=3.5cm,node contents={{Thermodynamically consistent constraints}}]
        \tikzstyle{stuff_fill2}=[rectangle,draw=none,fill=red!10,text width=3.0cm,node contents={Thermodynamically consistent ML tools}]
\begin{tikzpicture}

\draw [-{Stealth[slant=0]},line width=2pt] (-1,0)--(12,0) node[pos=.9,below] {\Large{Data}};
\draw [-{Stealth[slant=0]},line width=2pt] ( 0.,-1.) -- ( 0.,5.0) node[above] {\Large{Trustworthiness}};

\draw[line width=2pt] ( 0.,0.) -- ( 10.,4.0);
\draw[dashed, blue,line width=2pt] ( 0.,1.5) -- ( 10.,4.3);
\draw[dashed, red,line width=2pt] ( 0.,2.6) -- ( 10.,4.4);
\draw [->,blue,line width=2pt] (5,2.) to [out=180,in=290](4.1,2.5);
\draw [->,red,line width=2pt] (3,2.4) to [out=180,in=290](2,2.9);
\node at (7.1,1.8) [stuff_fill];
\node at (2.1,3.8) [stuff_fill2];
\end{tikzpicture}
    \caption{Trustworthiness of material behavior descriptions in the context of data-driven material modeling. The trustworthiness qualitatively increases with the amount of available data. However, both thermodynamically consistent modeling constraints, as well as ML tools that adhere to thermodynamic laws, allow for increasing the trust in the model even in the low data domain. }
    \label{fig:Trustworhtiness}
\end{figure}
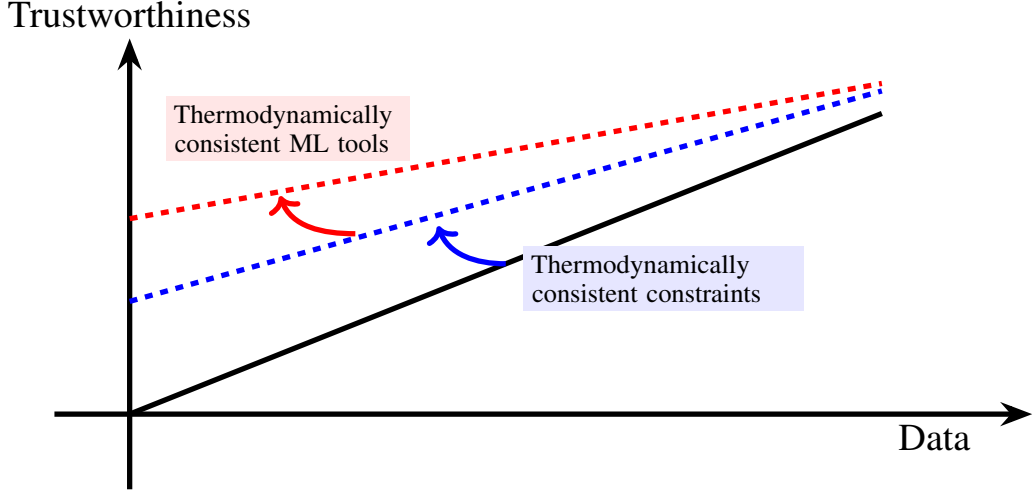

In this work, we explore a thermodynamically consistent framework that positions itself in between the big data domain of (purely) data-driven plasticity modeling and their phenomenological counterparts by proposing a data-driven hardening representation that is free of user-chosen functional forms and which is intended to be trained from simple uniaxial experimental data. The approach is based on modular elastoplastic modeling that is constrained by thermodynamic principles and adheres to mechanistic assumptions that have been commonly utilized in the derivation of phenomenological laws.   
However, there is a distinction between the consistency of the theoretical model and the consistency of the data-driven tools, which is especially critical in the low data domain, see Figure \ref{fig:Trustworhtiness}. This is commonly the case with soft constraints added in the loss function of neural networks.
Hence, we propose data-driven models that intrinsically conform to convexity and monotonicity requirements based on mechanistic assumptions that are needed to maintain thermodynamic consistency.
This work is structured as follows.
In Section \ref{sec::2} we briefly review the basic concepts of elastoplasticity and discuss them in the context of modular data-driven modeling given experimental data from uniaxial tests.
Section \ref{sec::3} focuses on the main contribution of this work. We explore how constrained data-driven tools can be used to model hardening behavior directly from stress-strain curves without needing to specify functional forms.
The proposed framework is demonstrated on benchmark tests as well as on experimental data in Section \ref{sec::4}.
The paper is concluded in Section \ref{sec::5} with a summary of the developments and ideas for future work.
\begin{figure}
    \centering
\def\spc{7pt} 
\setlist[itemize]{wide=0pt, leftmargin=*, itemsep=0pt, topsep=0pt, after=\vspace*{-\baselineskip}, rightmargin=-\leftmargini}
\setlength{\extrarowheight}{3pt}
\begin{tikzpicture}
    \node[inner sep=\spc] (t)
        {
\begin{tabularx}{0.85\textwidth}{|X|X|X|}
\hline
Constraints & ML discovery & Examples\\
\hline
\begin{itemize}
\item Explicit internal variables
\item Fixed model functional form
\item Fixed flow direction
\item Fixed hardening types
\end{itemize}
&
\begin{itemize}
 \item
Model parameters
\end{itemize}
&
\begin{itemize}
\item
Phenomenological modeling
 \item
NLK models \cite{chaboche1986time}
\end{itemize}
 \\
  \arrayrulecolor{black!30}\midrule
  \begin{itemize}
\item Explicit internal variables
\item Fixed flow direction
\item Fixed hardening types
\end{itemize}
&
\begin{itemize}
 \item
Functional forms
\end{itemize}
&
\begin{itemize}
\item
\color{red}{Our work}
\end{itemize}
 \\
   \arrayrulecolor{black!30}\midrule
  \begin{itemize}
\item Explicit internal variables
\item Fixed flow direction
\item Projection on uniaxial data
\end{itemize}
&
\begin{itemize}
 \item Functional forms
\item Hardening types
\end{itemize}
&
\begin{itemize}
\item Tang et. al. \cite{tang2020map123}
\end{itemize}
 \\
  \arrayrulecolor{black!30}\midrule
  \begin{itemize}
\item Explicit internal variables
\item Fixed flow direction
\end{itemize}
&
\begin{itemize}
 \item Functional forms
\item Hardening types
\end{itemize}
&
\begin{itemize}
\item Masi et. al. \cite{masi2021thermodynamics}
\item Vlassis and Sun \cite{vlassis2021sobolev}
\item Jones et. al. \cite{jones2018machine}
\end{itemize}
 \\
   \arrayrulecolor{black!30}\midrule
  \begin{itemize}
\item Latent space\newline  internal variables
\item Fixed flow direction
\end{itemize}
&
\begin{itemize}
 \item Functional forms
\item Hardening types
\end{itemize}
&
\begin{itemize}
\item Vlassis and Sun \cite{vlassis2022geometric}
\end{itemize}
 \\
    \arrayrulecolor{black!30}\midrule
  \begin{itemize}
  \item Thermodynamically \newline constrained flow
\item Latent space \newline  internal variables
\end{itemize}
&
\begin{itemize}
\item Stress evolution law
\end{itemize}
&
\begin{itemize}
\item Jones et. al. \cite{jones2021neural}
\end{itemize}
 \\
     \arrayrulecolor{black!30}\midrule
  \begin{itemize}
  \item No thermodynamic \newline constraints
\item Explicit internal variables
\end{itemize}
&
\begin{itemize}
\item Stress evolution law
\end{itemize}
&
\begin{itemize}
\item Huang et. al. \cite{huang2020machine}
\end{itemize}
 \\
      \arrayrulecolor{black!30}\midrule
  \begin{itemize}
\item No thermodynamic \newline constraints
\item Latent space/ implicit \newline  internal variables
\end{itemize}
&
\begin{itemize}
\item Stress evolution law
\end{itemize}
&
\begin{itemize}
\item Mozaffar et. al. \cite{mozaffar2019deep}
\item Abueidda et. al. \cite{abueidda2021deep}
\end{itemize}
 \\
\bottomrule
\end{tabularx}
};
\draw [-{Stealth[slant=0]},line width=2pt] (-7.5,-8.0)--(-7.5,8.0) node[pos=1.0,above] {\Large{Constraints}};
\draw [-{Stealth[slant=0]},line width=2pt] (7.5,8.0)--(7.5,-8.0) node[pos=1.0,below] {\Large{Data}};

\end{tikzpicture}
    \caption{Quantitive categorization of data-driven modeling of plastic material behavior. More available data correlates to needing less constitutive modeling constraints in order to obtain sufficiently working data-driven models. }
    \label{fig:Categorization}
\end{figure}

\section{Plasticity theory in light of modular elastoplastic modeling and uniaxial experimental data}\label{sec::2}
Towards incorporating more mechanistic information in our data-driven approach, which we will cover in the following section,  we focus on a small strain time-independent elastoplastic framework under isothermal conditions. This allows us to additively decompose the strain into its elastic and plastic components
\begin{equation}
    \bm{\epsilon} = \bm{\epsilon}^{e} + \bm{\epsilon}^{p}.
\end{equation}
Non-associative plasticity in general requires the description of a free energy $\psi$, a yield surface $f$, and a dissipation potential $F$.
We postulate that the free-energy function can be split into its elastic and plastic contributions
\begin{equation}
    \psi( \bm{\epsilon}^{e},\bm{\Gamma}_{1},  \ldots , \bm{\Gamma}_{n}) = \psi^{e}( \bm{\epsilon}^{e}) + \psi^{p}(\bm{\Gamma}_{1}, \ldots, \bm{\Gamma}_{n} )
\end{equation}
where $\bm{\Gamma}_{i}$, $i=1,\ldots, n $ denote a set of internal hardening variables of arbitrary order. 
The first and second laws of thermodynamics define the intrinsic dissipation $\mathcal{D}$ to be non-negative 
\begin{equation}\label{eq::Diss}
    \mathcal{D} = \bm{\sigma} : \dot{\bm{\epsilon}}^{p} -  \sum_{i} \bm{A}_{i}  :  \dot{\bm{\Gamma}}_{i}\geq 0 
\end{equation}
where we have defined 
\begin{equation}\label{eq::ThermoForces}
\begin{aligned}
\bm{\sigma} = \frac{\partial \psi}{\partial \bm{\epsilon}^{e}}, \qquad
\bm{A}_{i} =  \frac{\partial \psi}{\partial \bm{\Gamma}_{i}}
\end{aligned}
\end{equation}
following the Coleman-Noll procedure \cite{coleman1974thermodynamics}.
In general elastoplastic modeling these variables can be used to define the scalar-valued yield function $f(\bm{\sigma},\bm{A}_{i})$ which defines the plastically admissible domain as the set of stresses that satisfy $f(\bm{\sigma},\bm{A}_{i})\leq 0$. 

The stress $ \bm{\sigma}$ and the hardening thermodynamic forces $\bm{A}_{i} $ have to be defined such that eq. \eqref{eq::Diss} is always fulfilled.
A common approach is based around the definition of a pseudo-potential (or dissipation potential) $F(\bm{\sigma}, \bm{A}_{i})$ \cite{chaboche1991some,chaboche2008review,besson2009non,skrzypek2015mechanics} which determines the direction of plastic flow using the generalized normality rule \cite{halphen1975materiaux}
\begin{equation}\label{eq::GenNormRule1}
\begin{aligned}
\dot{\bm{\epsilon}}^{p} = \dot{\lambda} \frac{\partial F}{\partial \bm{\sigma}}, \qquad
\dot{\bm{\Gamma}}_{i} = \dot{\lambda} \frac{\partial F}{\partial \bm{A}_{i}}
\end{aligned}
\end{equation}
where $\dot{\lambda}$ is known as the plastic multiplier.
In order to enforce thermodynamic consistency, the pseudo-potential is required to adhere to the following principles
\begin{itemize}
    \item $F$ is convex with respect to all its arguments,
    \item $F$ is positive at the onset of plastic flow ($F\geq f = 0$),
    \item $F$ fulfills the condition
    $F(\bm{\sigma}, \bm{A}_{i}) - F(\bm{0}, \bm{0})\geq 0$.
\end{itemize}
A common choice to fulfill these conditions on $F$ is given by \cite{lemaitre1994mechanics,auricchio1995two}
\begin{equation}\label{eq::DissSplit}
    F(\bm{\sigma}, \bm{A}_{i}) = f(\bm{\sigma}, \bm{A}_{i}) + \phi ( \bm{A}_{i})
\end{equation}
where $f$ and $\phi$ are convex, $f(\bm{0}, \bm{0})< 0$,  $\phi(\bm{0})= 0$ and $\phi( \bm{A}_{i})\geq 0$,  and the latter function adds further nonlinearities to the formulation.
In essence, this elastoplastic modeling framework consists of 5 components, which would need to be specialized:\label{page:5Comp}
\begin{enumerate}
    \item A choice of the hardening parameters $\bm{A}_{i}$,
    \item An elastic model $\bm{\sigma} = \hat{\bm{\sigma}}(\bm{\epsilon}^{e})$ that describes the material behavior inside the plastically admissible domain which is derived from the elastic component of the free energy,
    \item A convex model for the yield function $f$,
    \item The plastic portion of the free energy function $\psi^{p}$,
    \item A convex model $\phi(\bm{A}_{i})$.
\end{enumerate}

In general, any of the above components can be discovered from data, or directly prescribed using existing models, allowing for a modular elastoplastic modeling approach depending on data availability; this modularity is what we will exploit in our framework. As discussed in the introduction, the first component, namely, the choice of internal variables, can be explicitly determined by the user or implicitly obtained from e.g. a latent space. 
For the remaining components (2-5), the choice has to be made depending on prior knowledge, the amount of data, and arising complexities of each individual component.
This allows for a seamless transition and exchange between classical constitutive modeling and the new data-driven approaches that have been proposed for the elastic part \cite{fuhg2021local,fuhg2022physics} or the yield function \cite{fuhg2022machine,fuhg:hal-03619186}. It should be remarked that our approach is in essence similar to the data-driven elastoplastic framework proposed in \cite{vlassis2021sobolev}, but in their work, the authors only make an implicit distinction between the initial yield function and its evolution.

Specifying the elastic response (component 2) as well as the form of the initial yield function -- prior to yielding -- (component 3) are path-independent problems which means we can rely on one-to-one pairs of labeled data to develop fits with machine learning tools. We briefly review the process of selecting models of these two components (2,3) in the context of the proposed modular framework. It is crucial to discuss these choices under consideration of uniaxial data in the following section. In Section \ref{sec::3} we discuss the more complex issue of the remaining path-dependent components (4,5) in the context of data-driven hardening.

\subsection{Modeling of the elastic response}\label{subsec::ElasticModeling}
\begin{figure}
\begin{subfigure}[b]{0.48\linewidth}
        \centering
    \includegraphics[scale=0.3]{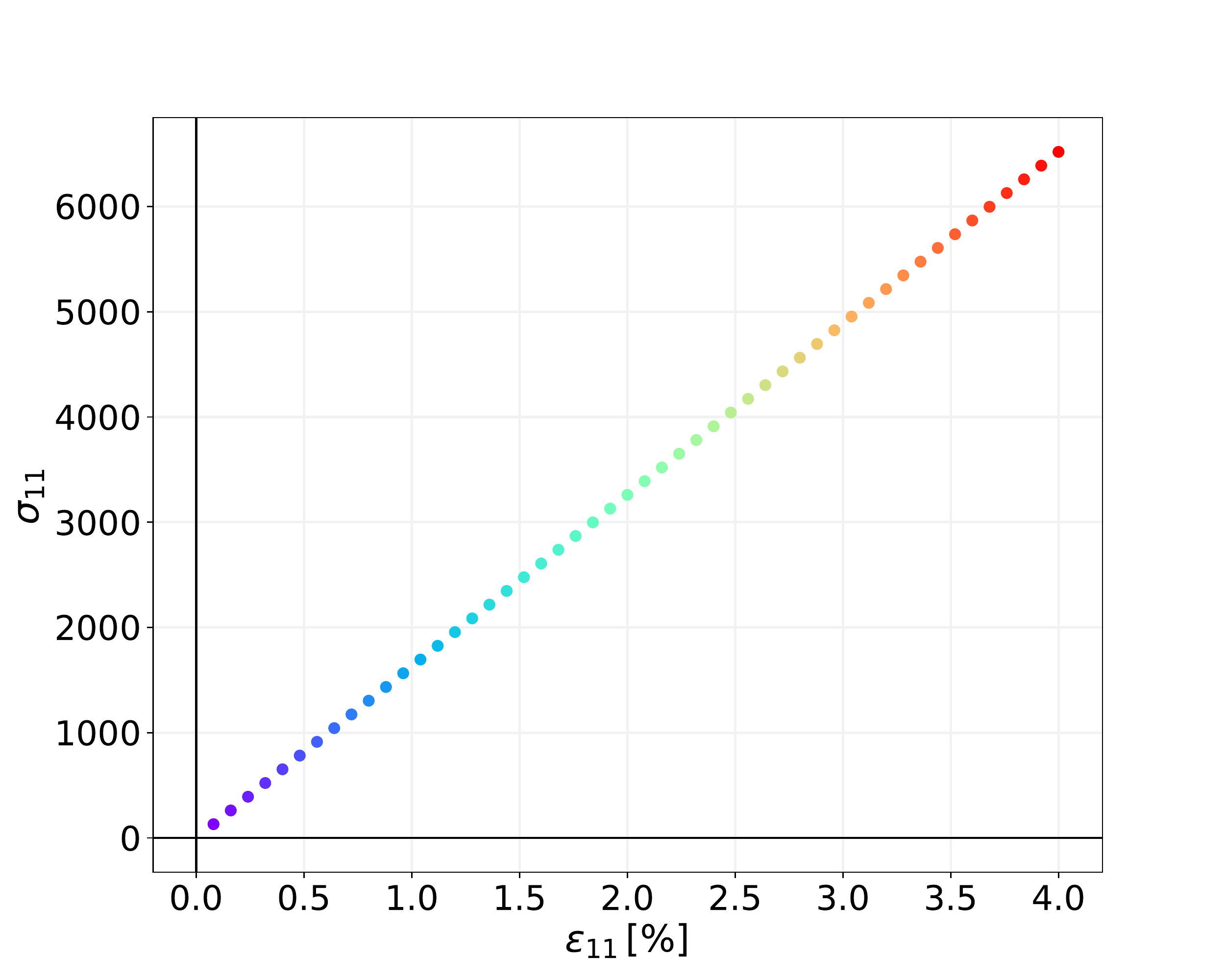}
    \caption{}\label{fig::ElasticUnia}
\end{subfigure}
\begin{subfigure}[b]{0.48\linewidth}
        \centering
    \includegraphics[scale=0.3]{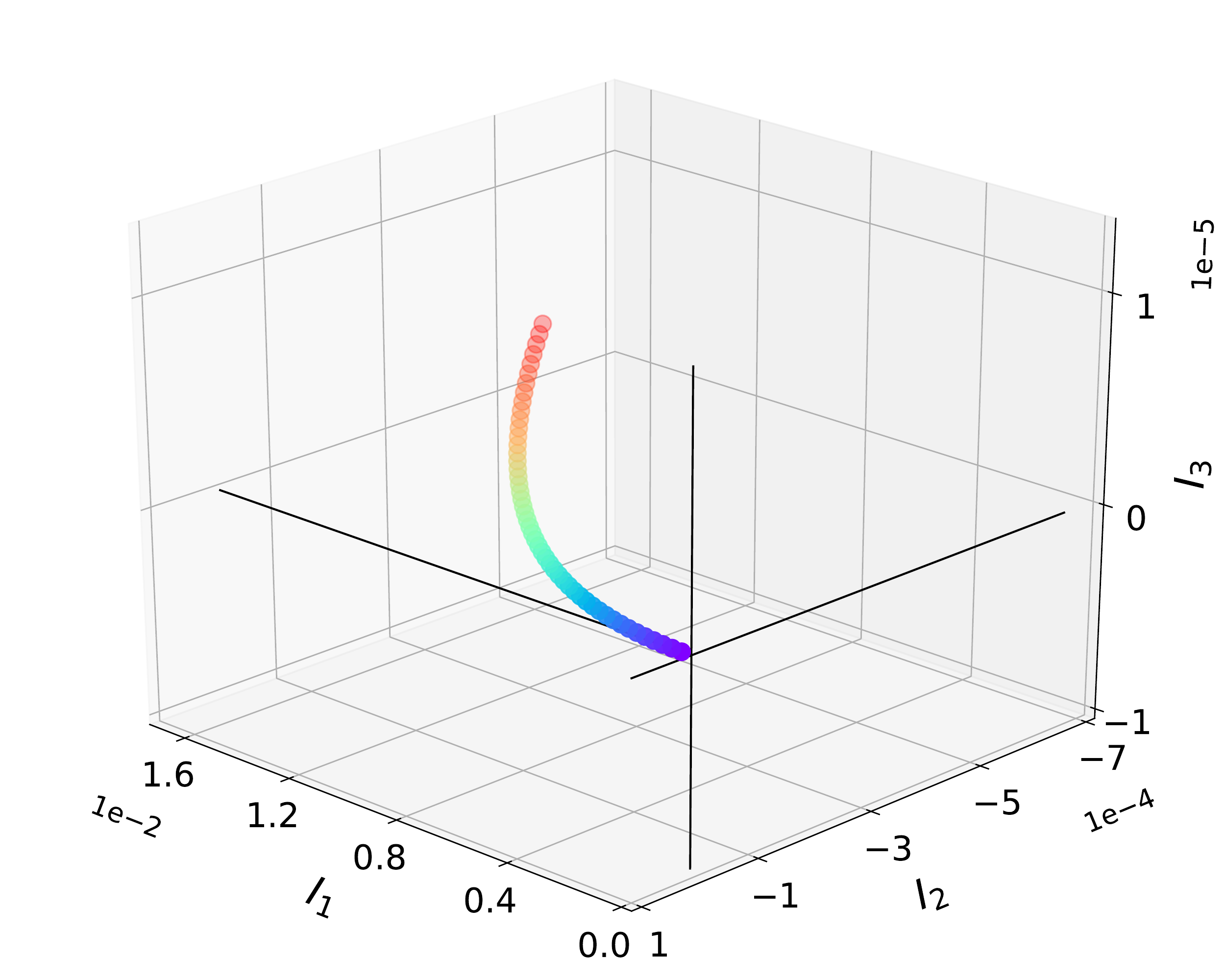}
    \caption{}\label{fig::ElasticUnib}
\end{subfigure}
    \caption{Invariant space path of a uniaxial tension experiment. (a) Linear elastic stress-strain curve, (b) Invariant space representation of curve shown in (a). Colors indicate matching path positions.}\label{fig::ElasticUni}
    \end{figure}

It has been recently showcased that the incorporation of physical constraints in data-driven modeling of elastic material behavior  \cite{frankel2020tensor,fuhg2022physics,klein2022polyconvex,kalina2022automated,FUHG2022105022} leads to more robust models compared to earlier machine learning-based approaches. Here we briefly review using the theory of representation for tensor functions towards this goal. If we assume an isotropic material behavior then the free energy is dependent on three invariants $\psi^{e}(\bm{\epsilon^{e}}) = \psi^{e} (I_{1},I_{2}, I_{3})$
which read
\begin{equation}
\begin{aligned}
I_{1} = \text{tr}(\bm{\epsilon}), \qquad 
I_{2} = 0.5 (\text{tr}(\bm{\epsilon})^{2} - \text{tr}(\bm{\epsilon}^{2})), \qquad
I_{3} = \text{det}(\bm{\epsilon}) .
\end{aligned}
\end{equation}
Employing the representation theorem for an isotropic tensor function the resulting stress response can always be decomposed into a three-part linear combination given by 
\begin{equation}\label{eq::IsoMapping}
\bm{\sigma}(I_{1},I_{2},I_{3}) = c_{1}(I_{1},I_{2},I_{3}) \bm{I} + c_{2}(I_{1},I_{2},I_{3})  \bm{\epsilon} + c_{3}(I_{1},I_{2},I_{3})  \bm{\epsilon}^{-1}
\end{equation}
where the scalar functions $c_{i}$ are only dependent on the invariants. Hence, given a dataset consisting of strain-stress pairs, we can learn a functional mapping of the form
\begin{equation}
\begin{bmatrix}
I_{1} \\ I_{2} \\ I_{3}
\end{bmatrix}
\rightarrow
\begin{bmatrix}
c_{1} \\ c_{2} \\ c_{3}
\end{bmatrix}
\end{equation}
and take the prediction values (here denoted by $\hat{\bullet}$) to get the stress prediction
\begin{equation}
    \hat{\bm{\sigma}} = \hat{c}_{1}(I_{1},I_{2},I_{3}) \bm{I} + \hat{c}_{2}(I_{1},I_{2},I_{3})  \bm{\epsilon} + \hat{c}_{3}(I_{1},I_{2},I_{3})  \bm{\epsilon}^{-1}.
\end{equation}
Similar representations are available for anisotropic behavior. This data-driven framework has proven to possess excellent interpolation and extrapolation qualities \cite{fuhg2022physics} for problems at finite deformations where data of the whole (but limited) stress-strain space is available. 
However, as seen in Figure \ref{fig::ElasticUni} a simple uniaxial experiment (here just tension) of an elastic material only offers very limited information (spread) of the input space, see Figure \ref{fig::ElasticUni} (b).
Hence, training a full data-driven tool that takes the invariant information as input while relying only on this dataset, would not lead to a reliable and trustworthy model.
On the other hand, phenomenological models which also constrain the functional space of the model fit have been shown to yield reliable results given this limited amount of data.
For example, assuming linear elasticity for the elastic part of the free energy reads
\begin{equation}\label{eq:LinElasticity}
    \psi^{e} = \frac{1}{2} \bm{\epsilon}^{e}: \mathbb{C}: \bm{\epsilon}^{e}
\end{equation}
where for an isotropic material the fourth order modulus tensor $\mathbb{C}$ is given by
\begin{equation}
    \mathbb{C}_{ijkl} = \frac{E \nu}{(1+\nu) (1-2\nu)} \delta_{ij} \delta_{kl} + \frac{E}{2 (1+ \nu)} (\delta_{ik} \delta_{jl} + \delta_{il} \delta_{jk}).
\end{equation}
Here, $E$ and $\nu$ denote Young's modulus and Poisson's ratio respectively. These two material parameters can easily be obtained from stress-strain curves (e.g. as given by Figure \ref{fig::ElasticUni}), of course only, if a linear elastic response can be observed. 
Similarly, a two-parameter nonlinear elastic law such as the Neo-Hookean law could be assumed in case the response follows a nonlinear trend \cite{holzapfel2000nonlinear}. 

Hence, if no additional knowledge (other than a uniaxial test) about the elastic response of a material is known, phenomenological models have to be the preferred choice due to their ability to provide trustworthy fits even for this limited amount of data. If additional information, for example from microstructural simulations, is available, the use of data-driven models as suggested above might be advantageous.
In the following, we will represent the mapping to obtain the elastic response by $\hat{\bm{\sigma}}(\bullet)$ which could either be derived from a phenomenological or a data-driven model depending on the amount of available information without loss of generality.
\subsection{Modeling of initial yield function}\label{subsec::YieldModeling}
\begin{figure}
\begin{subfigure}[b]{0.48\linewidth}
        \centering
    \includegraphics[scale=0.3]{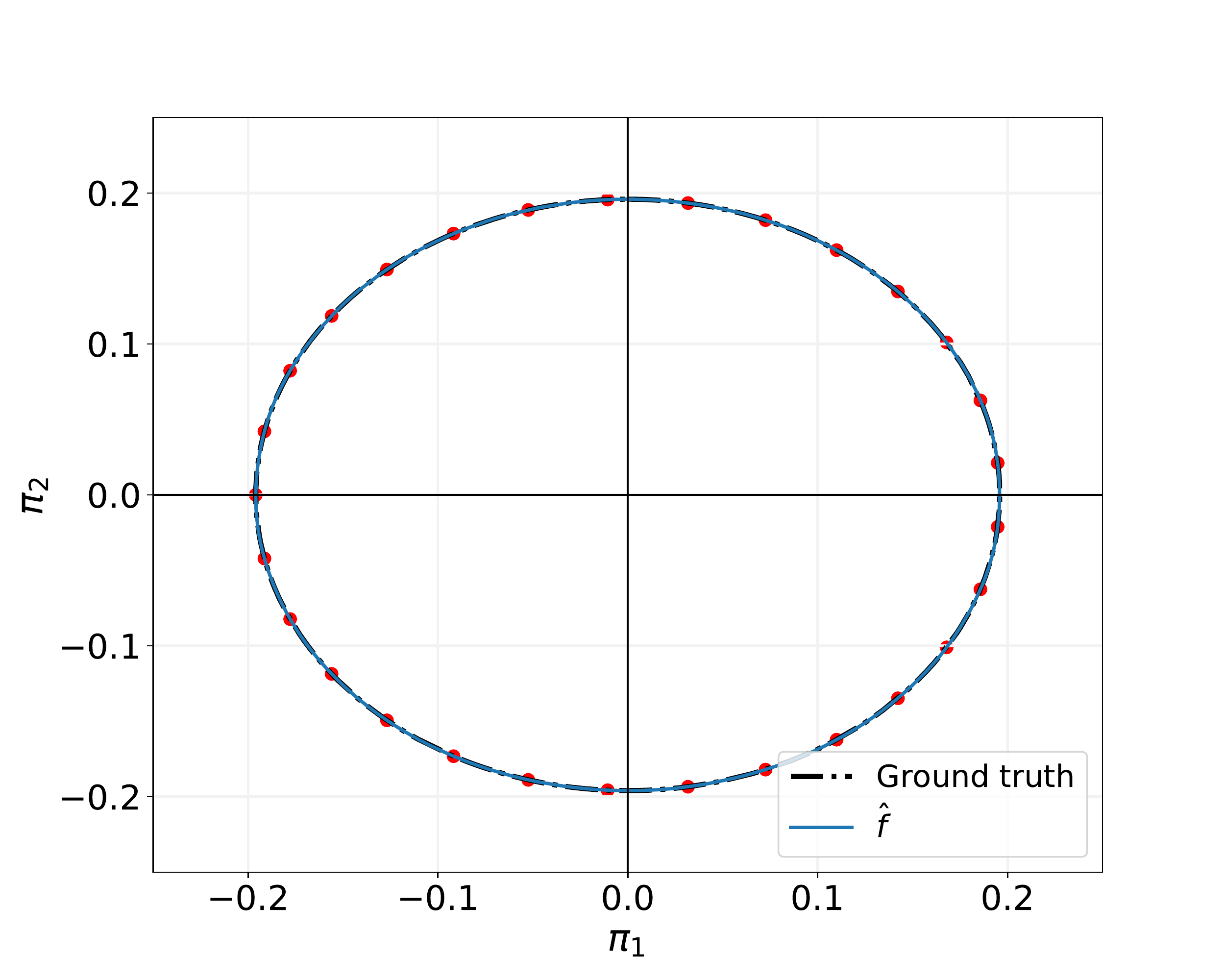}
    \caption{Von Mises yield surface}
\end{subfigure}
\begin{subfigure}[b]{0.48\linewidth}
        \centering
    \includegraphics[scale=0.3]{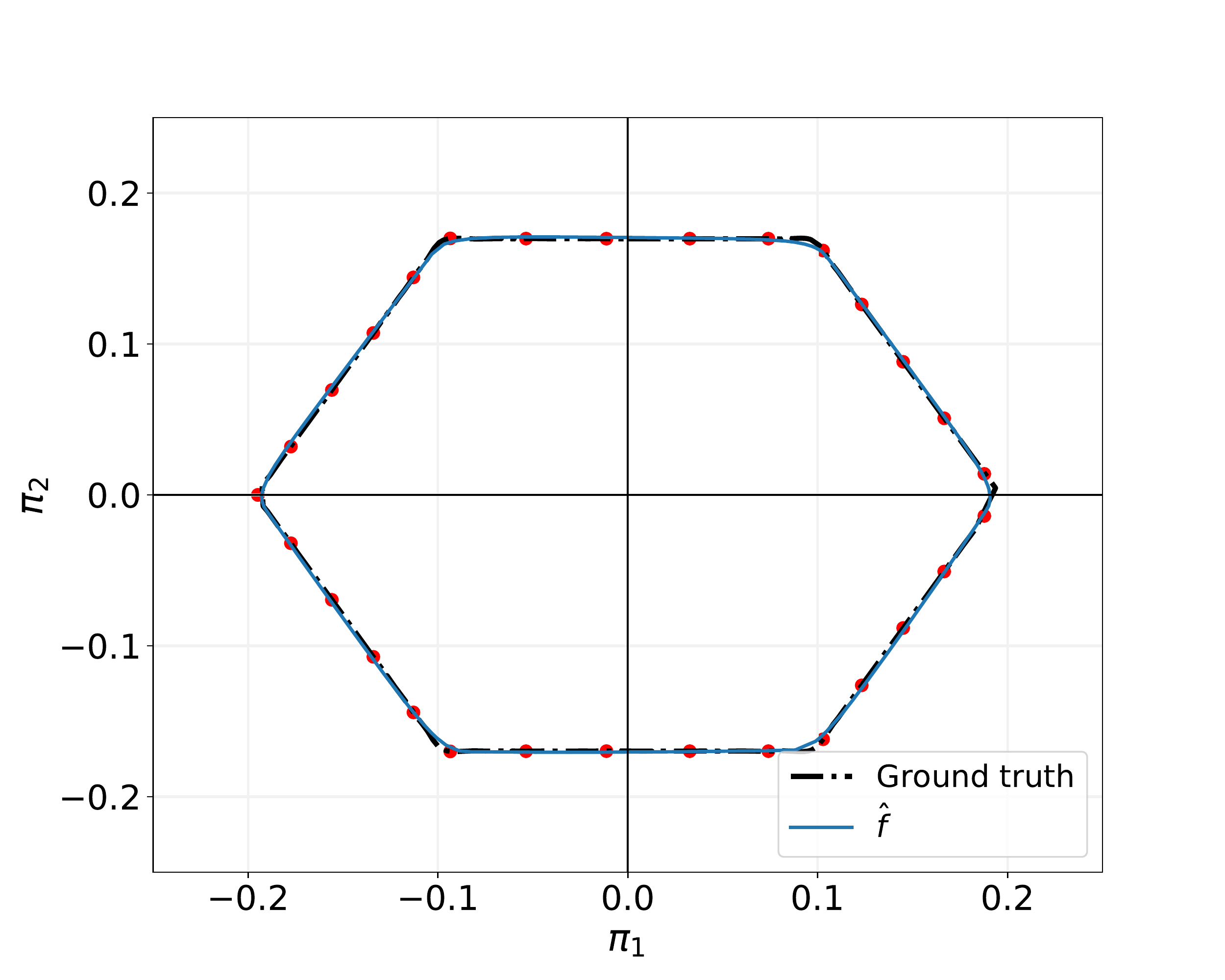}
    \caption{Tresca yield surface}
\end{subfigure}

\begin{subfigure}[b]{1.0\linewidth}
        \centering
    \includegraphics[scale=0.3]{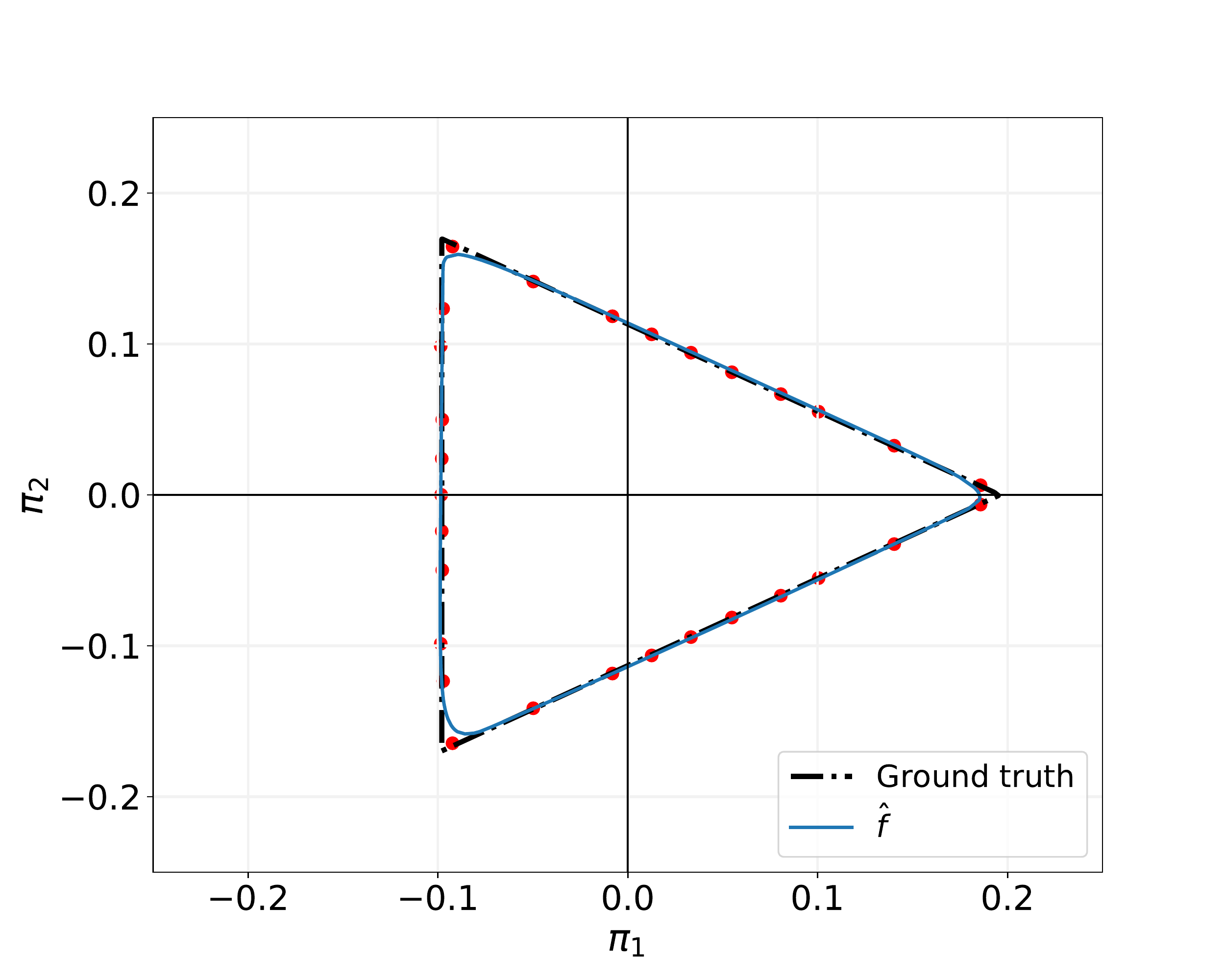}
    \caption{Ivlev yield surface}
\end{subfigure}
    \caption{Data-driven yield function modeling in $\pi$-plane using input convex neural networks as suggested by \cite{fuhg2022machine}. Red dots indicate training data positions.}\label{fig:DataDrivenYieldFun}
    \end{figure}

Data-driven modeling of yield functions has recently received more and more attention \cite{vlassis2021sobolev,fuhg2022machine,flaschel2022discovering, vlassis2022component, fuhg:hal-03619186} due to the flexibility of these models compared to their traditional phenomenological counterparts.
Assuming an isotropic shape of the virgin yield surface, the yield function is only dependent on the principal stresses $(\sigma_{1}, \sigma_{2}, \sigma_{3} )$. The principal stress space is also known as the Haigh–Westergaard space \citep{slater1977engineering}. 
\begin{figure}
    \centering
    \includegraphics[scale=0.7]{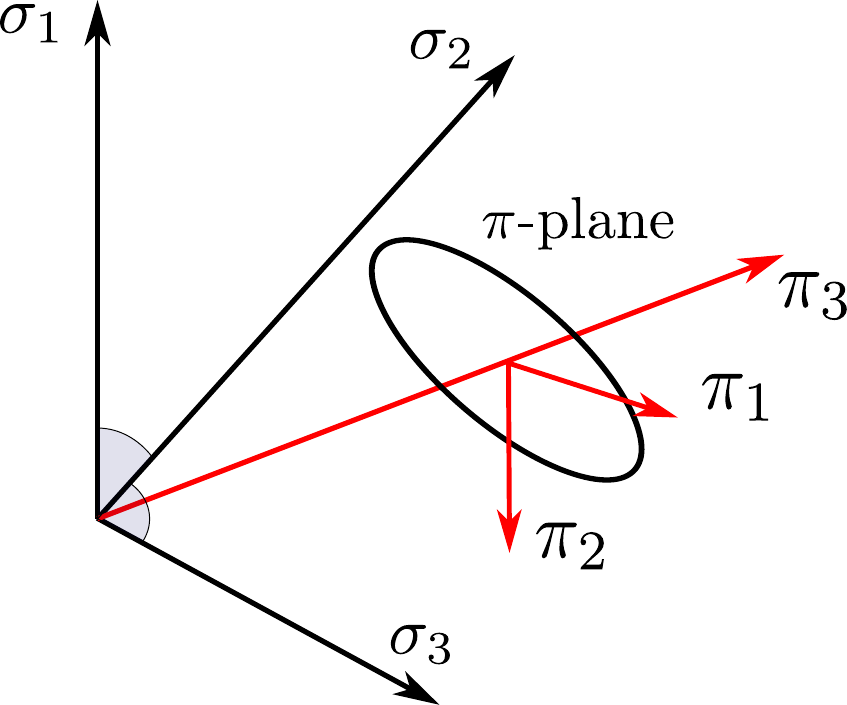}
    \caption{Representations of the Haigh–Westergaard space $(\sigma_{1}, \sigma_{2}, \sigma_{3} )$ and the $\pi$-plane with the coordinate system $(\pi_{1},\pi_{2},\pi_{3})$ where $\pi_{3}$ is aligned with the hydrostatic axis.}
    \label{fig:piPlaneRep}
\end{figure}
In order to decompose the stress tensor into deviatoric and hydrostatic components the coordinates $(\pi_{1},\pi_{2},\pi_{3})$ can be derived from
\begin{equation}
    \begin{bmatrix}
    \pi_{1}\\
    \pi_{2}\\
    \pi_{3}
    \end{bmatrix} = 
    \begin{bmatrix}
\sqrt{\frac{2}{3}} & - \sqrt{\frac{1}{6}}& - \sqrt{\frac{1}{6}} \\
0 &  \sqrt{\frac{1}{2}} & - \sqrt{\frac{1}{2}}\\
 \sqrt{\frac{1}{3}} & \sqrt{\frac{1}{3}} & \sqrt{\frac{1}{3}}
    \end{bmatrix}
           \begin{bmatrix}
    \sigma_{1}\\
    \sigma_{2}\\
    \sigma_{3}
    \end{bmatrix} .
\end{equation}
    Introduced in \cite{de1962elasticidade} the $\pi$-plane represents a cross-sectional cut through the yield surface at $\pi_{3}$=const. Since the $\pi_{3}$-coordinate represents the hydrostatic axis, the yield surface in the $\pi$-plane does not depend on $\pi_{3}$ for pressure-insensitive materials, see Figure 
    \ref{fig:piPlaneRep}. In in these cases $f(\pi_{1},\pi_{2})$ is a sufficient representation of the yield function.
    It can be proven that if a yield function is convex in the three-dimensional space of the principal stresses (and hence also in the space of the $\pi$-coordinate system) then it is also convex with regards to the more general six-dimensional stress space \cite{lippmann1970matrixungleichungen}. Hence, data-driven models for the yield function have been proposed that are intrinsically convex allowing to model convex yield surfaces very reliably and accurately \cite{fuhg2022machine,fuhg:hal-03619186} if training data on the yield surface is available, see Figure \ref{fig:DataDrivenYieldFun}.

However, a single uniaxial experiment only offers a single point to describe the whole initial yield surface.
Hence, similarly to modeling the elastic case, in the low data-regime classical phenomenological yield functions allow for more trustworthy modeling.
In the case of a single uniaxial experiment, an initially isotropic yield function is commonly assumed. The functional form is a user-choice, but a common representation is given the following
von Mises \cite{mises1913mechanik} which can be reformulated (from their classical functional form) in terms of the $\pi$-plane coordinates \cite{flaschel2022discovering} as 
\begin{equation}\label{eq:vonMisesRefor}
    f (\pi_{1}, \pi_{2}) = \sqrt{\frac{3}{2}} \sqrt{\pi_{1}^{2} + \pi_{2}^{2}}  - \sigma_{y}
\end{equation}
where $\sigma_{y}$ can be fit to the one available data point. 
If additional knowledge about the shape of the yield function is available (obtained from microstructural simulations or other experiments), then data-driven yield function representations may be preferred.
In this work this reformulation of eq. \eqref{eq:vonMisesRefor} is done so that the phenomenological models of the yield functions follow the same functional mapping $(\pi_{1}, \pi_{2}, \pi_{3}) \rightarrow \mathbb{R}$ as the data-driven tools. This will allow both data-driven and classical models to be easily used and exchanged with each other in the proceeding formulations of hardening.

\section{Data-driven hardening for uniaxial experimental data and modular elastoplasticity}\label{sec::3}
An important aspect of path-dependent material modeling, as highlighted by the 5 components of the elastoplastic framework of this work introduced on p. \pageref{page:5Comp}, is the identification of the appropriate internal variables. A data-driven approach for this process is commonly based on implicit or latent space internal variables that are directly obtained from data such as microstructural information \cite{frankel2019predicting,frankel2022mesh}. This enables the extraction of the important features that model a path-dependent process without significant user intervention.
However, in the low data regime, based on experimental data, this process cannot easily be conducted.
Hence, in this case, which is of interest to the current approach, internal variables can be chosen based on prior mechanistic knowledge and informed user choices. The most common choice which has been used in the field for over 50 years \cite{eisenberg1968nonlinear,dafalias1976plastic, chaboche1979modelization} is based around the following two internal variables $\lbrace r, \bm{\alpha}\rbrace$, where $r$ is a scalar variable for isotropic hardening and $\bm{\alpha}$ is a tensor-valued kinematic hardening variable. Since this work is focusing on the case of limited available data, we explicitly follow the latter approach.

Employing this strategy, the free energy function can be split into its elastic and plastic component
\begin{equation}\label{eq::NewPsi}
    \psi = \psi(\bm{\epsilon}^{e},r, \bm{\alpha}) = \psi^{e}(\bm{\epsilon}^{e}) + \psi^{p}(r, \bm{\alpha}).
\end{equation}
Following eqs. \eqref{eq::Diss} and \eqref{eq::ThermoForces} allow us to obtain the corresponding thermodynamic forces for each hardening variable as
\begin{equation}\label{eq::NewThermoForce}
    R = \frac{\partial \psi^{p}}{\partial r}, \qquad \bm{X} = \frac{\partial \psi^{p}}{\partial \bm{\alpha}}.
\end{equation}
Hence, compared with eq. \ref{eq::DissSplit}, we can write the dissipation potential as a combination of a yield function $ f (\bm{\sigma},R,\bm{X})$ 
and a convex function $ \phi(R,\bm{X})$, i.e.
\begin{equation}
 F (\bm{\sigma}, R, \bm{X})  =   f (\bm{\sigma},R,\bm{X}) + \phi(R,\bm{X}).
\end{equation}
This means that a general (isotropic) yield function could for example be represented by the invariants of the input tensors
\begin{equation}
    f (\pi_{1}, \pi_{2}, \pi_{2},R,I_{1,X},I_{2,X},I_{3,X})
\end{equation}
where $\lbrace I_{1,X}, I_{2,X}, I_{3,X} \rbrace$ are a set of invariants of $\bm{X}$. This representation results in a potentially seven-dimensional input space that needs to be explored with data. To reduce the dimension of the input space it is common practice to make the yield function dependent on the difference between $\bm{\sigma}$ and $\bm{X}$ instead of focusing on both separately. Hence, we can write $f (\bm{\sigma}-\bm{X},r)$ which allows us then two write the yield function as
\begin{equation}\label{eq:GeneralR}
    f (\overline{\pi}_{1}, \overline{\pi}_{2}, \overline{\pi}_{3},R)
\end{equation}
where $\lbrace \overline{\pi}_{1}, \overline{\pi}_{2}, \overline{\pi}_{3} \rbrace$ represents the set of $\pi$-coordinates (as introduced in Figure \ref{fig:piPlaneRep}) of the tensor $\bm{\sigma} - \bm{X}$.
Using this functional form, the authors in \cite{vlassis2021sobolev,vlassis2022geometric} propose a level-set hardening framework. However, this approach requires access to a lot of data of evolving level sets that can not be obtained from simple experiments.  
Hence, in this work, we adjust the general formulation of eq. \eqref{eq:GeneralR} by allowing the thermodynamic force $R$ to be the ratio of a homothetic transformation of the initial yield function given by 
\begin{equation}
    f (R \, \overline{\pi}_{1}, R \, \overline{\pi}_{2}, R \, \overline{\pi}_{3})
\end{equation}
which is equivalent to the classical understanding of isotropic hardening as a widening or narrowing of the yield surface around its center. Here the value of $R \geq 0$ defines the type of isotropic hardening
\begin{equation}
     \begin{cases}
        &R > 1: \text{ softening} \\
        &R < 1: \text{ hardening} \\
        &R = 1: \text{ no isotropic hardening.}
    \end{cases}
\end{equation}
This approach has been proposed by 
\cite{boehler1987rational} and can also be extended to anisotropic yield functions.

If we furthermore assume a pressure-independent yield function which is a common assumption for metals then we arrive at the expression
\begin{equation}
    f (R \, \overline{\pi}_{1}, R \, \overline{\pi}_{2}).
\end{equation}
This finally leads to the following formulation for the dissipation potential
\begin{equation}
    F (\bm{\sigma}, R, \bm{X})   = f(R \, \overline{\pi}_{1}, R \, \overline{\pi}_{2})    + \phi(R,\bm{X})
\end{equation}
which can be used with the generalized normality rule as introduced in eq. \eqref{eq::GenNormRule1} to find
\begin{equation}\label{eq::GenNormRule2}
\begin{aligned}
        \dot{\bm{\epsilon}}^{p} = \dot{\lambda} \frac{\partial F}{\partial \bm{\sigma}}, \qquad 
        \dot{r} = - \dot{\lambda} \frac{\partial F}{\partial R}, \qquad 
        \dot{\bm{\alpha}} = - \dot{\lambda} \frac{\partial F}{\partial \bm{X}}.
\end{aligned}
\end{equation}
The (pseudo) time-derivatives of the thermodynamic forces read
\begin{equation}
\begin{aligned}
        \dot{R} &=  \dot{\frac{\partial \psi}{\partial r}} =  \frac{\partial^{2} \psi}{\partial \bm{\epsilon}^{e}\partial r} : \dot{\bm{\epsilon}}^{e} +  \frac{\partial^{2} \psi}{\partial r^{2}} \dot{r} +  \frac{\partial^{2} \psi}{\partial r \partial \bm{\alpha}} : \dot{\bm{\alpha}}
        = -  \dot{\lambda}  \left( \frac{\partial^{2} \psi}{\partial r^{2}}  \frac{\partial F}{\partial R} + \frac{\partial^{2} \psi}{\partial r \partial \bm{\alpha}} :  \frac{\partial F}{\partial \bm{X}}\right)
\end{aligned}
\end{equation}
and
\begin{equation}\label{eq:EvoX1}
\begin{aligned}
    \dot{\bm{X}} &=  \dot{\frac{\partial \psi}{\partial \bm{\alpha}}} = \frac{\partial^{2} \psi}{\partial \bm{\epsilon}^{e} \partial\bm{\alpha}} : \dot{\bm{\epsilon}}^{e} +  \frac{\partial^{2} \psi}{\partial \bm{\alpha} \partial r} \dot{r} +   \frac{\partial^{2} \psi}{\partial \bm{\alpha} \partial \bm{\alpha}} : \dot{\bm{\alpha}}  
= -  \dot{\lambda}  \left(  \frac{\partial^{2} \psi}{\partial \bm{\alpha}\partial r}  \frac{\partial F}{\partial R} + \frac{\partial^{2} \psi}{\partial \bm{\alpha} \partial \bm{\alpha}} : \frac{\partial F}{\partial \bm{X}} \right).
\end{aligned}
\end{equation}
Lastly, the typical loading-unloading criterion for plastic modeling requires that $f=0$ and $\dot{f}=0$ during plastic flow. The latter reads
\begin{equation}
    \frac{\partial f}{\partial \bm{\sigma}} : \dot{\bm{\sigma}} + \frac{\partial f}{\partial R} \dot{R} + \frac{\partial f }{\partial  \bm{X}} : \dot{\bm{X}} = 0.
\end{equation}

\subsection{Specific modeling choices for uniaxial experimental data}
\begin{figure}
\begin{subfigure}[b]{0.33\linewidth}
        \centering
    \includegraphics[scale=0.22]{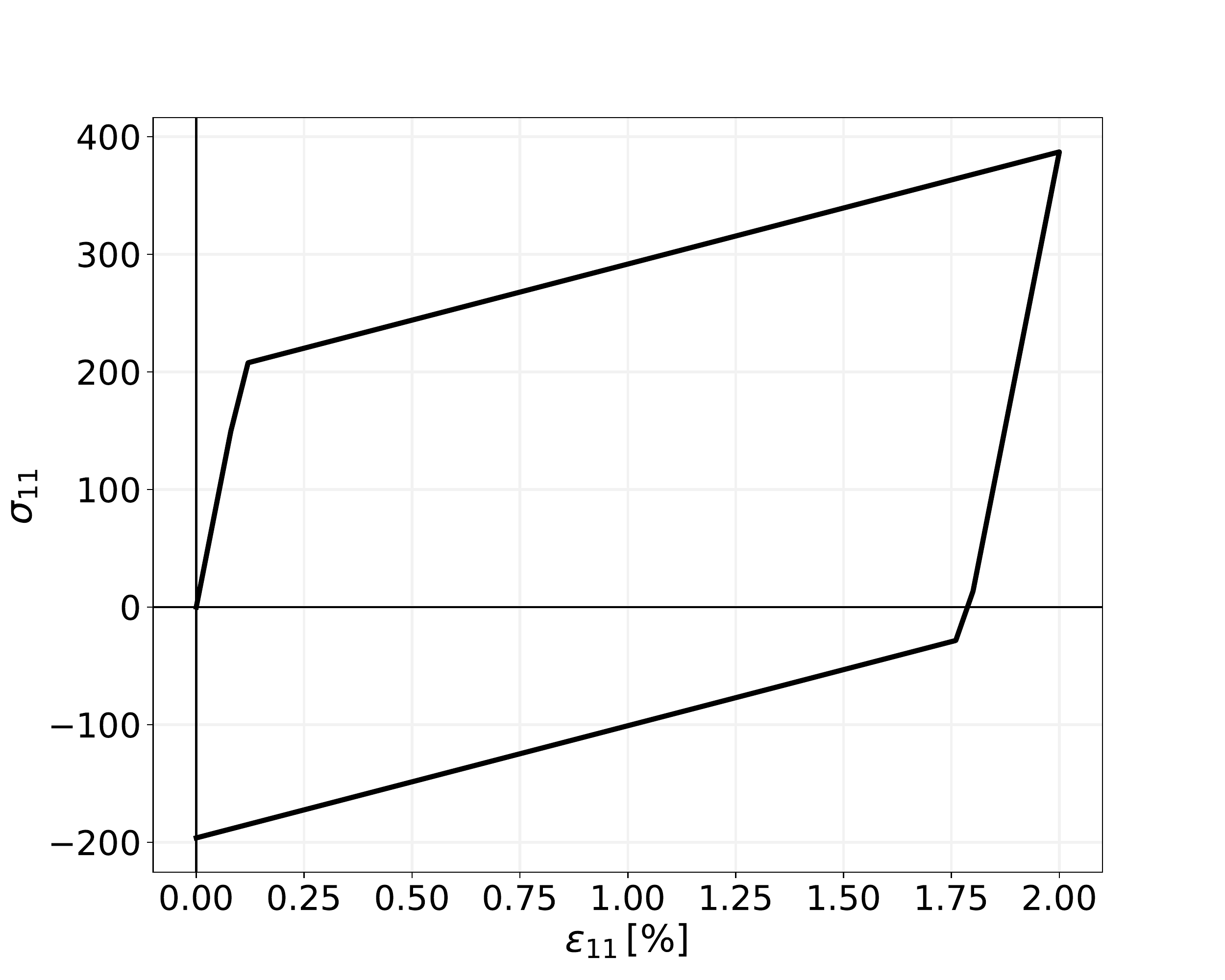}
    \subcaption{Linear kinematic hardening}\label{fig:HowNonKina}
\end{subfigure}
\begin{subfigure}[b]{0.33\linewidth}
        \centering
    \includegraphics[scale=0.22]{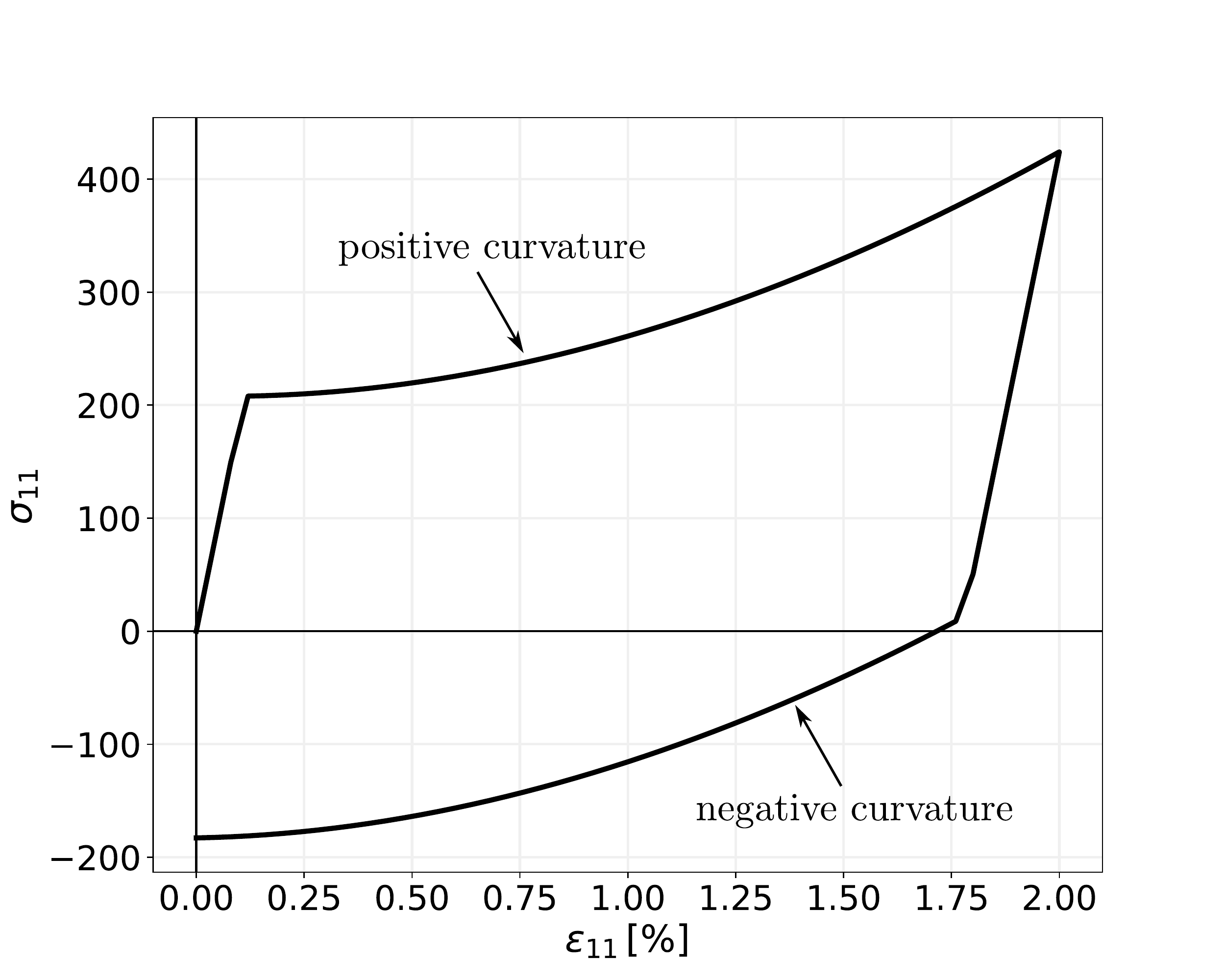}
    \caption{Nonlinear kinematic hardening ($\phi = 0$)}\label{fig:HowNonKinb}
\end{subfigure}
\begin{subfigure}[b]{0.33\linewidth}
        \centering
    \includegraphics[scale=0.22]{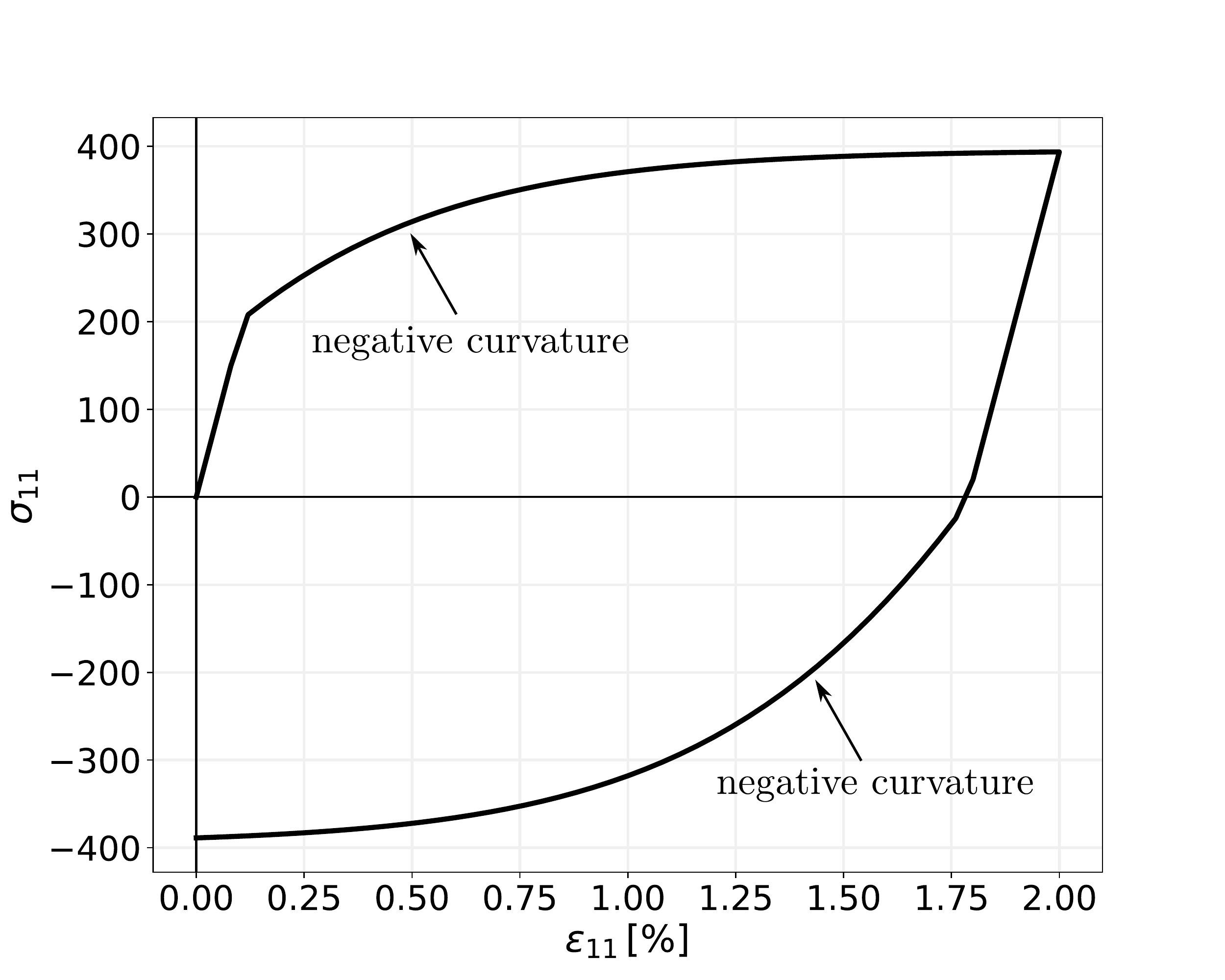}
    \caption{Nonlinear kinematic hardening ($\phi \neq 0$)}\label{fig:HowNonKinc}
\end{subfigure}
    \caption{How to achieve nonlinear kinematic hardening that fits experimental observations. A linear elastic law with $E=212,000$, $\nu=0.26$ and von Mises yield function $\sigma_{0} = 208$ without isotropic hardening is utilized. Uniaxial stress-strain data are plotted for the following cases:
(a) Linear kinematic hardening; $\bm{X} = C\bm{\epsilon}^{p}$ with $C =10000$, 
(b) Nonlinear kinematic hardening law enforced through higher-order nonlinearity in the free energy; $\psi^{p}_{2} = C \trace{(\bm{\alpha}^{3})}$ with $C=300,000$ which leads to
$\dot{\bm{X}} = C \bm{\epsilon}^{p} \dot{\bm{\epsilon}}^{p} $,
(c) Nonlinear kinematic hardening law enforced through nonlinearity in dissipation potential $F$, here Armstrong-Frederick law; $\dot{\bm{X}} = C \dot{\bm{\epsilon}}^{p} + \gamma \bm{X} \dot{\lambda} $ with $C=50000$ and $\gamma=400$.}\label{fig:HowNonKin}

\end{figure}

As discussed in Sections \ref{subsec::ElasticModeling} and \ref{subsec::YieldModeling} a phenomenological model form for the elastic response and the initial yield function have to be preferred in cases where uniaxial experimental data is the only source of information. 
Without loss of generality we opt for a linear elastic isotropic material response (see eq. \eqref{eq:LinElasticity}) and model the initial yield surface using the von Mises yield function defined in eq. \eqref{eq:vonMisesRefor}. But both of these choices can in general be replaced by any other phenomenological or data-driven model.
This leaves two functions, in particular $\psi^{p}$ and $\phi$, still open to be established to fully define the general hardening framework that was discussed in the last section.
\paragraph{Choice of the plastic free energy component} 
For simplicity, we can assume that $\psi^{p}$ can be additively decomposed into two parts
\begin{equation}\label{eq:psipSplit}
    \psi^{p}(r, \bm{\alpha}) = \psi_{1}^{p}(r) + \psi_{2}^{p}(\bm{\alpha}).
\end{equation}
From eqs. \eqref{eq::NewPsi} and \eqref{eq::NewThermoForce} we can see that $R$ now acts as a function of $r$.
Furthermore, this allows us to simplify the expression of the evolution law of eq. \eqref{eq:EvoX1} to 
\begin{equation}\label{eq:EvoX2}
\begin{aligned}
    \dot{\bm{X}} 
&=- \dot{\lambda} \left( \frac{\partial^{2} \psi}{\partial \bm{\alpha} \partial \bm{\alpha}} : \frac{\partial F}{\partial \bm{X}} \right) 
= - \dot{\lambda} \left( \frac{\partial^{2} \psi^{p}_{2}}{\partial \bm{\alpha} \partial \bm{\alpha}} : \left(\frac{\partial f}{\partial \bm{X}}  +  \frac{\partial \phi}{\partial \bm{X}}\right) \right).
\end{aligned}
\end{equation}
Hence, from a data-driven material modeling perspective, it could be assumed from eq. \eqref{eq:psipSplit} that a general elastoplastic material model would rely on ML representations for both $\psi_{1}$ and $\psi_{2}$. However,
in the following, we briefly want to discuss the first term inside the parentheses, $\frac{\partial^{2} \psi^{p}_{2}}{\partial \bm{\alpha} \partial \bm{\alpha}}$, which has received significant attention in the literature over the years
\cite{eisenberg1968nonlinear,chaboche1993cyclic}. For these purposes let us assume that $f$ is a general von Mises yield function. In this case, we can see that 
$\psi^{p}_{2}$ needs to be at least quadratic in $\bm{\alpha}$ for $\dot{\bm{X}} \neq \bm{0}$. If we assume no isotropic hardening and only a quadratic dependency, i.e. $\psi^{p}_{2} = C \bm{\alpha}:\bm{\alpha}$, and $\phi=0$, we obtain a kinematic hardening law that is linear $\dot{\bm{X}} = C \dot{\bm{\epsilon}}^{p}$, see Figure \ref{fig:HowNonKina}. However, a majority of metals appear to have a significant nonlinear response during hardening which can (at least partly) be attributed to nonlinear kinematic hardening behavior \cite{mcdowell1992nonlinear,josefson1995nonlinear,desmorat2010non}.
Therefore, in order to model nonlinear kinematic hardening, we can distinguish two cases
\begin{enumerate}
    \item We can assume a higher than quadratic order dependence of $\psi^{p}_{2}$ on $\bm{\alpha}$ as suggested by \cite{kadashevich1958theory}, for example $\psi^{p}_{2}=C \trace{(\bm{\alpha}^{3})}$. This choice leads to the stress-strain response as seen in
    Figure \ref{fig:HowNonKinb}. We can see that the stress response is characterized by a positive curvature during loading and a negative curvature during unloading. This is a well-recorded phenomenon  \cite{eisenberg1968nonlinear,chaboche1986time,chaboche2008review, chaboche1993cyclic} which leads to a "one-to-one nonlinearity" \cite{lemaitre1994mechanics} which does not fit the observations from experimental cyclic loading cases which display a negative curvature in both the loading and the unloading phase.
    \item In eq. \eqref{eq:EvoX2}, the other option to represent nonlinear kinematic hardening is through the choice of $\phi$. If, for example, we assume a quadratic behavior  $\phi(R,\bm{X})=\frac{\gamma}{C} \bm{X} : \bm{X}$, this describes the commonly used Armstrong-Frederick law \cite{armstrong1966mathematical}. The stress-strain curve of such a response is plotted in Figure \ref{fig:HowNonKinc}. We can see that both the loading and the unloading behavior are characterized by a negative curvature that matches the general signature of experimental observations.
\end{enumerate}
This is the reason why
\cite{desmorat2010non} denotes the general update formula of $\bm{X}$ as 
\begin{equation}\label{eq::generalNKH}
    \dot{\bm{X}} = \frac{2}{3} C \dot{\bm{\epsilon}}^{p} - \bm{\mathcal{B}} (\bm{X},r, \bm{\sigma}) \dot{\mathcal{P}} (\bm{X}, \bm{\sigma}, \dot{\bm{\epsilon}}^{p})
\end{equation}
where the first term is linear in $\dot{\bm{\epsilon}}^{p}$. 
Later, \cite{xiao2012generalized} (c.f. Table 1) identified a non-exhaustive list of around 20 proposed kinematic hardening laws (up to the year 2012) that can be represented by the general formula of eq. \ref{eq::generalNKH} which all exhibit a linear relationship in the first term.
Hence, in the following we choose the plastic component of the free energy function as
\begin{equation}
    \psi^{p}(r,\bm{\alpha}) = \psi_{1}^{p}(r) + C \bm{\alpha}:\bm{\alpha} 
\end{equation}
which leads to the trainable parts of the model being the  material parameter $C$ and also the function $ \psi_{1}^{p}$ (through its derivative $R = \frac{\partial \psi_{1}^{p}}{\partial r}$). Additionally, the function $\phi$ remains to be determined, which as discussed, will facilitate the necessary nonlinearity of the response.

\paragraph{Choice of the nonlinear kinematic hardening function}
As pointed out, the form of $\phi$ plays a significant role in modeling nonlinear hardening behavior. 
The arguments of $\phi(R,\bm{X})$ are the two thermodynamic hardening forces $R$ and $\bm{X}$.
Since we treat $\phi$ as a means of introducing nonlinear kinematic hardening behavior we argue that we can assume it is independent of $R$, which is the thermodynamic force we associate with isotropic hardening.
Hence, we can reduce $\phi(R, \bm{X})$ to $\phi = \phi(\bm{X})$. However, to train this representation sufficiently well we would need to have access to data of the whole span defined by the invariants of $\bm{X}$ which are not accessible when only uniaxial tests are available (even in the case of uniaxial cyclic tests). Therefore, in order to obtain a representation that is reliably trainable on this limited amount of data we reduce the dependence of $\phi$ to the squared Frobenius norm of $\bm{X}$. 
We, therefore, arrive at the following expression for the plastic potential
\begin{equation}
    F (\bm{\sigma}, R, \bm{X})   =  f(R(r) \, \overline{\pi}_{1}, R(r) \, \overline{\pi}_{2})  + \phi(\norm{\bm{X}}_{F}^{2})
\end{equation}
where $\norm{\bm{X}}_{F}^{2} = \bm{X}:\bm{X}$. We highlight again that this choice is only made because we are concerned with uniaxial experimental datasets. For data from other sources, e.g. numerical homogenization, a function involving the full dependency $\phi(R, \bm{X})$ might be trained. We will explore this in future work which will revolve around data from microstructural simulations.

\paragraph{Constraints}\label{sub:ParaConstraints}
To enforce thermodynamic consistency for the dissipation potential $F$, the yield function $f$ and the nonlinear kinematic hardening function $\phi$ have to comply with the following constraints 
\begin{itemize}
    \item $f$ has to be convex and $f(R \overline{\pi}_{1}=0, R \overline{\pi}_{2}=0)< 0$,
    \item $\phi$ has to be convex, $\phi(\norm{\bm{X}}_{F}^{2}=0)= 0$ and $\phi( \norm{\bm{X}}_{F}^{2})\geq 0$.
\end{itemize}
In the following, we remind the reader that a suitable data-driven or phenomenological model for $f$ has already been selected as discussed in Section \ref{subsec::YieldModeling}. 
It is crucial for the machine learning tools to also be consistent with the thermodynamic constraints.
To achieve this, we employ two neural networks $\mathcal{N}_{\phi}$ and $\mathcal{N}_{R}$ to approximate the responses of $\phi$ and $R$.
However, these neural networks can be replaced with other tools such as Gaussian process or support vector regressors. We choose neural networks since they allow for simple enforcement of the thermodynamic constraints as will be shown in the following. 
To remain consistent with the thermodynamic conditions, we need to establish a neural network $\mathcal{N}_{\phi}$ that satisfies the following constraints 
$\mathcal{N}_{\phi}$:
\begin{itemize}
    \item $\mathcal{N}_{\phi}(\norm{\bm{X}}_{F}^{2})\geq 0$,
    \item $\mathcal{N}_{\phi}(\norm{\bm{X}}_{F}^{2})$ has to be monotonically increasing and convex with regards to its input $\norm{\bm{X}}_{F}^{2}$.
\end{itemize}
We can furthermore assume consistent isotropic hardening behavior without softening. There, $\mathcal{N}_{R}$ is subject to the constraints:
\begin{itemize}
    \item $\mathcal{N}_{R}(r)\geq 0$,
    \item $\mathcal{N}_{R}(r)$ has to be monotonically decreasing with regards to its input $r$.
\end{itemize}
Knowing that the inputs to both neural networks are always positive, since $r\geq 0$ and $\norm{\bm{X}}_{F}^{2}\geq 0$, we can define three different neural network types:
\paragraph{Positive neural networks}
Let $\mathcal{N}: \mathbb{R} \rightarrow \mathbb{R}$ be a feedforward neural network with $L$ hidden layers.
The updating formula of the neural networks reads
\begin{equation}
    \begin{aligned}
            x_{0} &\in \mathbb{R}_{\geq 0} \\
            \bm{x}_{1} = \sigma_{1} \left( x_{0} \bm{W}_{1}^{T} + \bm{b}_{1} \right) &\in \mathbb{R}^{n^{1}} \\
            \bm{x}_{l} = \sigma_{l} \left( \bm{x}_{l-1} \bm{W}_{l}^{T} + \bm{b}_{l} \right) &\in \mathbb{R}^{n^{l}}, \qquad l=1, \ldots, L-1 \\
            x_{L} = \bm{x}_{L-1} \bm{W}_{L}^{T} + \bm{b}_{L}, &\in \mathbb{R}
    \end{aligned}
\end{equation}

\begin{corollary}
The output of a neural network is positive if $x_{0}\geq 0$, $W_{l}, b_{l} \geq 0$ and $\sigma_{l}: \mathbb{R}_{\geq 0} \rightarrow \mathbb{R}_{\geq 0}$ for $l=1, \ldots, L$ then $x_{L} \geq 0$. We call this network positive.
\end{corollary}
\begin{proof}
If the input to the $l$-th layer is elementwise positive, i.e. $\bm{x}_{l-1}\geq 0$ and we can assume $W_{l}, b_{l} \geq 0$, then $ \bm{x}_{l-1} \bm{W}_{l}^{T} + \bm{b}_{l} \geq 0$. Since, the activation function $\sigma_{l}$ is applied per element the output of the $l$-th layer $\bm{x}_{l} $ is also greater or equal zero. Therefore, since the initial input is positive, the output of the network will also be positive.\\
We can prove this more rigorously by induction: For $l=1$
\begin{equation}
     \bm{x}_{1} = \sigma_{1} \left( x_{0} \bm{W}_{1}^{T} + \bm{b}_{1} \right) \geq 0, \text{ since } \bm{W}_{1},\bm{b}_{1} \geq 0, \text{ and } \sigma_{1}: \mathbb{R}_{\geq 0} \rightarrow \mathbb{R}_{\geq 0}.
\end{equation}
Now assume $\bm{x}_{l-1} \geq 0$, then
\begin{equation}
\bm{x}_{l} = \sigma_{l} \left( \bm{x}_{l-1} \bm{W}_{l}^{T} + \bm{b}_{l} \right) , \text{ since } \bm{W}_{l},\bm{b}_{l} \geq 0, \text{ and } \sigma_{l}: \mathbb{R}_{\geq 0} \rightarrow \mathbb{R}_{\geq 0}.
\end{equation}
\end{proof}
\paragraph{Positive, monotonically increasing neural networks}
A function $f: [a,b] \rightarrow \mathbb{R}$ is called monotonically increasing if $f(x) \leq f(y)$, $\forall x,y \in [a,b]$ where $x\leq y$. Equivalently a function is monotonically increasing if $f'(x) \geq 0$ for all $x \in [a,b]$.
\begin{corollary}
A positive neural network is monotonically increasing if $\sigma'_{l} : \mathbb{R}_{\geq 0} \rightarrow \mathbb{R}_{\geq 0}$. We call this network positive and monotonically increasing.
\end{corollary}
\begin{proof}
The derivative of the network output $x_{L}$ with regards to the input $x_{0}$ is given by (c.f. \cite{ratku2022derivatives})
\begin{equation}
    \frac{d x_{L}}{d x_{0}} = \prod_{l=0}^{L-1} \lbrace \left[ (\sigma_{L-l}^{'}( \underbrace{\bm{x}_{L-l-1} \bm{W}_{L-l}^{T} + \bm{b}_{L-l}}_{\bm{y}_{L-l}}))^{T} \bm{j}_{L-l} \right] \circ \bm{W}_{L-l} \rbrace
\end{equation}
where $\bm{j}_{l}$ are row vectors of ones with same size as $\bm{x}_{l-1}$. For a positive neural network we know $x_{0} \geq 0$, $\bm{W}_{l}, \bm{b}_{l} \geq 0$ and 
$\sigma_{l}: \mathbb{R}_{\geq 0} \rightarrow \mathbb{R}_{\geq 0}$. Hence, we established that $\bm{y}_{L-l} \geq 0$. Therefore, if $\sigma'_{L-l} : \mathbb{R}_{\geq 0} \rightarrow \mathbb{R}_{\geq 0}$ we see that
\begin{equation}
    \frac{d x_{L}}{d x_{0}} \geq 0, \qquad \forall x_{0} \geq 0
\end{equation}
which is the definition for a positive, monotonically increasing function on $\mathbb{R}_{+}$.
\end{proof}
We remark that we can obtain a positive, monotonically decreasing neural network by taking the reciprocal of the output of a positive, monotonically increasing neural network.
\paragraph{Positive, monotonically increasing, input convex neural networks}
A function $f: [a,b] \rightarrow \mathbb{R}$ is convex if $f''(x) \geq 0$ for all $x \in [a,b]$.
\begin{corollary}
A positive, monotonically increasing neural network is convex with regards to its input if $\sigma_{l} : \mathbb{R}_{\geq 0} \rightarrow \mathbb{R}_{\geq 0}$ is convex. We call this network positive, monotonically increasing, and input convex.
\end{corollary}
\begin{proof}
The second derivative of the network output is given by
(c.f. \cite{ratku2022derivatives})
\begin{equation}
    \frac{d^{2} x_{L}}{d^{2} x_{0}} = \sum_{l=1}^{L} \bm{J}N^{l+1,L} \left(   \left[ \lbrace (\sigma_{l}^{''})^{T} \circ \bm{m} \rbrace  \bm{j}_{l} \right] \circ \bm{W}_{l} \right) \bm{J}N^{1,l-1}
\end{equation}
where 
\begin{equation}
    \begin{aligned}
    \bm{J}N^{p,q} &=  \prod_{k=L-q}^{L-p} \lbrace \left[ (\sigma_{L-k}^{'}( \bm{x}_{L-k-1} \bm{W}_{L-k}^{T} + \bm{b}_{L-k}))^{T} \bm{j}_{L-k} \right] \circ \bm{W}_{L-k} \rbrace,  \\
        \sigma_{l}^{''} &= \sigma_{l}^{''}( \bm{x}_{l-1} \bm{W}_{l}^{T} + \bm{b}_{l}) , \\
        \bm{m} &= \bm{W}_{l} \bm{J}N^{1,l-1}.
    \end{aligned}
\end{equation}
Hence, since for a positive, monotonically increasing neural networks we know that $x_{0} \geq 0$, $\bm{W}_{l}, \bm{b}_{l} \geq 0$, 
$\sigma_{l}: \mathbb{R}_{\geq 0} \rightarrow \mathbb{R}_{\geq 0}$ and $\sigma_{l}': \mathbb{R}_{\geq 0} \rightarrow \mathbb{R}_{\geq 0}$ we can see that 
\begin{equation}
     \frac{d^{2} x_{L}}{d^{2} x_{0}} \geq 0, \qquad \forall x_{0} \geq 0,
\end{equation}
if $\sigma_{l}'': \mathbb{R}_{\geq 0} \rightarrow \mathbb{R}_{\geq 0}$ which is equivalent to saying that $\sigma_{l}$ is convex $\forall x_{0} \geq 0$.
\end{proof}
Finally, in order to conform with the established constraints, we choose $\mathcal{N}_{\phi}$ to be a positive, monotonically increasing, convex neural network and $\mathcal{N}_{R}$ to be the reciprocal of a positive, monotonically increasing neural network.

In addition to the suggested constraints (i.e. convexity, monotonicity) the two trainable functions also have to adhere to the following initial value constraints
\begin{equation}
\begin{aligned}
R(r=0) = 1, \qquad
\phi(0) = 0.
\end{aligned}  
\end{equation}
These convey that there should be no hardening for the material model prior to yielding. Since neural networks are based on average loss definitions, the output of the networks can not be guaranteed to fit these initial conditions.
However, we can implicitly comply with the constraint by perturbing the neural network output with a correction term. Following \cite{as2022mechanics, huang2022variational} where scaling terms have also recently been used to constrain the output of neural networks we can define the final model output of the isotropic hardening force with
\begin{equation}
    R(r) = \mathcal{N}_{R}(r) - \mathcal{N}_{R}(0) + 1.0 
\end{equation}
as well as the output of $\phi$ by
\begin{equation}
    \phi(\norm{\bm{X}}_{F}^{2})= \mathcal{N}_{\phi}(\norm{\bm{X}}_{F}^{2}) - \mathcal{N}_{\phi}(0).
\end{equation}
In the following, we briefly discuss details concerning the incremental formulation as well as the training algorithm employed in this work.

\subsection{Implementation highlights}
We assume access to a model $\hat{\bm{\sigma}}$ that yields the elastic response as well as a model for the initial yield function $f$. 
In this work we rely on the well-established implicit return mapping algorithm \cite{simo2006computational,wriggers2008nonlinear,de2011computational}. Given the increment of strain $\Delta \bm{\epsilon} = \bm{\epsilon}_{n+1} - \bm{\epsilon}_{n}$ corresponding to the time increment $[t_{n}, t_{n+1}]$ let the elastic trial strain be defined by
$\bm{\epsilon}_{n+1}^{e \, trial} = \bm{\epsilon}^{e}_{n} + \Delta \bm{\epsilon}$. 
The trial stress is then given by 
\begin{equation}
    \bm{\sigma}_{n+1}^{trial} = \hat{\bm{\sigma}} (\bm{\epsilon}_{n+1}^{e \, trial}).
\end{equation}
If the resulting relative stress lies inside the yield surface $f< 0$, then this time increment is purely elastic and we can update the variables by
\begin{equation}
\begin{aligned}
\bm{\epsilon}^{e}_{n+1} &= \bm{\epsilon}_{n+1}^{e \, trial}, \\
        \bm{X}_{n+1} &= \bm{X}_{n},\\
        r_{n+1} &= r_{n},\\
R_{n+1} &= R_{n}. \\
\end{aligned}
\end{equation}
Otherwise we need to solve the following system of nonlinear equations  $\bm{\epsilon}^{e}_{n+1}$, $\bm{X}_{n+1}$, $r_{n+1}$, $R_{n+1}$ and $\Delta \lambda$ 
\begin{equation}
\begin{aligned}
\bm{\epsilon}^{e}_{n+1} &= \bm{\epsilon}_{n+1}^{e \, trial} + \Delta \lambda \frac{\partial F}{\partial \bm{\sigma}_{n+1}}, \\
        \bm{X}_{n+1} &= \bm{X}_{n} -  \Delta \lambda 2 C \left( \frac{\partial f}{\partial \bm{X}_{n+1}} + \frac{\partial \phi}{\partial \bm{X}_{n+1}}  \right),\\
        r_{n+1} &= r_{n} - \Delta \lambda \frac{\partial f}{\partial R_{n+1}} ,\\
R_{n+1} &= R_{n} - \Delta \lambda \frac{\partial^{2} \psi_{1}^{p}}{\partial r_{n+1}^{2}} \frac{\partial f}{\partial R_{n+1}}, \\
    f(\bm{\epsilon}^{e}_{n+1},  \bm{X}_{n+1}) &= 0.
\end{aligned}
\end{equation}
In this work we employ a standard Newton-Raphson optimizer \cite{ypma1995historical} which gives the iterative update
\begin{equation}\label{eq::NewtonRaphson}
    \bm{x}^{v+1} = \bm{x}^{v} - [\bm{J}^{v}]^{-1} \bm{F}(\bm{x}^{v}).
\end{equation}
The explicit expressions for each term are summarized in Section \ref{subsec:NewtonRaphsonVal}. In this work, we obtain the Jacobian $\bm{J}$ using automatic differentiation. 
After one iteration step is completed, the stress prediction can be updated using
\begin{equation}
    \bm{\sigma}_{n+1} = \hat{\bm{\sigma}} (\bm{\epsilon}^{e}_{n+1})
\end{equation}
The consistent elastoplastic tangent operator which is relevant for finite element computations can then be obtained 
\begin{equation}
    \mathbb{C}^{ep} = \frac{\partial \bm{\sigma}_{n+1}}{\partial \bm{\epsilon}_{n+1}^{e \, trial}}.
\end{equation}

Using this implicit return-mapping framework and given a dataset of uniaxial tension/compression data we can train all necessary representations required for the discussed framework.
The training algorithm which is based on a stochastic gradient descent scheme is summarized in Algorithm \ref{alg:Train}. It also includes steps to enforce the uniaxiality constraint of the stress response ($\sigma_{ij}=0 \text{ for } (i,j) \neq (1,1) $).
The constraints on the weights and biases of the neural networks models as well as the material parameters are enforced using gradient clipping \cite{zhang2019gradient}.
We remark that the networks are initialized using the Kaiming uniform initialization technique \cite{he2015delving} which is adjusted to normalized and scaled data. Since the models in this network are trained using an online-training approach where the training data is only implicitly available and can not a-priori be scaled we undertake two steps to allow for faster training of the model even on the unscaled data
\begin{itemize}
    \item We scale the input to the neural network $\mathcal{N}_{R}$ by a factor of $100$ before using it as an input. This is done for the following reason: We expect the value of $r$ to be around the order of the accumulated plastic strain which (since $\epsilon \approx 0.00-0.05$) puts $100 r$ closer to a value around $1$ after the onset of plasticity. For cases where this rough approximation is not possible this input factor might also be added as an additional trainable parameter. Alternatively, other network architectures could be used that were proposed to be self-normalizing, see e.g. \cite{shao2020normalization}. 
    \item Additionally, to make the networks more resistant against ill-scaled input and output values we utilize "scalable" or parameterized activation functions which have recently received more attention \cite{gnanasambandam2022self,jagtap2022important}.
Hence, instead of the standard set of activation functions, we use a parameterized version of the Softplus activation function for $\mathcal{N}_{\phi}$
\begin{equation}
    \sigma^{\mathcal{N}_{\phi}}_{l}(x) = \frac{1}{\beta} \log(1+\beta \exp{x})
\end{equation}
with $\beta>0$ being a different trainable parameter for each hidden layer of the model.
For $\mathcal{N}_{R}$ we employ a "scalable" version of the logistic function given by
\begin{equation} 
   \sigma^{\mathcal{N}_{R}}_{l}(x) =  \frac{1}{1+\exp{ - \beta_{1} (x-\beta_{2})}}
\end{equation}
where $\beta_{1}, \beta_{2}>0$ are additional trainable parameters for every hidden layer of the model.
It can be seen that both of these activation functions fulfill the conditions that allow for $\hat{\mathcal{N}}_{\phi}$ to be positive and convex and for $\hat{\mathcal{N}}_{R}$ to be positive and monotonically decreasing. To give the reader an impression of these activation functions, both are plotted for different values in Figure \ref{fig:ScaleAct}.

\begin{figure}
\begin{subfigure}[b]{0.45\linewidth}
        \centering
    \includegraphics[scale=0.3]{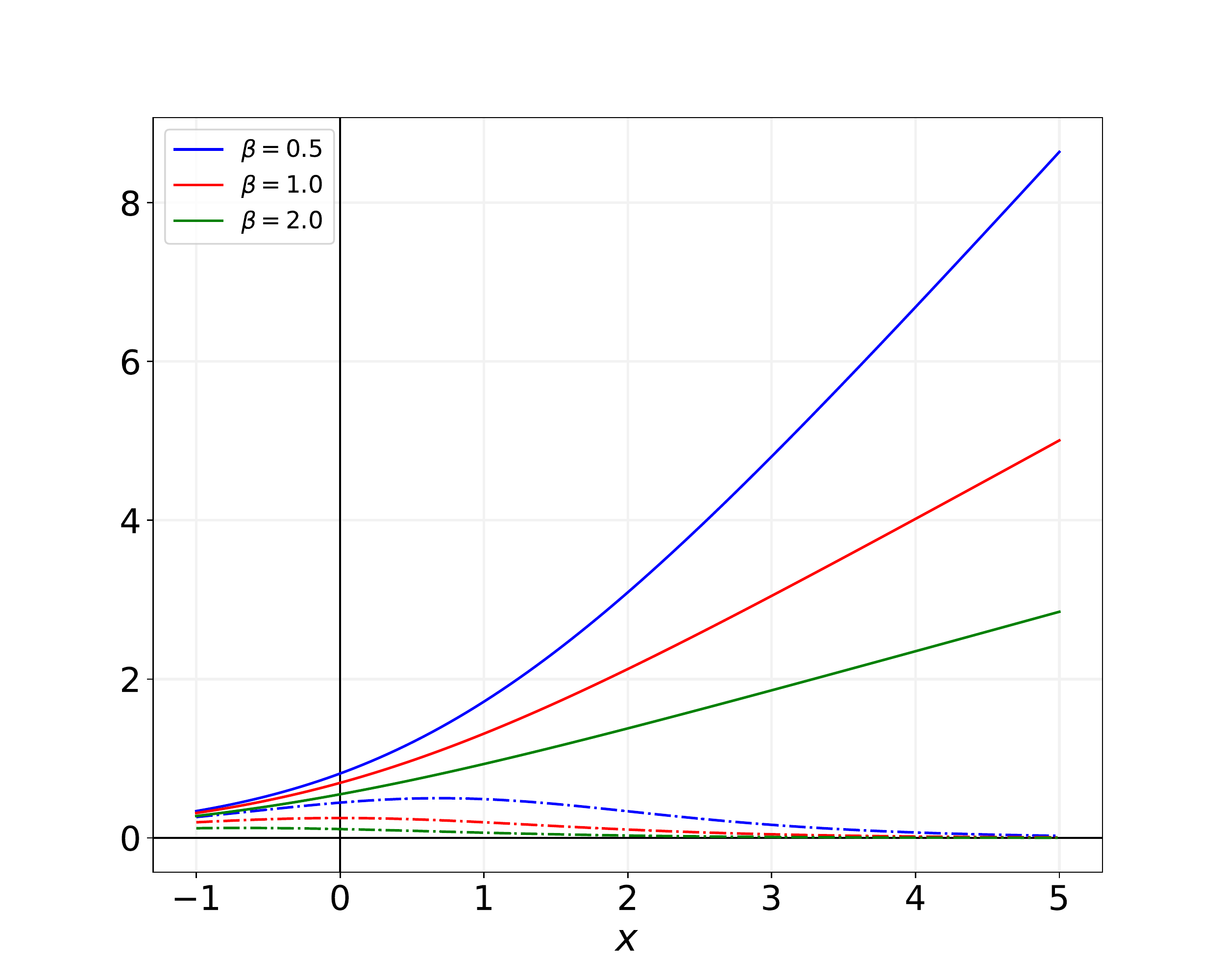}
    \caption{}
\end{subfigure}
\begin{subfigure}[b]{0.45\linewidth}
        \centering
    \includegraphics[scale=0.3]{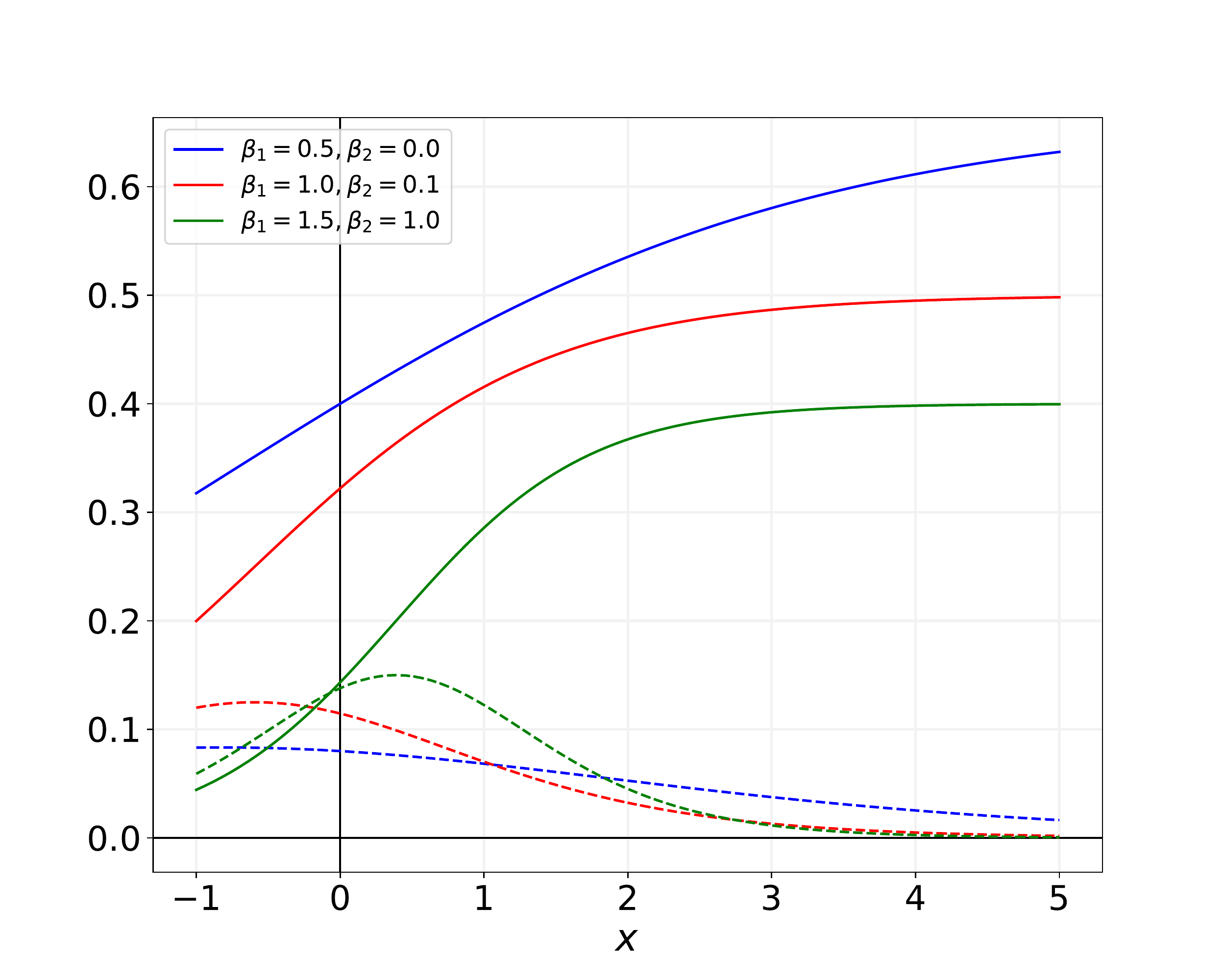}
    \caption{ }
\end{subfigure}

\caption{Example curves of scalable activation functions. (a) Parameterized version of the Softplus activation function used in $\mathcal{N}_{\phi}$ for different parameter values. The dotted lines represent the second derivative of the function which is important for the convexity constraint of the network, (b) Parameterized version of the Logistic regression activation function for different parameter values used in the positive, monotonically decreasing network $\mathcal{N}_{R}$. The dotted lines represent the first derivative of the respective curves.  }\label{fig:ScaleAct}
\end{figure}

\end{itemize}

\begin{algorithm}
\caption{Training algorithm for given uniaxial dataset}\label{alg:Train}
\begin{algorithmic}
\Require {Dataset of $N$ samples consisting of uniaxial stress-strain data (assume here $\epsilon_{11}$ and $\sigma_{11}$), i.e. $\mathcal{U} = \lbrace \epsilon_{11}^{i}, \sigma_{11}^{i}\rbrace_{i=1}^{N}$. Information about elastic mapping ($\hat{\sigma}: \mathbb{R}^{3\times 3} \rightarrow \mathbb{R}^{3\times 3} $) and yield function mapping $\hat{f}$.} \\
\Establish 
\begin{itemize}
    \item Initialize positive, monotonically decreasing network $R$ with trainable parameters $\bm{\theta}_{R}$ $\rightarrow$ optimizer $\mathcal{O}_{R}$
    \item Initialize positive, monotonically increasing, input convex network $\phi$ with trainable parameters $\bm{\theta}_{\phi}$ $\rightarrow$ optimizer $\mathcal{O}_{\phi}$
    \item Initialize other trainable material parameters $\bm{\theta}_{M}$ $\rightarrow$ optimizer $\mathcal{O}_{M}$
    \item Loading path with $N_{L}$ steps, Training steps $N_{T}$
\end{itemize}
\Begin \\
\textbf{Set:} $R_{0} \leftarrow 1$, $r_{0} \leftarrow 0$, $\bm{\epsilon}_{0}^{e} \leftarrow \bm{0}$, $\bm{X}_{0} \leftarrow \bm{0}$,  $\bm{\epsilon}_{0}^{p} \leftarrow \bm{0}$
\For {$i \leftarrow 1, N_{T}$}

$\mathcal{L}(\bm{\theta}_{R},\bm{\theta}_{\phi}, \bm{\theta}_{M}) \leftarrow 0$

\For {$n \leftarrow 0, N_{L}-1$}

 \quad  Get strain increment $\Delta \bm{\epsilon}$ from loading path

 \quad  $\bm{\epsilon}_{n+1} \leftarrow \bm{\epsilon}_{n} + \Delta \bm{\epsilon} $

 \quad  $R_{n+1}^{trial} \leftarrow R_{n}$, $r_{n+1}^{trial} \leftarrow r_{n}$, $\bm{X}_{n+1}^{trial}  \leftarrow \bm{X}_{n}$

 \quad  \textbf{Solve} for $\bm{\epsilon}_{n+1}^{e \, trial}$:  \Comment{Use $\bm{\epsilon}_{n}^{e}$ as starting guess}
\begin{fleqn}[\dimexpr(\leftmargini-\labelsep)*2]
          \setlength\belowdisplayskip{0pt}
 \begin{equation*}
                \begin{aligned}[c]
                \bm{\epsilon}_{n+1}^{e \, trial} - (\bm{\epsilon}_{n+1} - \bm{\epsilon}_{n}^{p}) &= \bm{0}\\
\hat{\sigma} (\bm{\epsilon}_{n+1}^{e \, trial})_{ij}  &= 0 \, \qquad \text{for } (i,j) \neq (1,1)
              \end{aligned}
          \end{equation*}
          \end{fleqn}\\

\If{$\hat{f}(\bm{\sigma}_{n+1}^{trial}, R_{n+1}^{trial},r_{n+1}^{trial}, \bm{X}_{n+1}^{trial};\bm{\theta}_{R},\bm{\theta}_{\phi}, \bm{\theta}_{M}) \leq 0$}

\qquad $ \bm{\epsilon}_{n+1}^{e} \leftarrow \bm{\epsilon}_{n+1}^{e \, trial}$, $R_{n+1} \leftarrow R_{n+1}^{trial}$, $r_{n+1} \leftarrow r_{n+1}^{trial}$, $\bm{X}_{n+1} \leftarrow \bm{X}_{n+1}^{trial}$, $ \bm{\epsilon}_{n+1}^{p} \leftarrow \bm{\epsilon}_{n+1} - \bm{\epsilon}_{n+1}^{e \, trial}$

\Else{}

     \qquad    \textbf{Solve} for $\bm{\epsilon}^{e}_{n+1}, \bm{X}_{n+1}, r_{n+1}, R_{n+1}, \Delta \lambda$:  \Comment{Use last values $\bullet_{n}$ as starting guess}
          \begin{fleqn}[\dimexpr(\leftmargini-\labelsep)*2]
          \setlength\belowdisplayskip{0pt}
 \begin{equation*}
                \begin{aligned}[c]
\bm{\epsilon}^{e}_{n+1} - \left( \bm{\epsilon}_{n+1}^{e \, trial} + \Delta \lambda \frac{\partial F}{\partial \bm{\sigma}_{n+1}} \right)&= 0 \\
        \bm{X}_{n+1} - \left[ \bm{X}_{n} -  \Delta \lambda  2 C \left( \frac{\partial f}{\partial \bm{X}_{n+1}} +  \frac{\partial \phi}{\partial \bm{X}_{n+1}}  \right) \right] & = 0\\
        r_{n+1} - \left( r_{n} - \Delta \lambda \frac{\partial f}{\partial R_{n+1}} \right) &= 0 \\
R_{n+1} -\left( R_{n} - \Delta \lambda \frac{\partial^{2} \psi_{1}^{p}}{\partial r_{n+1}^{2}} \frac{\partial f}{\partial R_{n+1}} \right) &=0 \\
    f(\bm{\epsilon}^{e}_{n+1},  \bm{X}_{n+1}) &= 0 \\
    \hat{\sigma} (\bm{\epsilon}^{e}_{n+1})_{ij}  &= 0 \, \qquad \text{for } (i,j) \neq (1,1)
              \end{aligned}
          \end{equation*}
          \end{fleqn}

\EndIf 

Get stress $\bm{\sigma}_{n+1} \leftarrow \hat{\sigma}(\bm{\epsilon}^{e}_{n+1})$

Find $\overline{\sigma}_{11}$ from $\mathcal{U}$: Interpolating between $\bm{\epsilon}_{n+1}$ and $\mathcal{U}$

$\mathcal{L} \leftarrow \mathcal{L} + \norm{ \sigma_{n+1, 11} - \overline{\sigma}_{11}}^{2}$

\EndFor 

Get gradients: $\bm{g}_{R} \leftarrow \nabla_{\bm{\theta}_{R}} \mathcal{L}, \bm{g}_{\phi} \leftarrow \nabla_{\bm{\theta}_{\phi}} \mathcal{L}, \bm{g}_{M} \leftarrow \nabla_{\bm{\theta}_{M}} \mathcal{L}$ 

Update parameters of optimizers $\mathcal{O}_{R}, \mathcal{O}_{\phi}, \mathcal{O}_{M}$ and use gradient clipping to ensure constraints are met

\EndFor
\End
\end{algorithmic}
\end{algorithm}

Now that the framework is fully established we can test its performance.

\section{Applications}\label{sec::4}
The training algorithm (Algorithm \ref{alg:Train}) is implemented\footnote{Codes relating to this work are available from the corresponding author under reasonable request.} in
Pytorch \cite{NEURIPS2019_9015} to take advantage of the automatic differentiation capabilities of the framework. The parameters are optimized using the AdamW optimizer \cite{loshchilov2017decoupled} where a learning rate of $1e-2$ is used for the neural network parameters while standalone parameters (material parameters) are trained with a learning rate of $5e-2$. We did not use a scheduler to adjust the learning rate.
Since we expect smooth ground truth curves for $\phi$ and $R$, we choose a neural network architecture consisting of 1 hidden layer with 10 neurons for both models.
We, therefore, end up with 268 combined trainable parameters which include the material parameter $C$. This amount is significantly higher than used in traditional phenomenological modeling where 10 trainable parameters are already considered a high number. 

In the following, we train until the loss plateaus (i.e. loss stops changing significantly) and then take the best-performing model in terms of training loss over the course of training as the final model for testing.
For simplicity, we assume that all material parameters as well as the experimental datasets are dimensionless. The linear response for all models is assumed to be isotropic linear elastic and the initial yield function can be described using the classical von Mises model.
\subsection{Fitting to existing constitutive laws}
In the first part of this section, we investigate if the proposed framework is able to accurately capture the hardening behavior of known phenomenological nonlinear kinematic hardening (NLK) laws.
Here, we differentiate between single and multi NLK models where in the multi NLK model the backstress is a superposition of multiple backstresses while the single NLK model has a single backstress evolution law.
\subsubsection{Single NLK-model fit}
\begin{figure}
    \centering
    \includegraphics[scale=0.3]{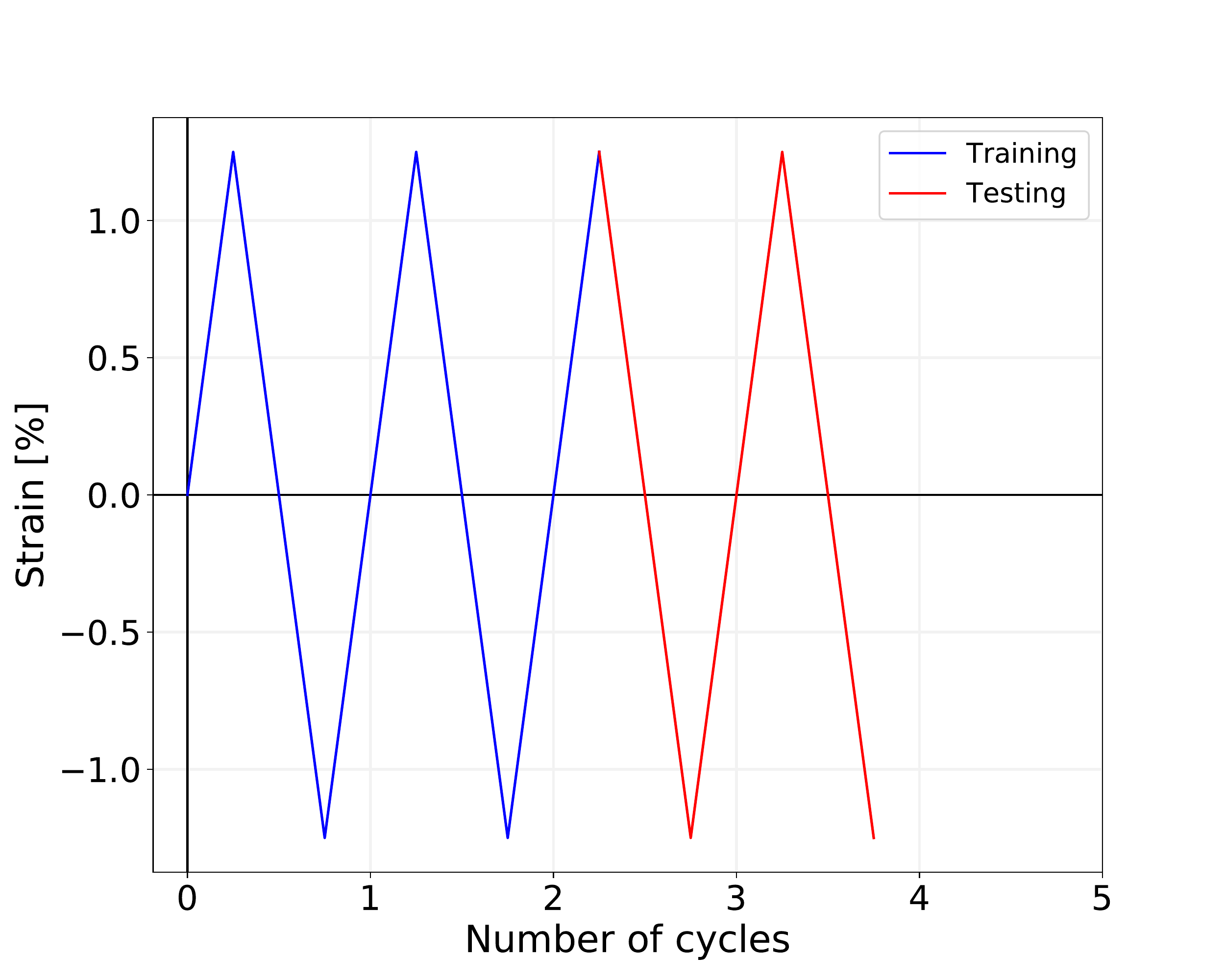}
    \caption{Loading curve of the single NLK model}
    \label{fig:singleNLKLoading}
\end{figure}

Consider a classical von Mises yield function of the form
\begin{equation}
f = J(\bm{\sigma} - \bm{X}) - \sigma_{y} - R \leq 0    
\end{equation}
where $J(\bullet)$ is the von Mises equivalent stress and $\sigma_{y}$ denotes the yield stress.
We assume a commonly applied isotropic hardening law of Voce-type, see e.g. \cite{flaschel2022discovering}, given by
\begin{equation}\label{eq:singleNLKVoce}
    R(r) = H_{1} r+ H_{2} (1 - \exp (- H_{3} r))
\end{equation}
where $H_{1}$, $H_{2}$ and $H_{3}$ are material parameters.
Additionally, a nonlinear kinematic hardening formulation is assumed
\begin{equation}\label{eq:singleNLKX}
    \dot{\bm{X}} = \frac{2}{3} C \dot{\bm{\epsilon}}^{p} - \gamma \text{tr}(\bm{X}^{2})^{m} \bm{X} \dot{r}
\end{equation}
which is similar to the functional form suggested in \cite{ohno1994kinematic}.
This means the plastic potential reads
\begin{equation}
    F = f + \frac{\gamma}{2 C (m+1)} \left( \text{tr}(\bm{X}^{2})^{m+1}   \right).
\end{equation}
All material parameters of this model are specified in Table \ref{tab::SingleNLKParam}.
Figure \ref{fig:singleNLKLoading} plots the loading paths that are used for the training and testing of the following models. The specimen is loaded three times until 1.25 $\%$ strain and the load is reversed two times. This data is used as the training data (blue line). For testing the specimen the loading-reverse-loading process is continued for two reverse-loading phases (red line).

\begin{table}
\begin{center}
\begin{tabular}{||c c c c c c c c c||} 
 \hline
$E$ & $\nu$ &$C$ & $\gamma$ & $m$ & $H_{1}$ & $H_{2}$ & $H_{3}$ & $\sigma_{y}$ \\ [0.5ex] 
 \hline\hline
200e3 & 0.3 & 15 & 550 & 0.9 & 0.1875 & 0.25 & 2.0 & 207\\ [1ex] 
 \hline
\end{tabular}
\end{center}
\caption{Material parameters of the single NLK model.}\label{tab::SingleNLKParamOpti}
\end{table}

\paragraph{Single NLK - Only isotropic hardening component}
In a first experiment we aim to to check whether our framework is able to recreate a model that is just subject to isotropic hardening. Hence, for the following results assume $C=\gamma=0$ in Table \ref{tab::SingleNLKParam}.

\begin{figure}
\begin{subfigure}[b]{0.45\linewidth}
        \centering
    \includegraphics[scale=0.3]{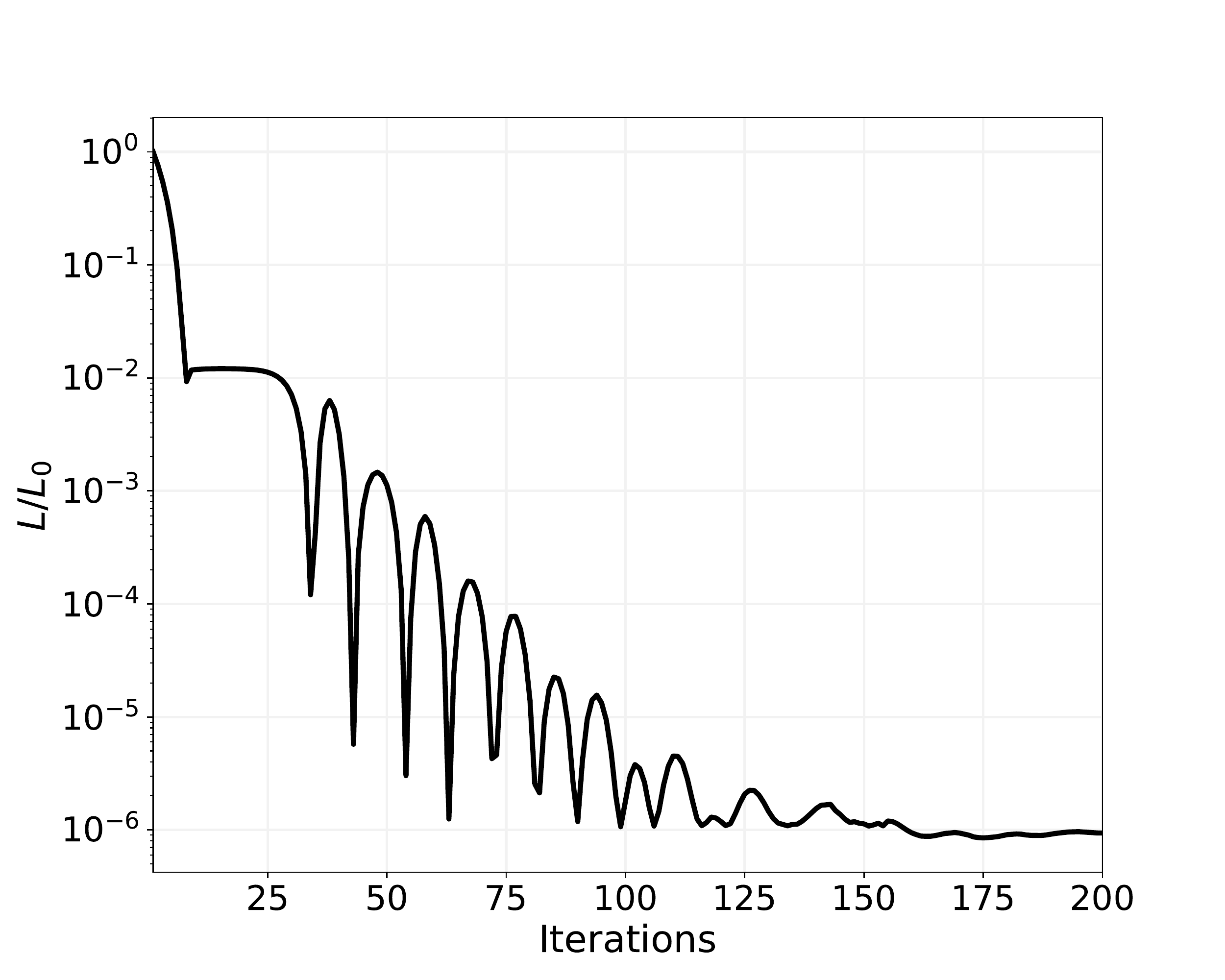}
    \caption{}\label{fig:isoOnlya}
\end{subfigure}
\begin{subfigure}[b]{0.45\linewidth}
        \centering
    \includegraphics[scale=0.3]{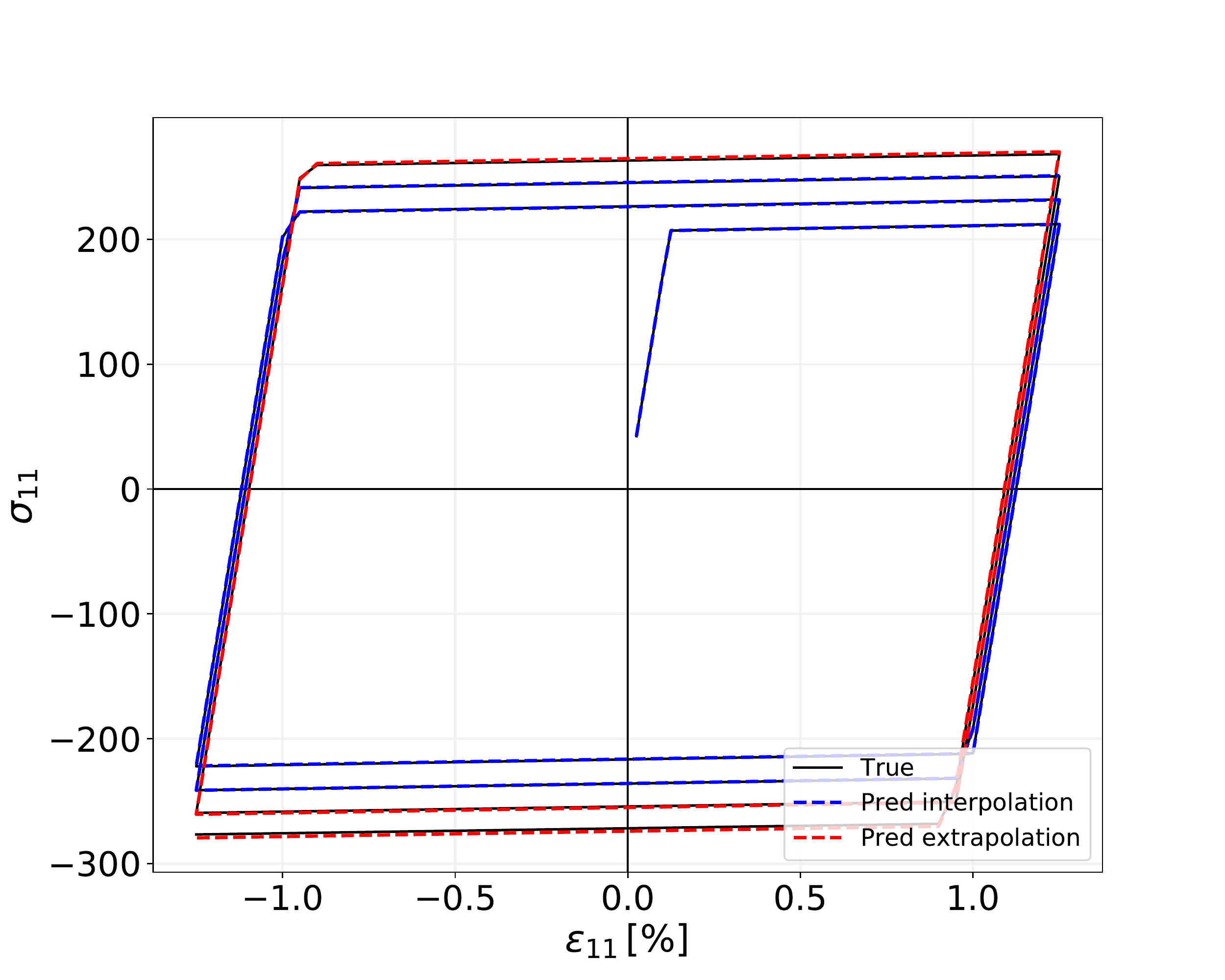}
    \caption{}\label{fig:isoOnlyb}
\end{subfigure}

\begin{subfigure}[b]{0.45\linewidth}
        \centering
    \includegraphics[scale=0.3]{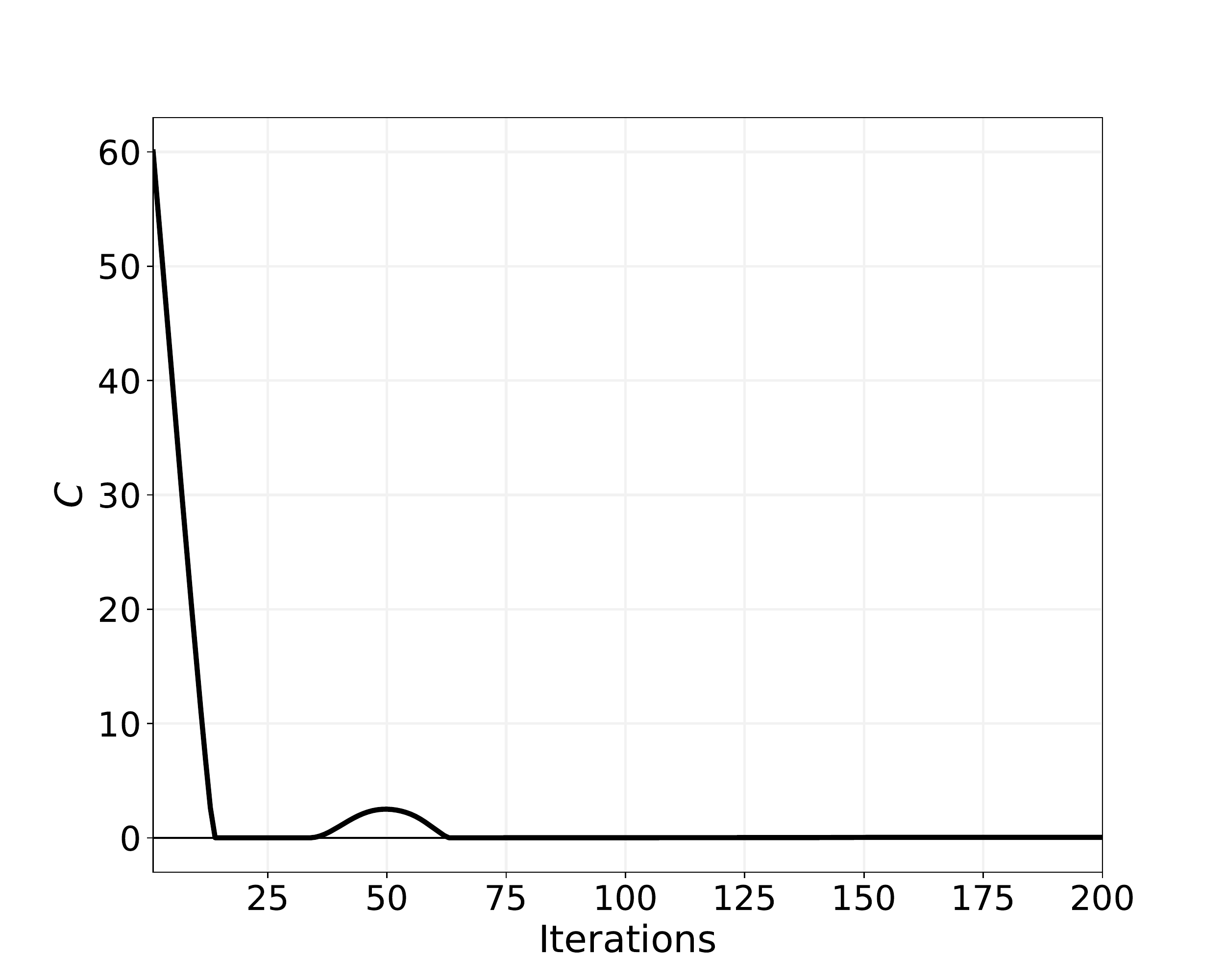}
    \caption{}\label{fig:isoOnlyc}
\end{subfigure}
\begin{subfigure}[b]{0.45\linewidth}
        \centering
    \includegraphics[scale=0.3]{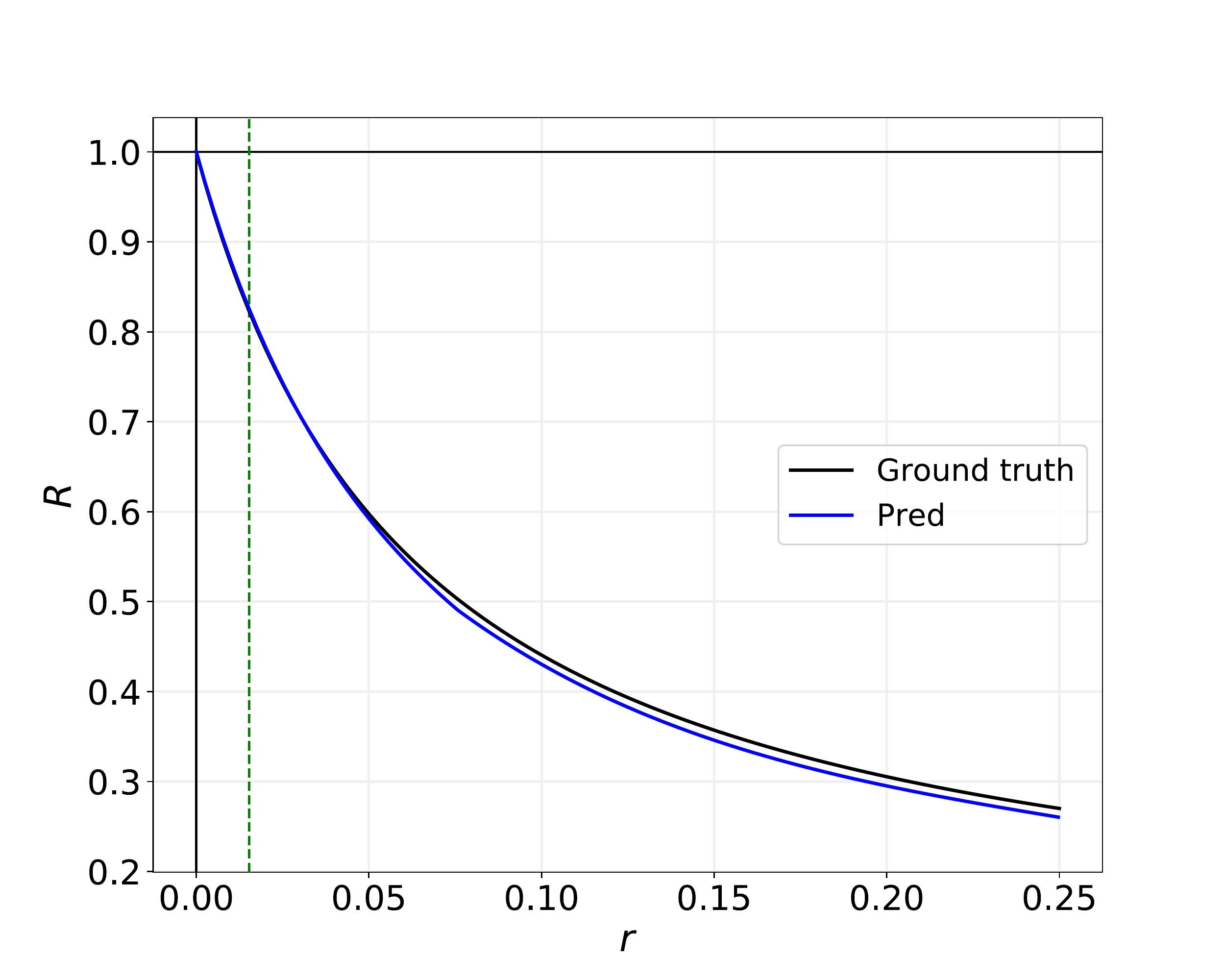}
    \caption{ }\label{fig:isoOnlyd}
\end{subfigure}
\caption{Fit to single NLK model but with only isotropic hardening. (a) Loss behavior, (b) Model fit (blue curve) as well as extrapolation to unseen data (red curve), (b) Evolution of the material parameter $C$ over the training process (ground truth value is $0$), (d) True and predicted isotropic hardening function. Vertical green dotted line indicates maximum seen input data during training.}\label{fig:isoOnly}
\end{figure}

Figure \ref{fig:isoOnlya} shows the normalized loss over $200$ training iterations.
We can see that the error reduces rapidly to a relative value of $1e-6$ after around $100$ iterations.

Using the model with the best loss over this training process we compare the model output to the ground truth in
Figure \ref{fig:isoOnlyb}. Here the blue and red curves indicate interpolation and extrapolation ranges, respectively of the model (which fit the loading as discussed in Figure \ref{fig:singleNLKLoading}).
It can be seen that the model is able to extrapolate well even far beyond the training domain. The reason for the remarkable extrapolation capabilities lies in the fit of the isotropic and kinematic hardening components, as will be discussed below.
Figure \ref{fig:isoOnlyc} shows the evolution of the material parameter $C$ over the training process of $200$ iterations. It can be seen that $C\rightarrow 0$ means that the model discovers the ground truth. i.e. that there is no underlying kinematic hardening behavior in the stress-strain data.
The ground truth and the final prediction of the isotropic hardening curve $R(r)$ are plotted in Figure \ref{fig:isoOnlyd}. Here the green dotted line highlights the maximum value of $r$ that the neural network has seen during training. It can be seen that the ground truth and the function prediction are (basically) identical inside the training domain and are close together even far outside the domain that the neural network has seen during training. The reason for this might the implicit constraint of the ML tool of $R$ to be monotonically decreasing. 

\begin{figure}
    \centering
    \includegraphics[scale=0.3]{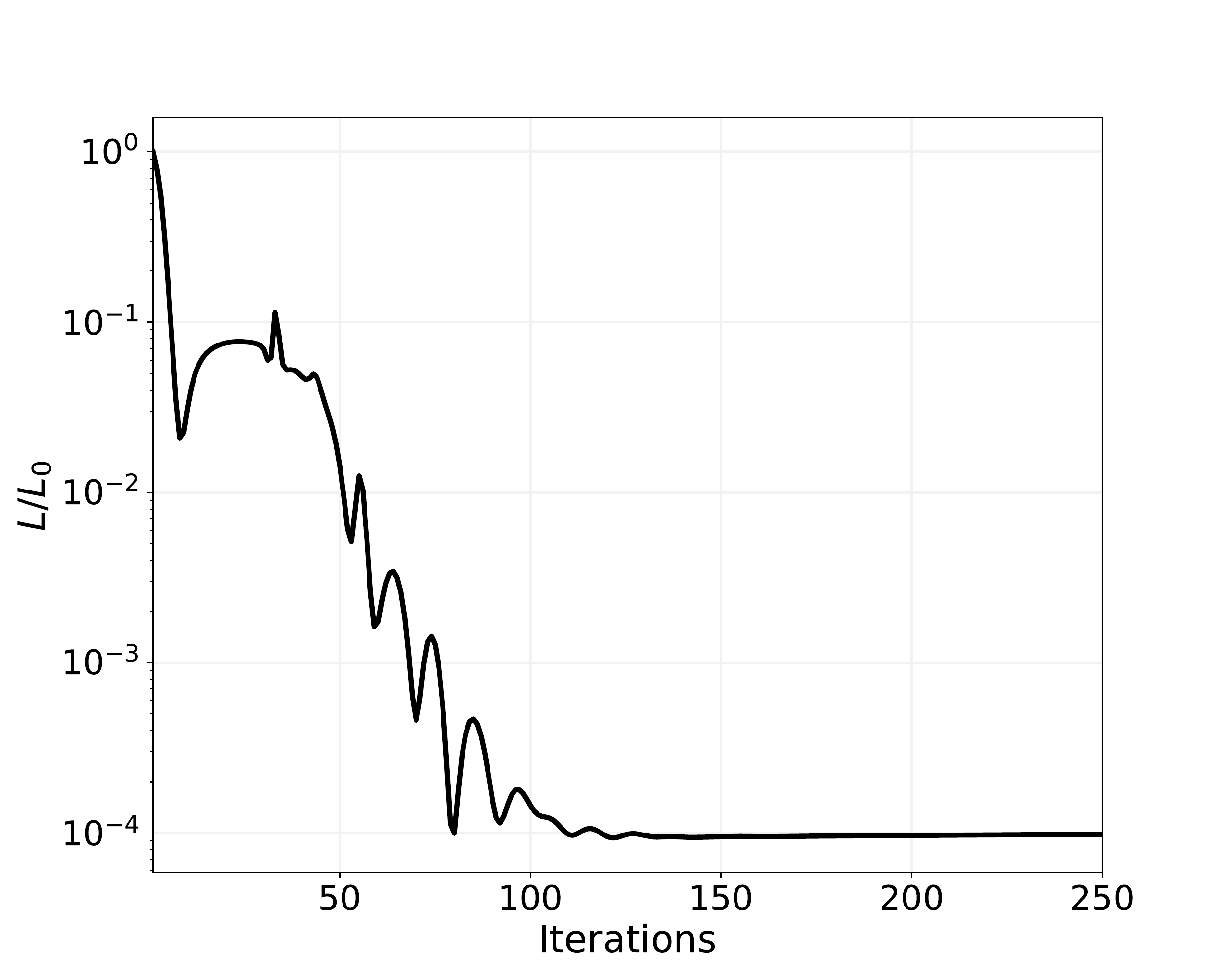}
    \caption{Fit to single NLK model but with only kinematic hardening. Training loss convergence over the training process}
    \label{fig:onlyKinLoss}
\end{figure}

\paragraph{Single NLK - Only kinematic hardening component}
Next, we test if the proposed framework can recover hardening behavior from data that is purely based on kinematic hardening.
For this reason, in the following, we set the ground truth isotropic hardening function as given in eq. \eqref{eq:singleNLKVoce} to zero, i.e. $R=0$.
Using the data generated from the interpolation domain of Figure \ref{fig:singleNLKLoading} (blue line) for training, the normalized loss of the material model is plotted in Figure \ref{fig:onlyKinLoss} over the whole training process of $250$ iterations. It can be seen that the model converged after around $100$ iterations showcasing a similar convergence behavior to the case with only isotropic hardening as discussed in the previous section.

\begin{figure}
\begin{subfigure}[b]{0.45\linewidth}
        \centering
    \includegraphics[scale=0.3]{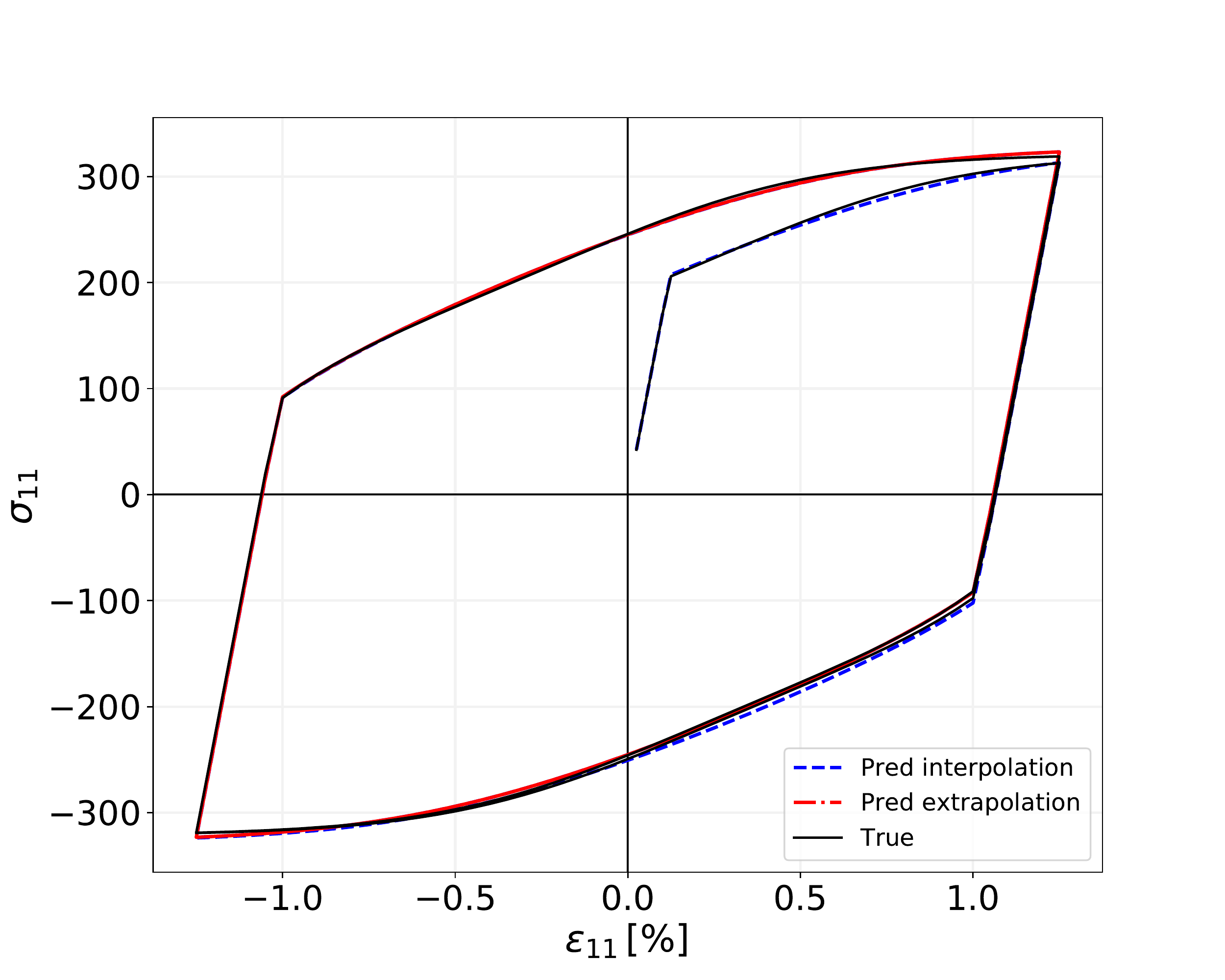}
    \caption{}\label{fig:onlyKina}
\end{subfigure}
\begin{subfigure}[b]{0.45\linewidth}
        \centering
        \includegraphics[scale=0.3]{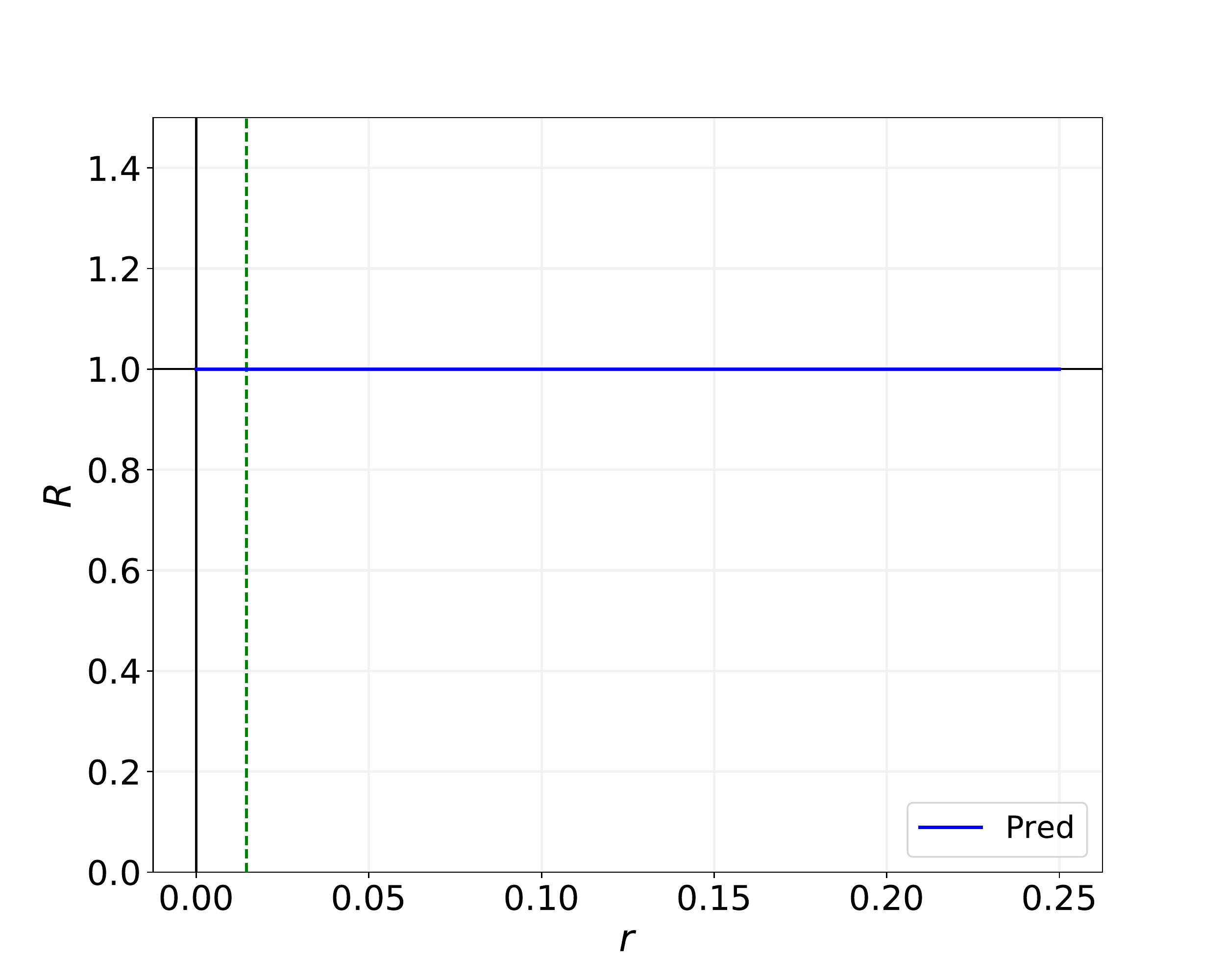}
    \caption{}\label{fig:onlyKinb}
\end{subfigure}

\begin{subfigure}[b]{0.45\linewidth}
        \centering
    \includegraphics[scale=0.3]{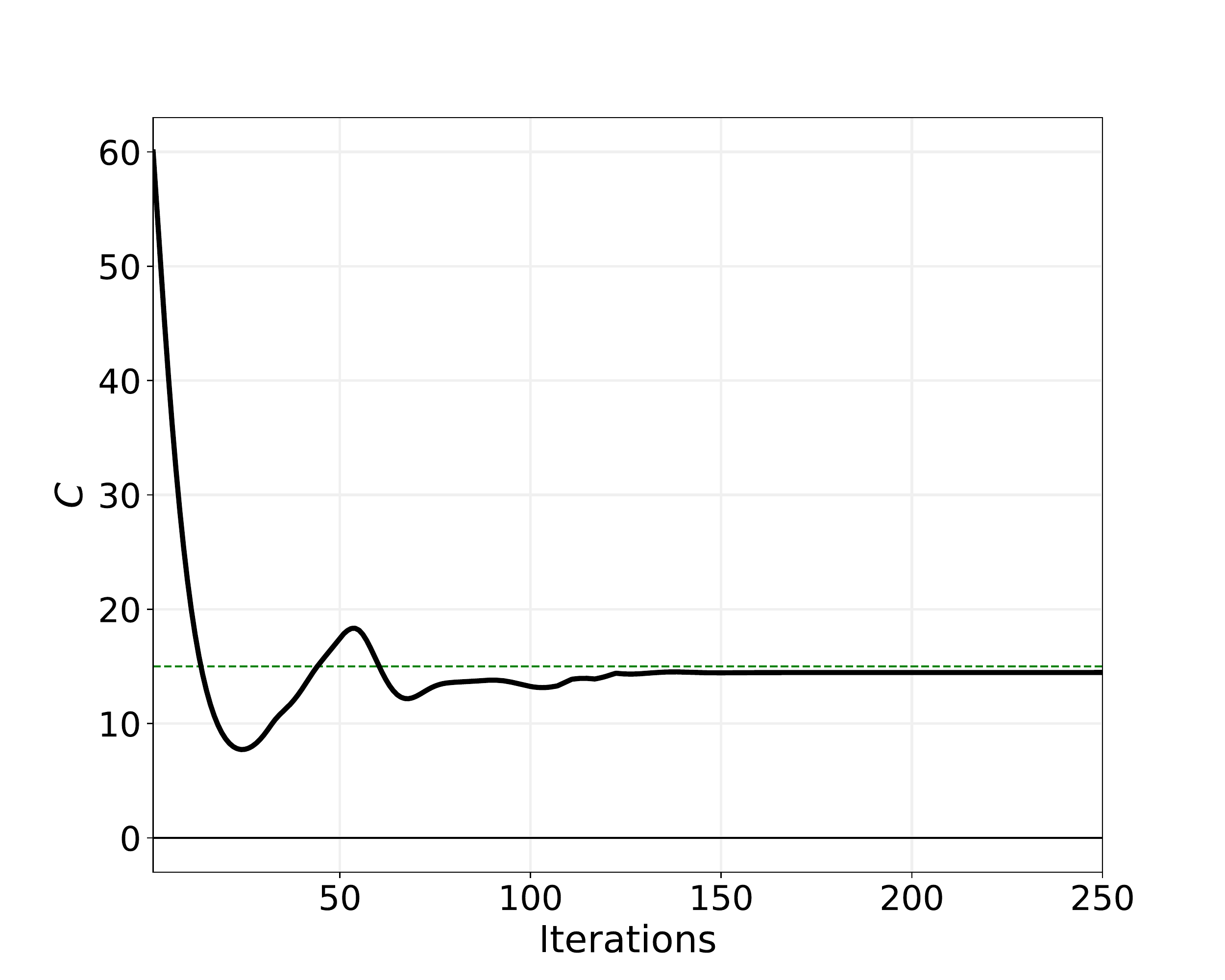}
    \caption{}\label{fig:onlyKinc}
\end{subfigure}
\begin{subfigure}[b]{0.45\linewidth}
        \centering
     \includegraphics[scale=0.3]{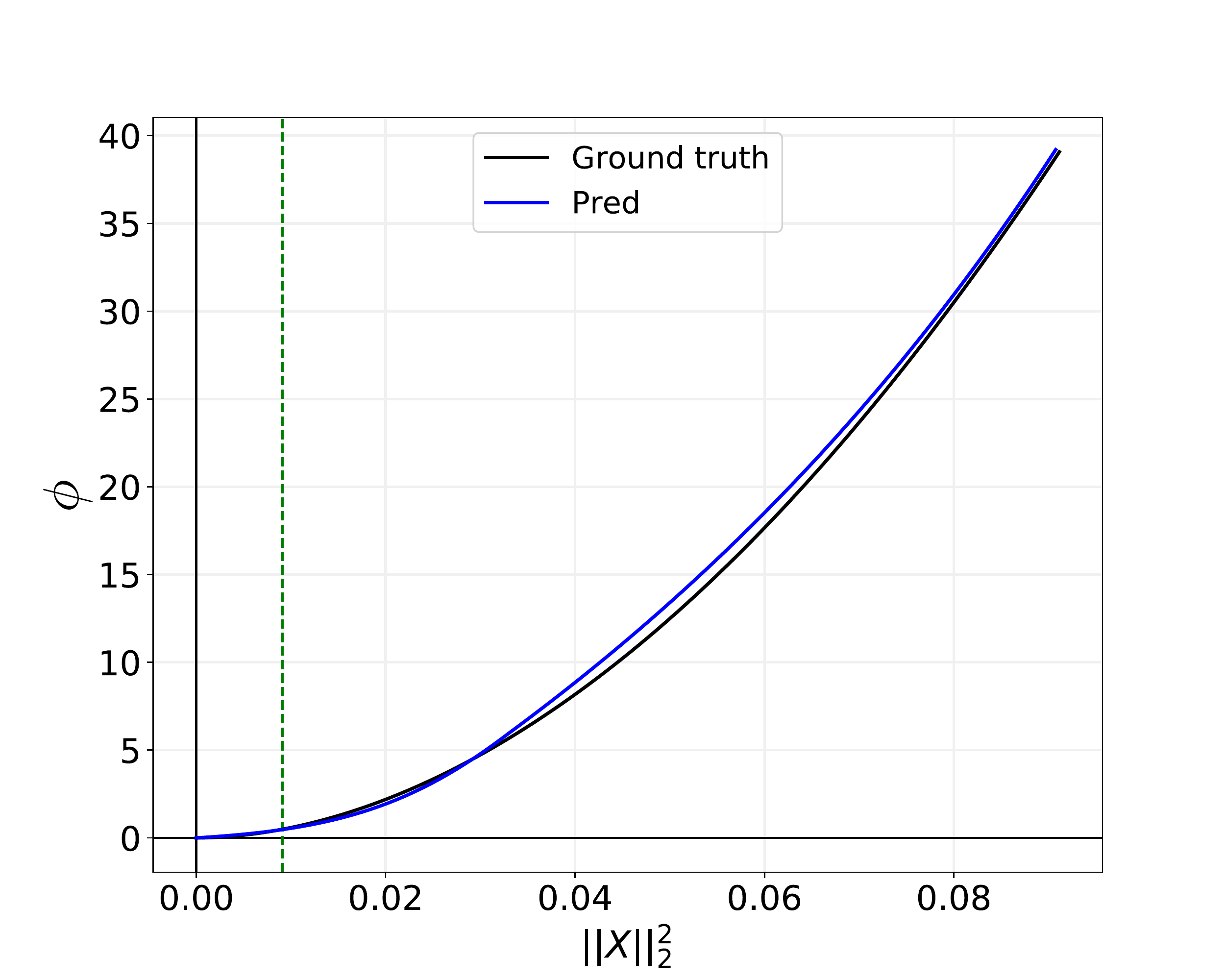}
    \caption{ }\label{fig:onlyKind}
\end{subfigure}
\caption{Fit to single NLK model but with only kinematic hardening. (a) Model fit (blue curve) as well as extrapolation to unseen data (red curve), (b) Predicted isotropic hardening function (ground truth is $R=0$). Green dotted line indicates maximum seen input data during training, (c) Evolution of the material parameter $C$ over the training process (ground truth value is $15$ as indicated by green line),  (d) Predicted nonlinear kinematic hardening function $\phi$. Green dotted line indicates maximum seen input data during training.}\label{fig:onlyKin}
\end{figure}

Figure \ref{fig:onlyKina} displays the stress-strain curve of the trained material model for the domains of interpolation and extrapolation (which correspond to training and testing data). The model captures both domains well. Since there is no isotropic hardening the predictions overlap after the initial cycle.
It can be seen that the model captures this behavior in 
Figure \ref{fig:onlyKinb} where the predicted isotropic hardening function $R$, obtained from the neural network, is plotted over its input $r$. Here, again the green dotted line indicates the maximum value of $r$ that the network has used as an input over the training process. We can see that the framework is able to accurately identify that no isotropic hardening is underlying the ground truth data.

Furthermore, the essential features of the kinematic hardening model are also captured by the proposed framework. Figure \ref{fig:onlyKinc} plots the evolution of the material parameter $C$ over the training process which seems to be converging towards the ground truth value of $15$.
The output of the neural network used to model $\phi$ as well as the graph of the true function are displayed in Figure \ref{fig:onlyKind}. The green dotted line indicates the maximum input value that the network has been trained with. We can see that the predicted function closely follows the ground truth and furthermore that the constraints on $\phi$ (convex, positive, monotonically increasing, initial value) are all fulfilled, which might be the reason for the goodness of fit.

\begin{figure}
    \centering
    \includegraphics[scale=0.25]{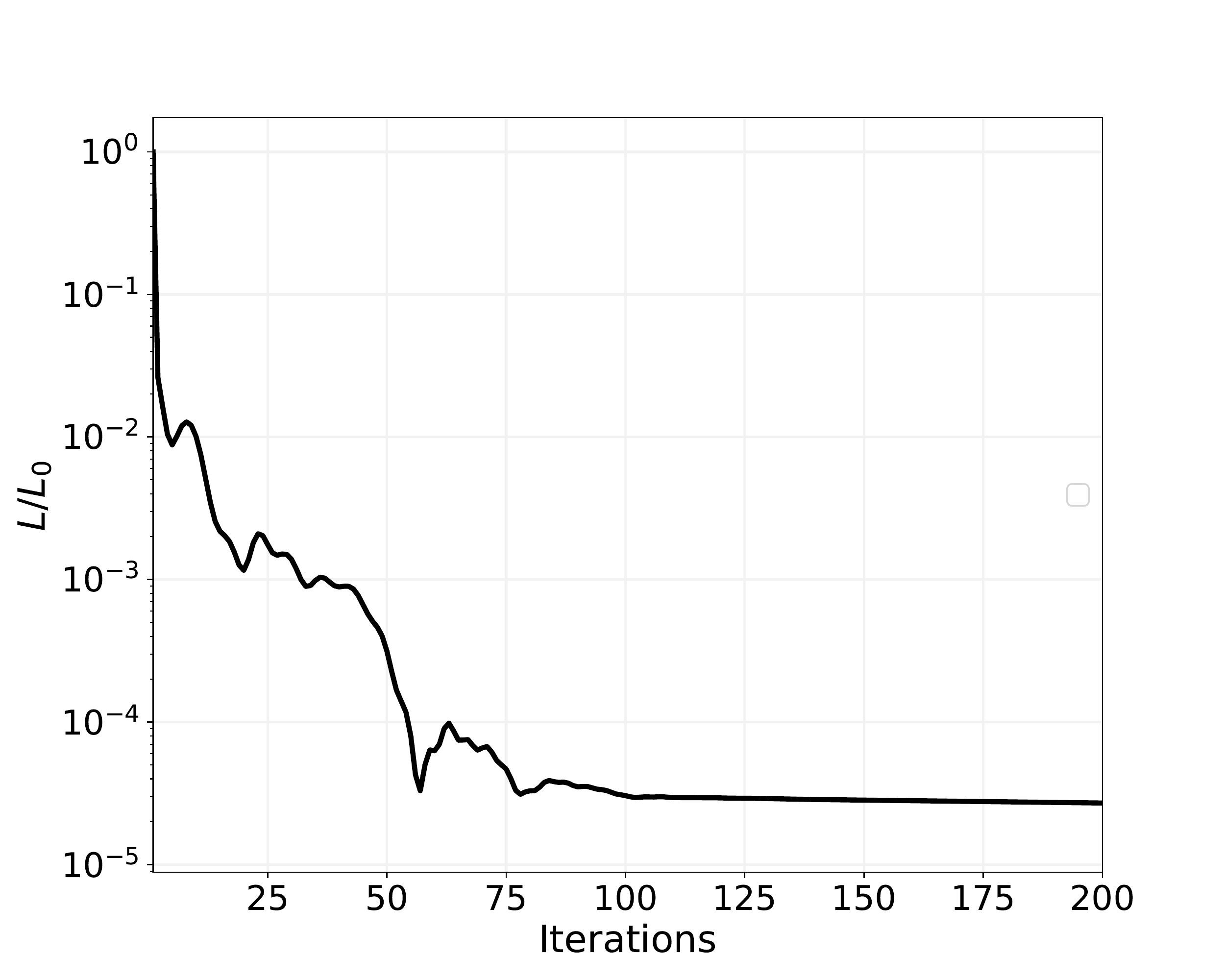}
    \caption{Fit to single NLK model with mixed hardening. Training loss convergence over the training process}
    \label{fig:singleNLKLoss}
\end{figure}

\begin{figure}
\begin{subfigure}[b]{0.45\linewidth}
        \centering
    \includegraphics[scale=0.3]{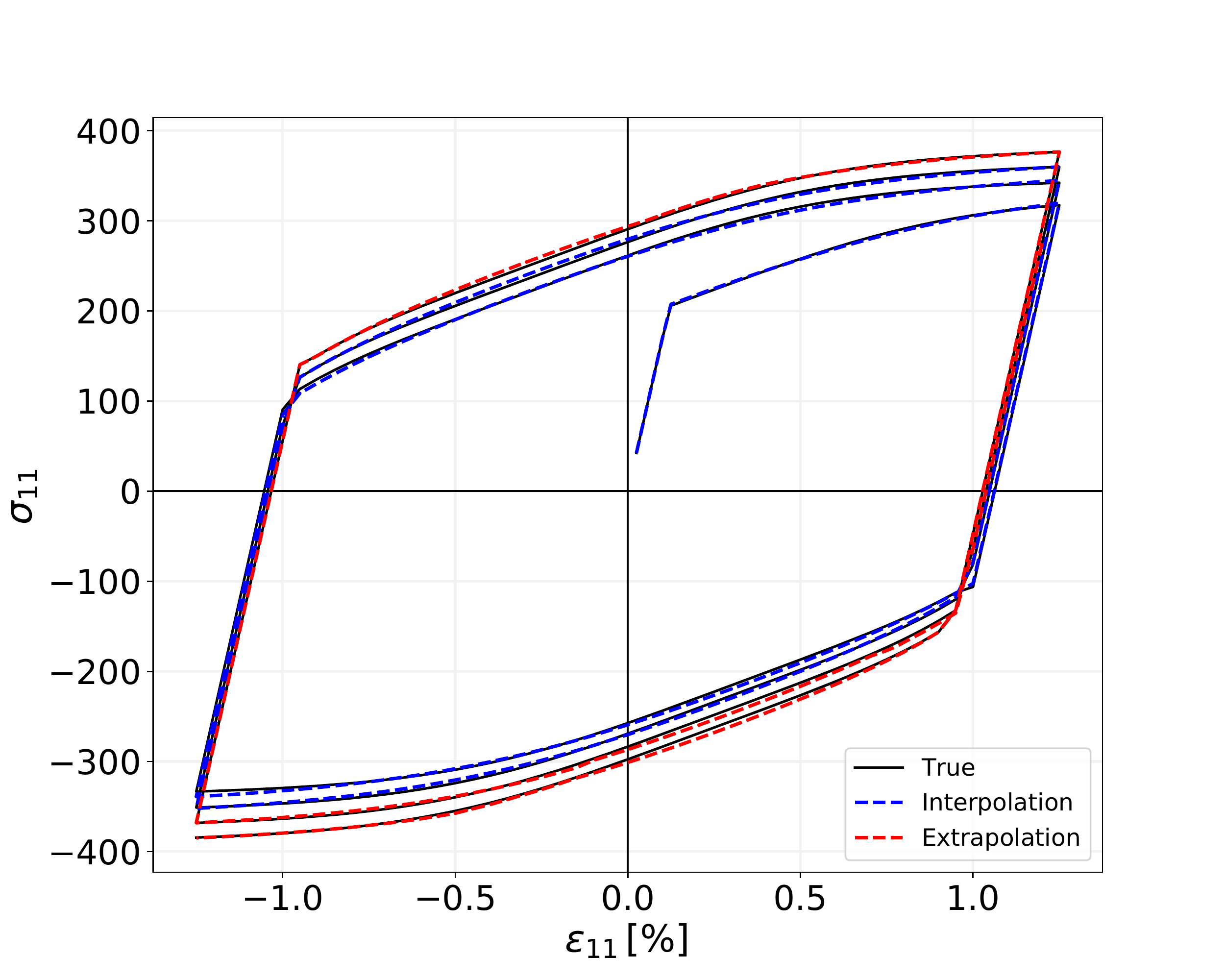}
    \caption{}\label{fig:singleNLKa}
\end{subfigure}
\begin{subfigure}[b]{0.45\linewidth}
        \centering
        \includegraphics[scale=0.3]{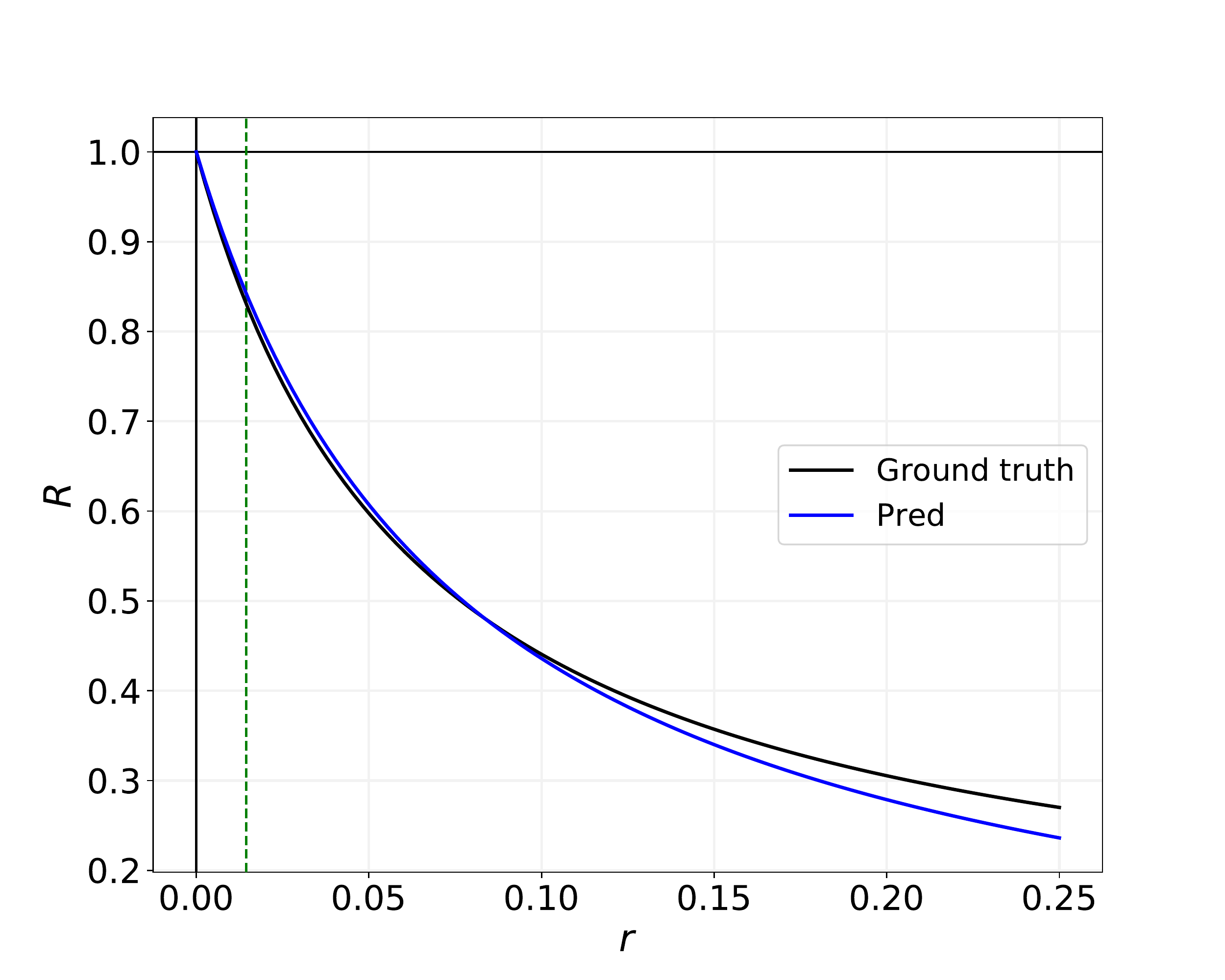}
    \caption{}\label{fig:singleNLKb}
\end{subfigure}

\begin{subfigure}[b]{0.45\linewidth}
        \centering
    \includegraphics[scale=0.3]{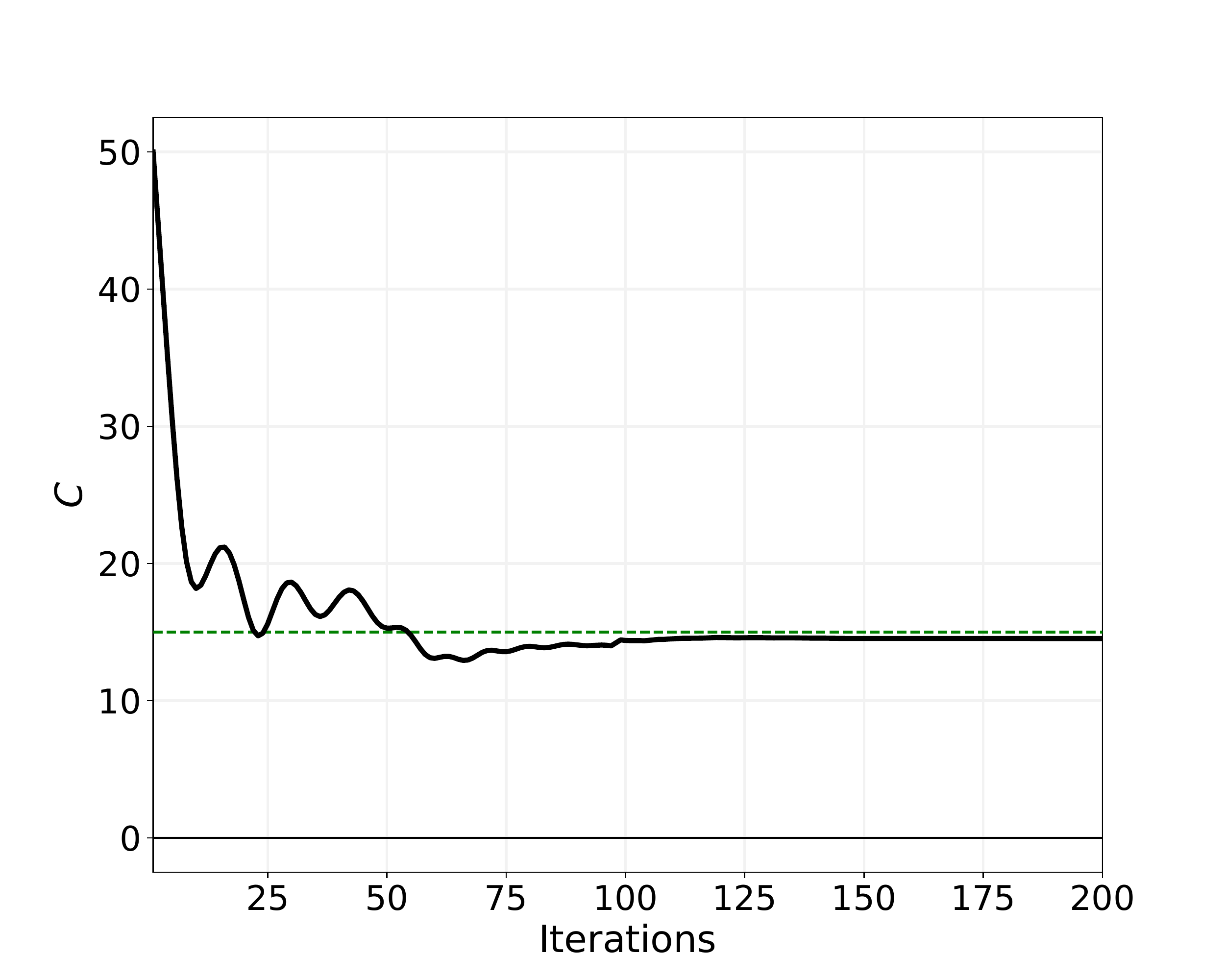}
    \caption{}\label{fig:singleNLKc}
\end{subfigure}
\begin{subfigure}[b]{0.45\linewidth}
        \centering
    \includegraphics[scale=0.3]{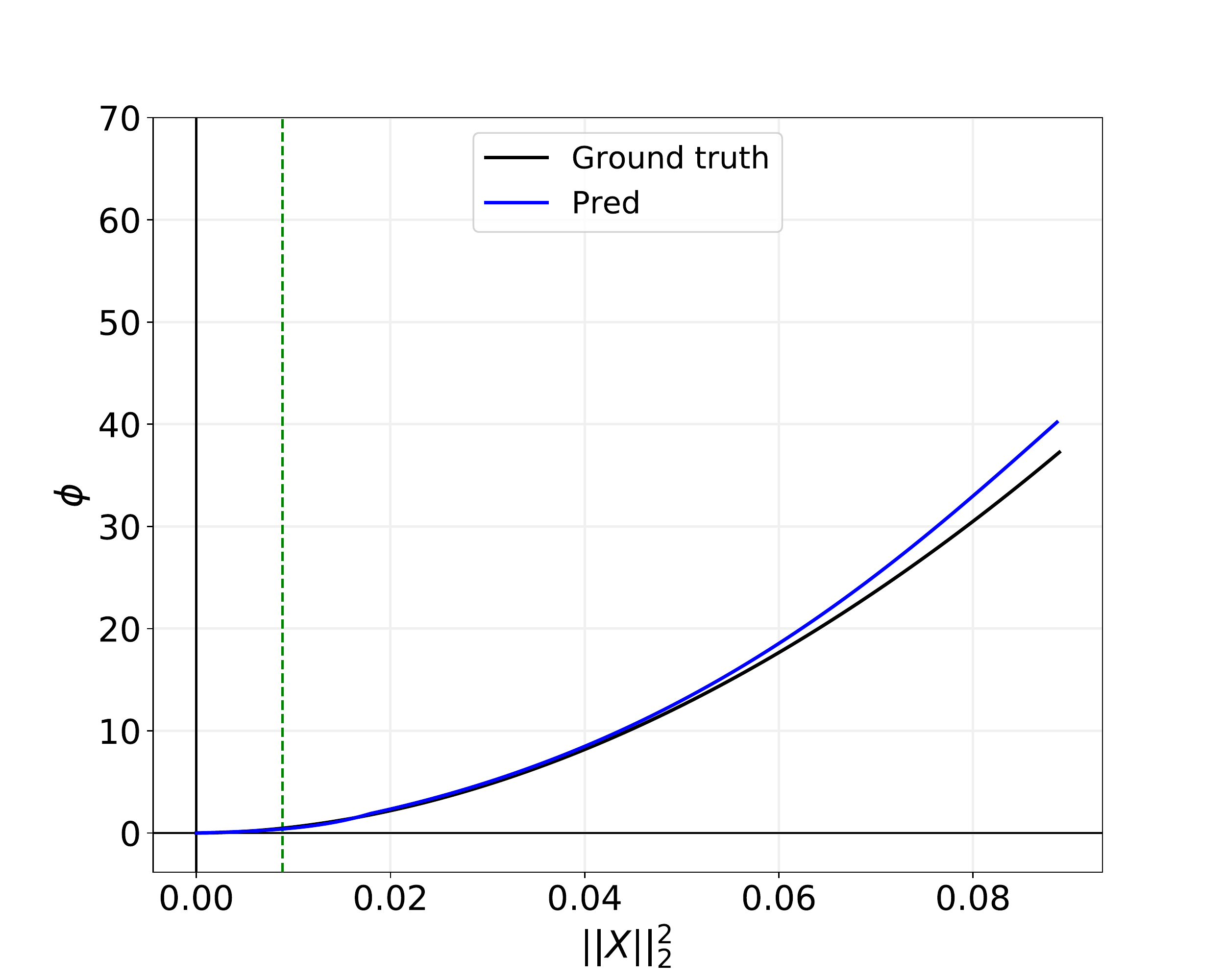}
    \caption{ }\label{fig:singleNLKd}
\end{subfigure}
\caption{Fit to single NLK model with mixed hardening. (a) Model fit (blue curve) as well as extrapolation to unseen data (red curve), (b) Predicted isotropic hardening function (ground truth is $R=0$). Green dotted line indicates maximum seen input data during training, (c) Evolution of the material parameter $C$ over the training process (ground truth value is $15$ as indicated by green line),  (d) Predicted nonlinear kinematic hardening function $\phi$. Green dotted line indicates maximum seen input data during training.}\label{fig:singleNLK}
\end{figure}

\paragraph{Single NLK - Mixed hardening}
In a last experiment involving the single NLK model, we check if the framework is able to capture both the isotropic and the kinematic hardening behavior that underlie the dataset.
Similarly to the previous two cases, the normalized loss converges after around $100$ iterations; see Figure \ref{fig:singleNLKLoss}.
The interpolation and extrapolation response of the trained model due to the applied loading of Figure \ref{fig:singleNLKLoading} is plotted in Figure \ref{fig:singleNLKa}. The trained model appears to fit the ground truth very well in both loading phases. 
Figure \ref{fig:singleNLKb} compares the true and the predicted response of the isotropic hardening function. The green line indicates the maximum value of $r$ that has been used as an input to the neural network over the training process. We can see that both curves are in good agreement even far outside the training domain.

The evolution of the material parameter $C$ over the training process is displayed in
Figure \ref{fig:singleNLKc}. It can be seen that the predicted value closely matches the ground truth value of $C=15$ (highlighted by the green line) after around $100$ iterations.
Lastly, the prediction of the nonlinear kinematic hardening function of this case is compared to the ground truth in Figure \ref{fig:singleNLKd}.
The maximum input value to the neural network is marked by the green dotted line. It can be seen that the predicted functional form adheres to its functional constraints (convexity, etc.) while showing a similar response as the true $\phi$-function.

Next, we compare the performance of the proposed framework to a classical fitting approach which involves the phenomenological model. Here, we assume the best-case scenario (for the phenomenological model), i.e. that the user knows that the isotropic and kinematic hardening behavior is given by eqs. \eqref{eq:singleNLKVoce} and \eqref{eq:singleNLKX} and now wants to fit the unknown hardening parameters. 
This is a best-case scenario in which the functional forms are assumed to be exactly known. If this is not the case, the whole functional form of the hardening behavior of these models is oftentimes fitted using genetic algorithms which involve a large number of function calls to the black-box to obtain the predicted stress-strain curve, e.g. 12500 calls in
\cite{furukawa1997inelastic}, 12500-17500 calls in \cite{jenab2016use}, 7000 calls in \cite{fernandez2018genetic}.

\begin{figure}
        \centering
    \includegraphics[scale=0.45]{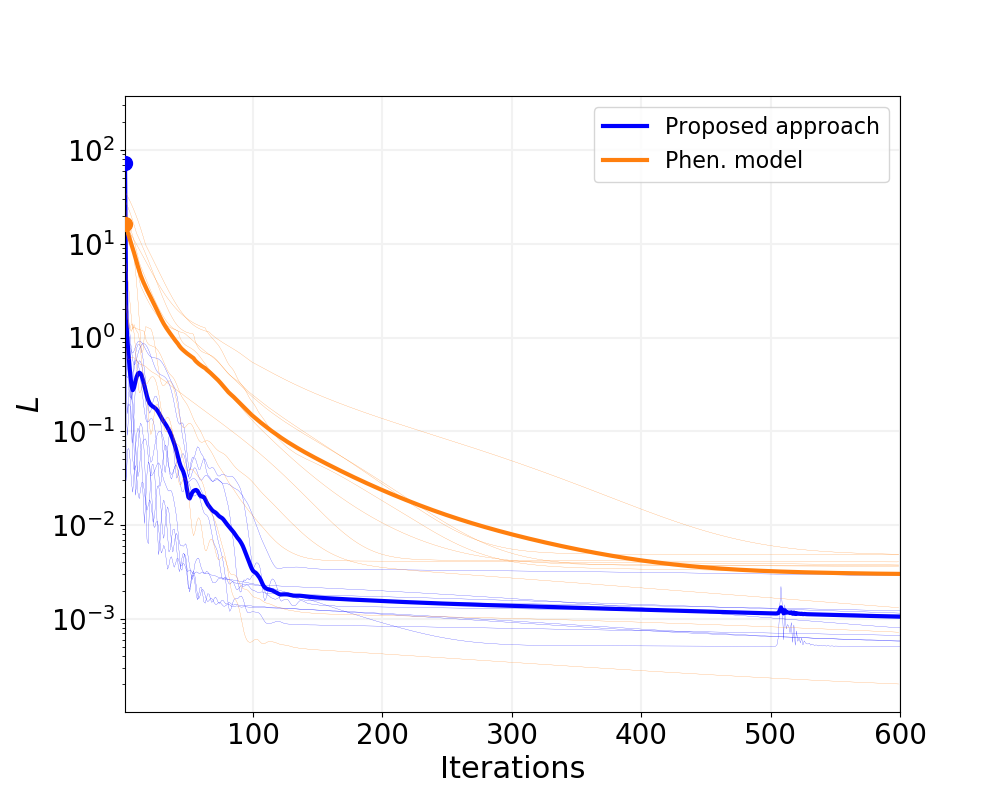}
    \caption{Comparison of training loss behavior between phenomenological fit (orange) and proposed approach (blue). 10 faint lines for each color indicate the losses for each of the 10 training runs while the bold line represents the respective averaged loss for each case. }\label{fig:ModelvsParamsingeNLK}
\end{figure}

Here, we assume that rough guesses for the 6 trainable parameters are known, given by the upper and lower bounds of Table \ref{tab::SingleNLKParam}. We see that these bounds roughly involve the true values.
Given these bounds, we uniformly sample 10 of the starting guesses for the parameter values and then apply the same gradient descent framework (Algorithm \ref{alg:Train}) that we use to fit the proposed neural network framework.
For a fair comparison of the performance of this parameter fitting process to our neural network-based approach,
the weights and biases of the neural networks are also randomly chosen (10 times) from the discussed initialization process using 10 different random seeds while the initial guess for the material parameter $C$ is chosen uniformly from the same range as indicated in Table \ref{tab::SingleNLKParam}.
Ten runs of each of these models are of course not necessarily a representative sample size but they can provide us with rough guidance on the convergence behavior of these two methods.
The training loss for both approaches over $600$ iterations is shown in 
Figure \ref{fig:ModelvsParamsingeNLK}. Here, the loss evolution of the 10 runs for each of the two cases is plotted in faint lines while the averaged loss behavior (over the 10 runs) is indicated with the bold-faced line. Here, we observe an interesting trend for the two optimization procedures. The training error of the proposed data-driven approach seems to converge quicker to its stable local minimum (after around $100$ iterations) compared to the phenomenological approach (around $500$ iterations). Furthermore, the final averaged error for the surrogate model is roughly $3$ times lower after $600$ iterations. This is surprising since the functional form of the phenomenological model is chosen to exactly match the true form, and the fit only involves $6$ trainable parameters. On the other hand, our proposed framework is not biased by any assumptions about the true functional form and has to fit around $270$ parameters. In light of this difference in the number of parameters, we were expecting the training loss of the phenomenological model to converge quicker than that of the surrogate model. Additionally, similar arguments can be posited for the error. But we have to acknowledge that the magnitude of the averaged  loss for both models is acceptable. 

Overall, this trend might be due to the fact that --contrary to our initial belief-- more parameters could allow for a quicker traversing of the loss landscape and to avoid local minima. This behavior is unexpected and might be due to the tightness of the bounds but studying it more deeply is outside of the scope of this paper and will be investigated in future works.

\begin{table}
\begin{center}
\begin{tabular}{||c | c c c c c c ||} 
 \hline
Parameter & $C$ & $\gamma$ & $m$ & $H_{1}$ & $H_{2}$ & $H_{3}$  \\ [0.5ex] 
 \hline\hline
Ground truth  & 15 & 550 & 0.9 & 0.1875 & 0.25 & 2.0 \\ 
 \hline
 Lower bound  & 1 & 1 & 0.5 & 0.01 & 0.01 &  0.01  \\ 
 \hline
 Upper bound  & 100 & 2000 & 1.5 & 5 & 5 & 5 \\ 
 \hline
\end{tabular}
\end{center}
\caption{Non-dimensional material parameters for Single NLK-model.}\label{tab::SingleNLKParam}
\end{table}

\paragraph{Single NLK - Mixed hardening FEM comparison}
We test the performance of the surrogate model by embedding it into a rewritten C$^\text{++}$ finite element (FE) code provided by \cite{yaghoobi2019prisms}. The goal is to ensure that the trained data-driven models work for complex stress states - beyond the uniaxial training data - and allow for a similar convergence behavior of the structural FE problem compared to the phenomenological model. 
To do so, we compare the behavior of the proposed elasto-plastic framework to the ground truth single NLK model on two standard structural benchmark tests.

\begin{figure}
\begin{subfigure}[b]{1.0\linewidth}
    \centering
    \includegraphics[scale=0.4]{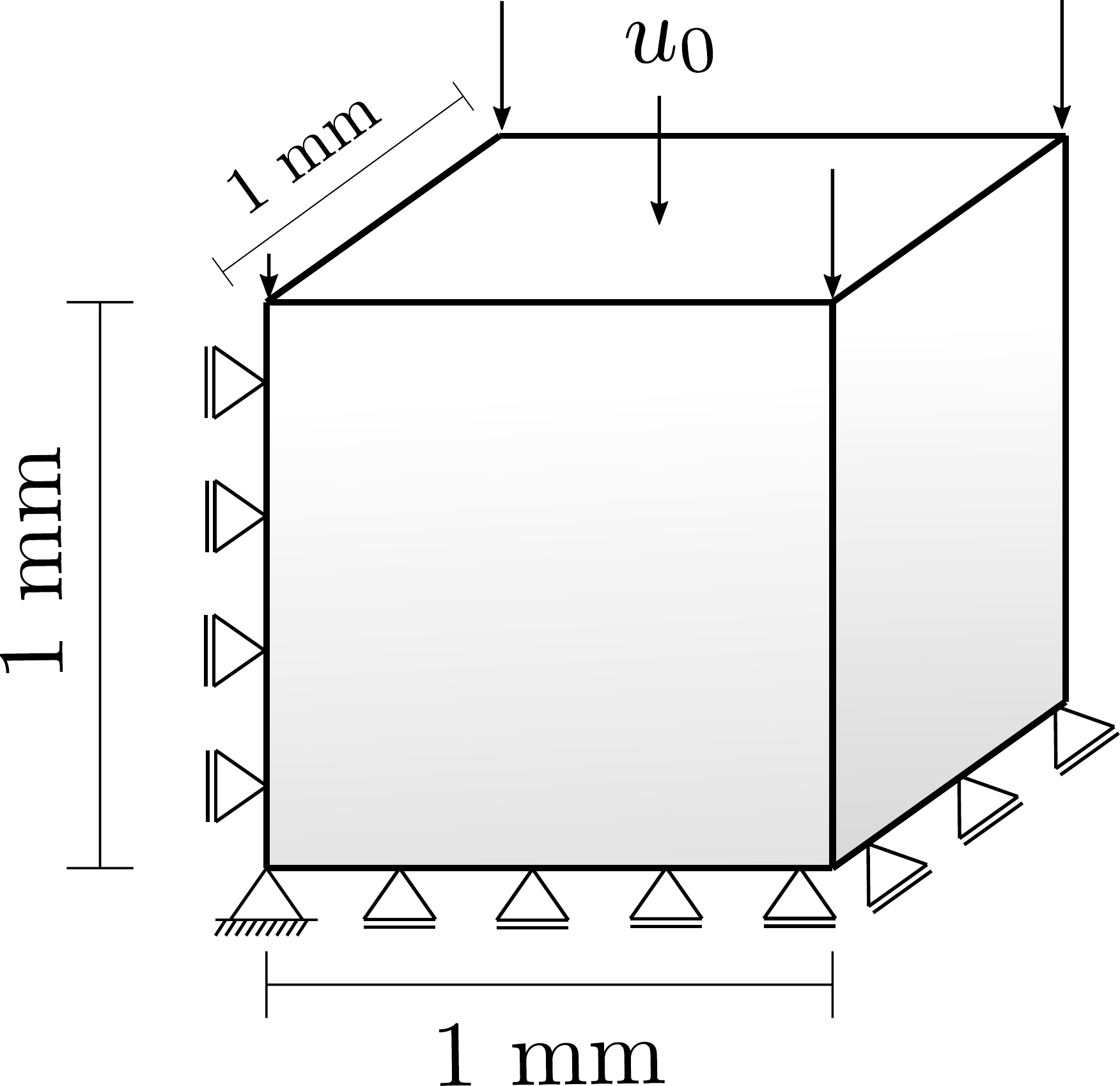}
    \caption{FEM structural problem: Punch test}
    \label{fig:punchTestApp}
    \end{subfigure}

\begin{subfigure}[b]{0.45\linewidth}
        \centering
    \includegraphics[scale=0.2]{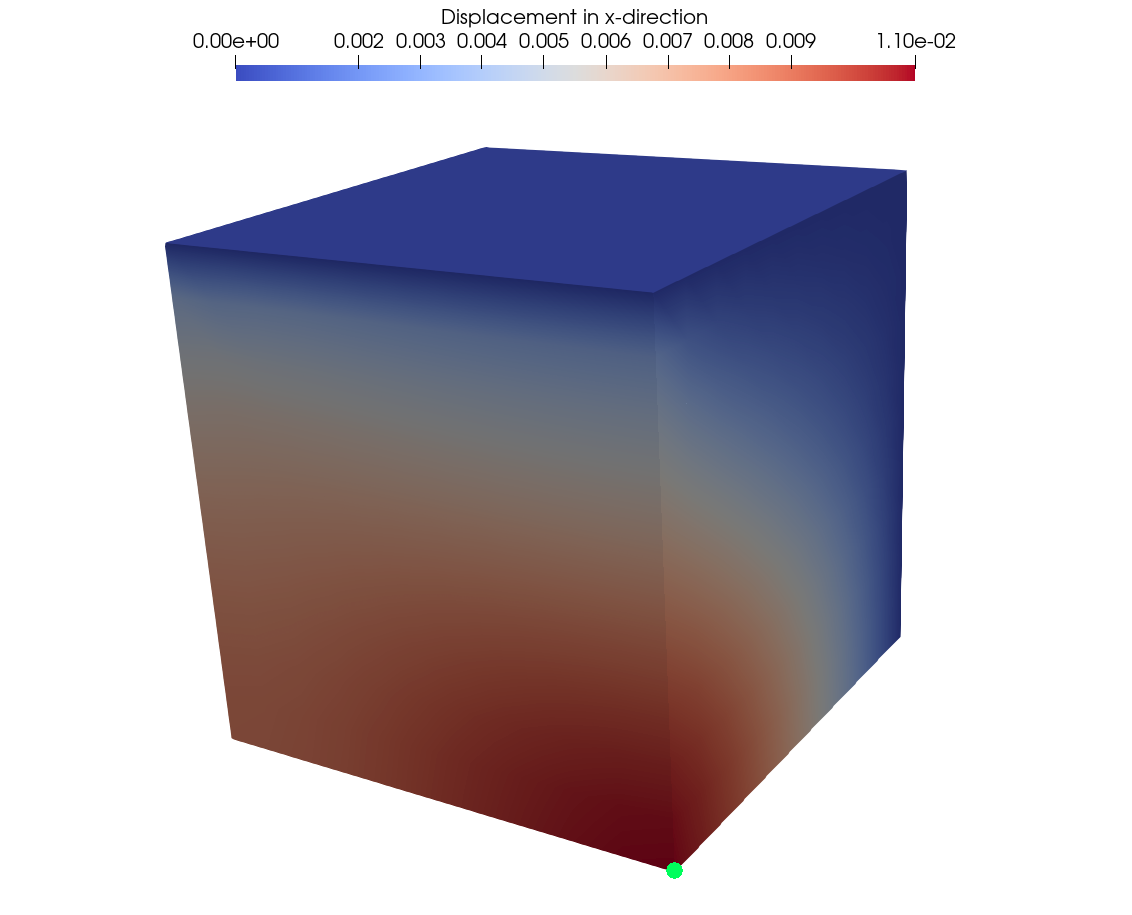}
    \caption{Ground truth response}\label{fig:punchTestAppModel}
\end{subfigure}
\begin{subfigure}[b]{0.45\linewidth}
        \centering
    \includegraphics[scale=0.2]{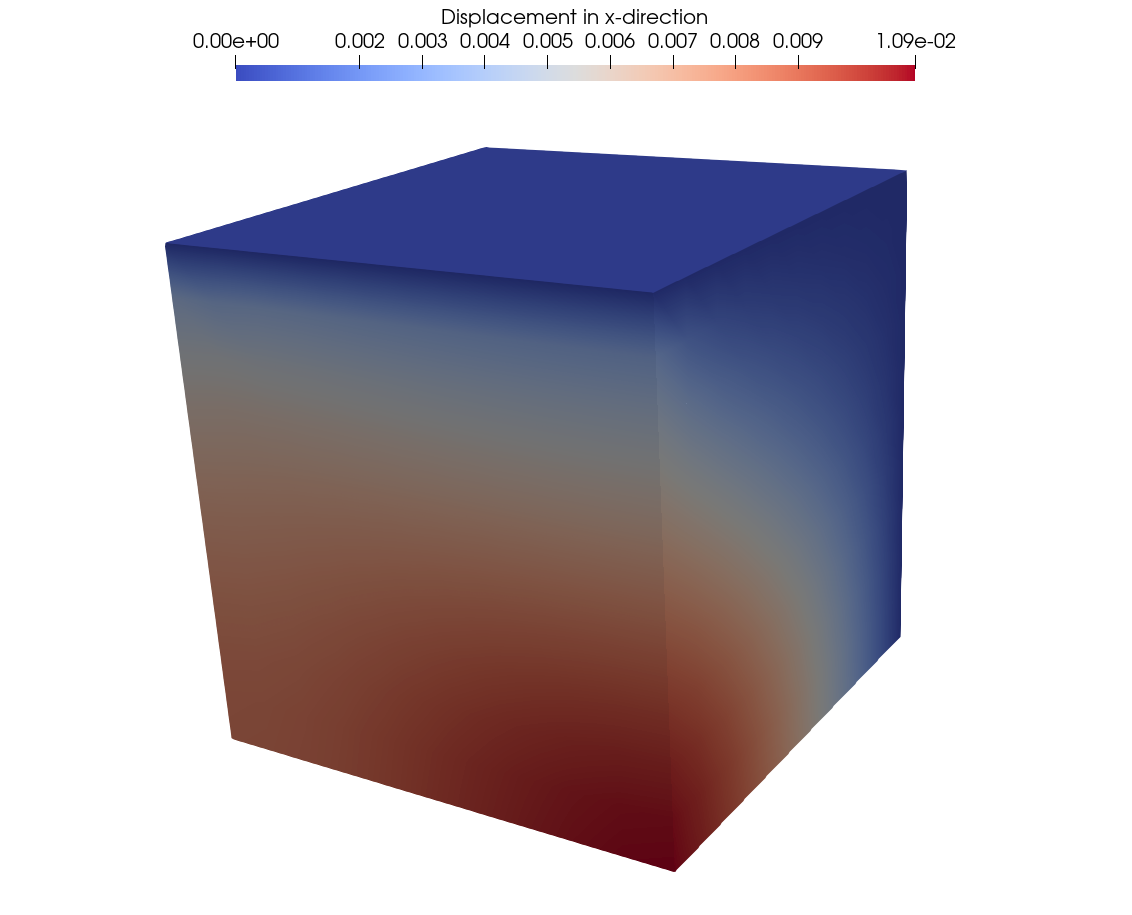}
    \caption{ Proposed framework response}\label{fig:punchTestAppNN}
\end{subfigure}
\caption{FEM benchmark test I. (a) Punch test setup (b) Displacement in $x$-direction using the ground truth phenomenological model. Green dot indicates position of loading stress-strain curves used in Figure \ref{fig:punchTestAppLoadDisp}, (c)  Displacement in $x$-direction using the developed hybrid framework.}
\end{figure}

\begin{figure}
\begin{subfigure}[b]{0.45\linewidth}
        \centering
    \includegraphics[scale=0.3]{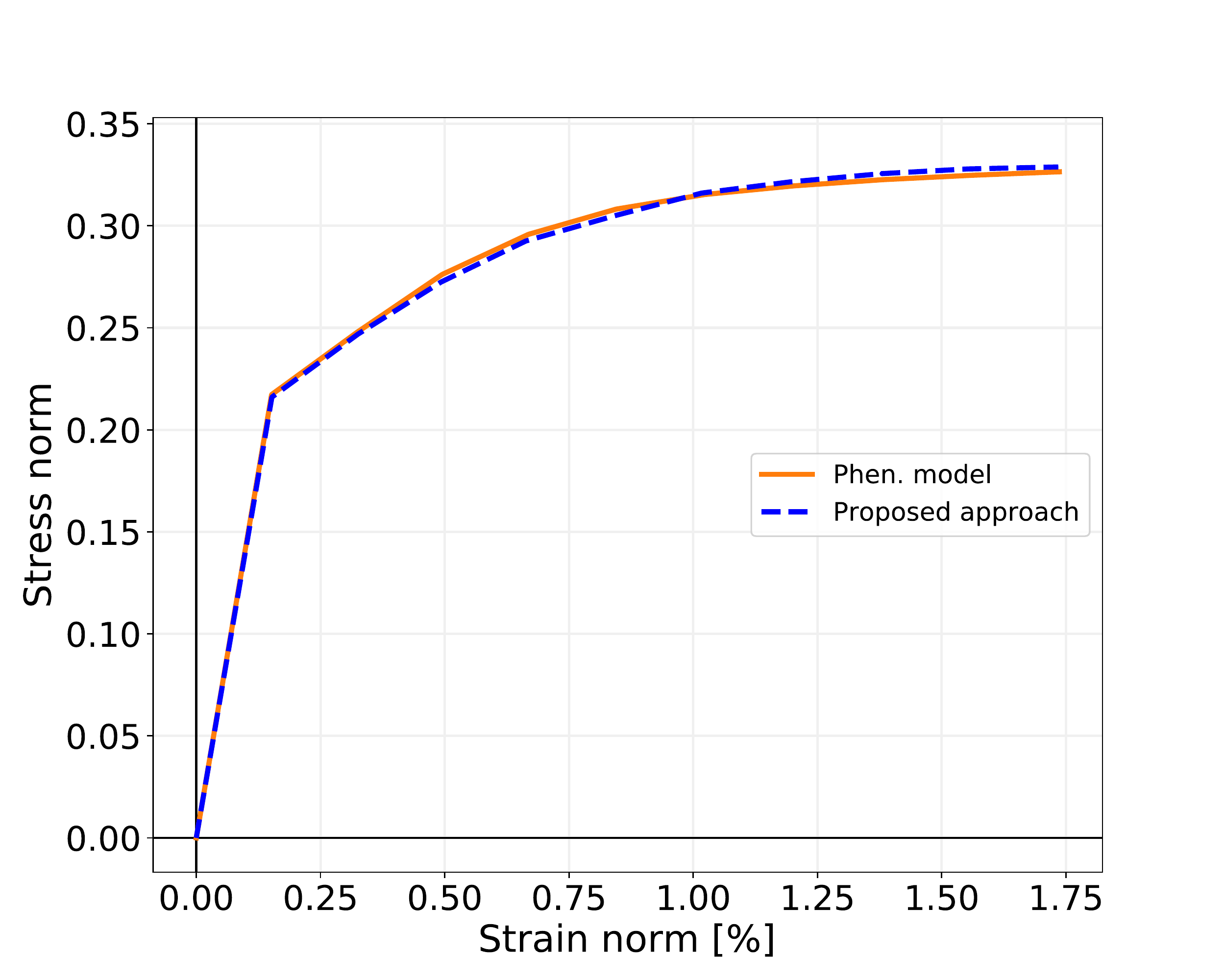}
    \caption{}\label{fig:punchTestAppLoadDisp}
\end{subfigure}
\begin{subfigure}[b]{0.45\linewidth}
        \centering
    \includegraphics[scale=0.3]{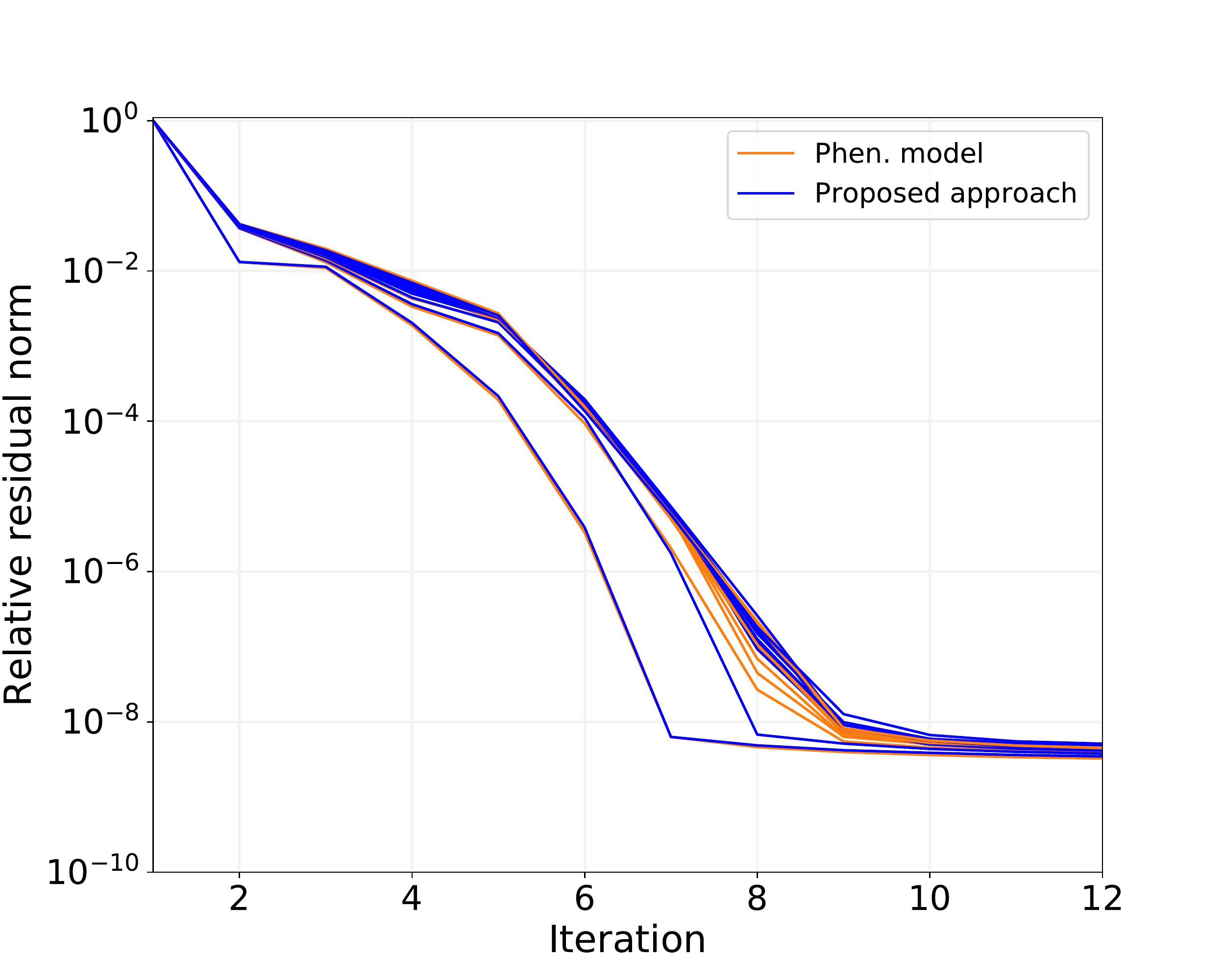}
    \caption{ }\label{fig:punchTestAppConv}
\end{subfigure}
\caption{ Punch test results: (a) Stress-strain norm curves at the point $(1.0,1.0,0.0)$ (green dot in Figure \ref{fig:punchTestAppLoadDisp}) over the loading process, (b) Numerical convergence behavior of the structural problem when using phenomenological and our proposed approach.}
\end{figure}

The first benchmark is known as the Punch test \cite{huang2020machine}.
It consists of a block which at the bottom is only fixed in the vertical direction while 
vertical displacement ($u_{0} = 0.015 mm$ ) is applied to the top of the block, see Figure \ref{fig:punchTestApp}. The model is solved by involving 
10 load increments each applied using 12 iterations.
Figures \ref{fig:punchTestAppModel} and \ref{fig:punchTestAppNN}
show the displacement in the $x$-direction for the ground truth model as well as the surrogate material model, respectively. It can be seen that the responses match. This indicated that the proposed framework is also able to reliably predict the true response under complex loading conditions.
This is also highlighted by
Figure \ref{fig:punchTestAppLoadDisp} where the stress-strain norm curves at the point $(1.0,1.0,0.0)$ (green dot in Figure \ref{fig:punchTestAppLoadDisp}) are shown over the loading process. Here again, the prediction closely fits the ground truth model.
Lastly, we can compare the convergence behavior of the structural problem of the two modeling approaches. Figure \ref{fig:punchTestAppConv} plots the relative residual norm of the global FE system over the number of iterations for each of the 10 loading increments. It can be seen that the convergence behavior of the proposed framework is indistinguishable from the classical phenomenological model. This is uncommon between data-driven constitutive models, where the convergence in a finite element setting usually lags compared to when using their analytical counterparts.

\begin{figure}
    \begin{subfigure}[b]{1.0\linewidth}
    \centering
    \includegraphics[scale=0.9]{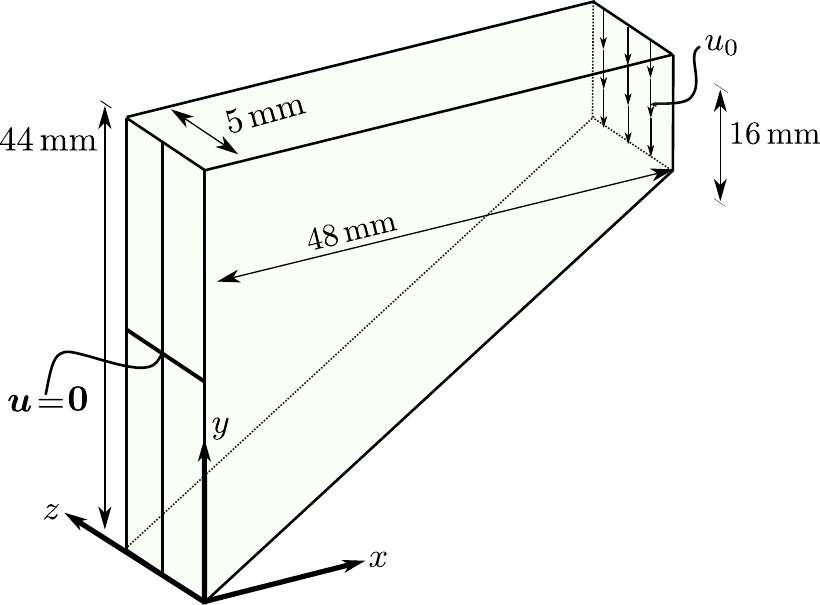}
    \caption{FEM structural problem: Cook's membrane test}
    \label{fig:cookMemApp}
    \end{subfigure}

    \begin{subfigure}[b]{0.48\linewidth}
        \centering
    \includegraphics[scale=0.2]{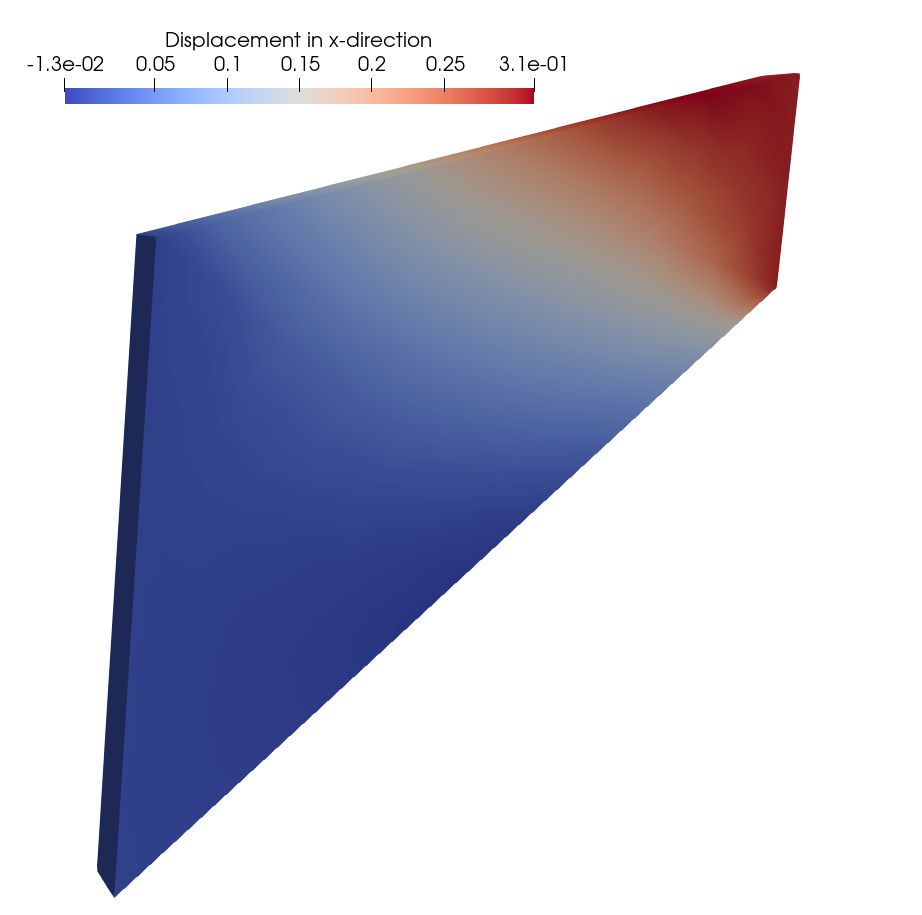}
    \caption{Ground truth response}\label{fig:cookMemAppModel}
\end{subfigure}
\begin{subfigure}[b]{0.48\linewidth}
        \centering
    \includegraphics[scale=0.2]{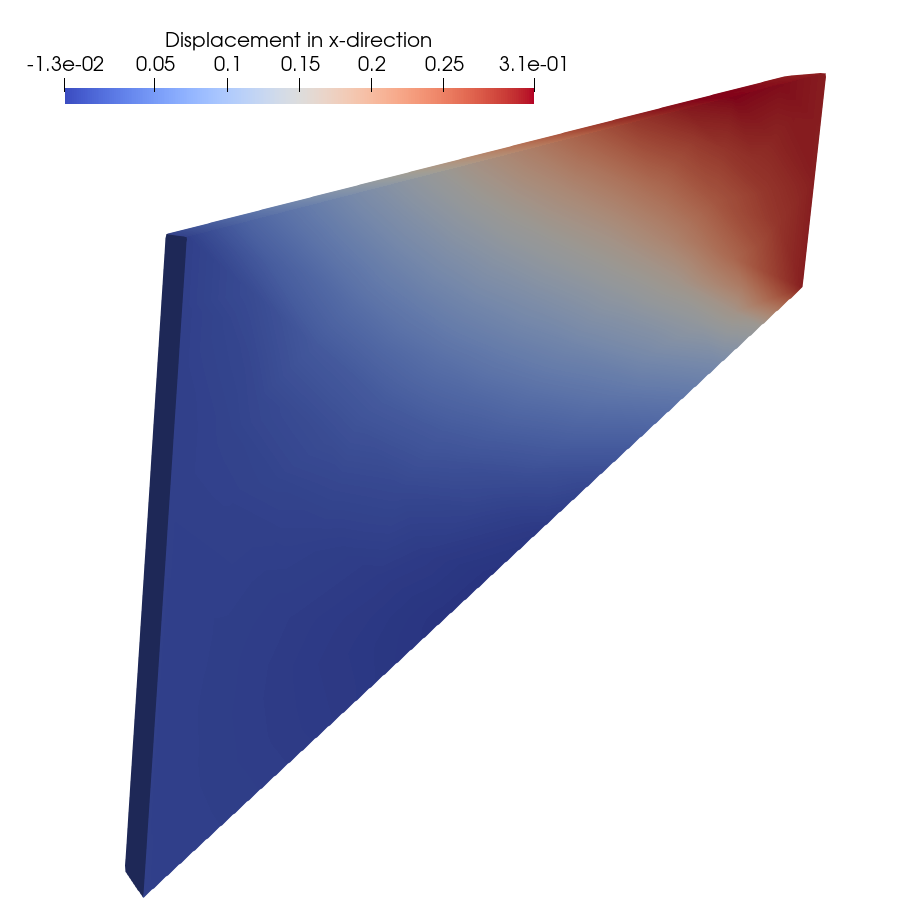}
    \caption{Proposed framework response }\label{fig:cookMemAppNN}
\end{subfigure}
\caption{FEM benchmark test II. (a) Cook's membrane setup (b) Displacement in $x$-direction using the ground truth phenomenological model, (c)  Displacement in $x$-direction using the developed hybrid framework.}
\end{figure}

\begin{figure}
\begin{subfigure}[b]{0.45\linewidth}
        \centering
    \includegraphics[scale=0.3]{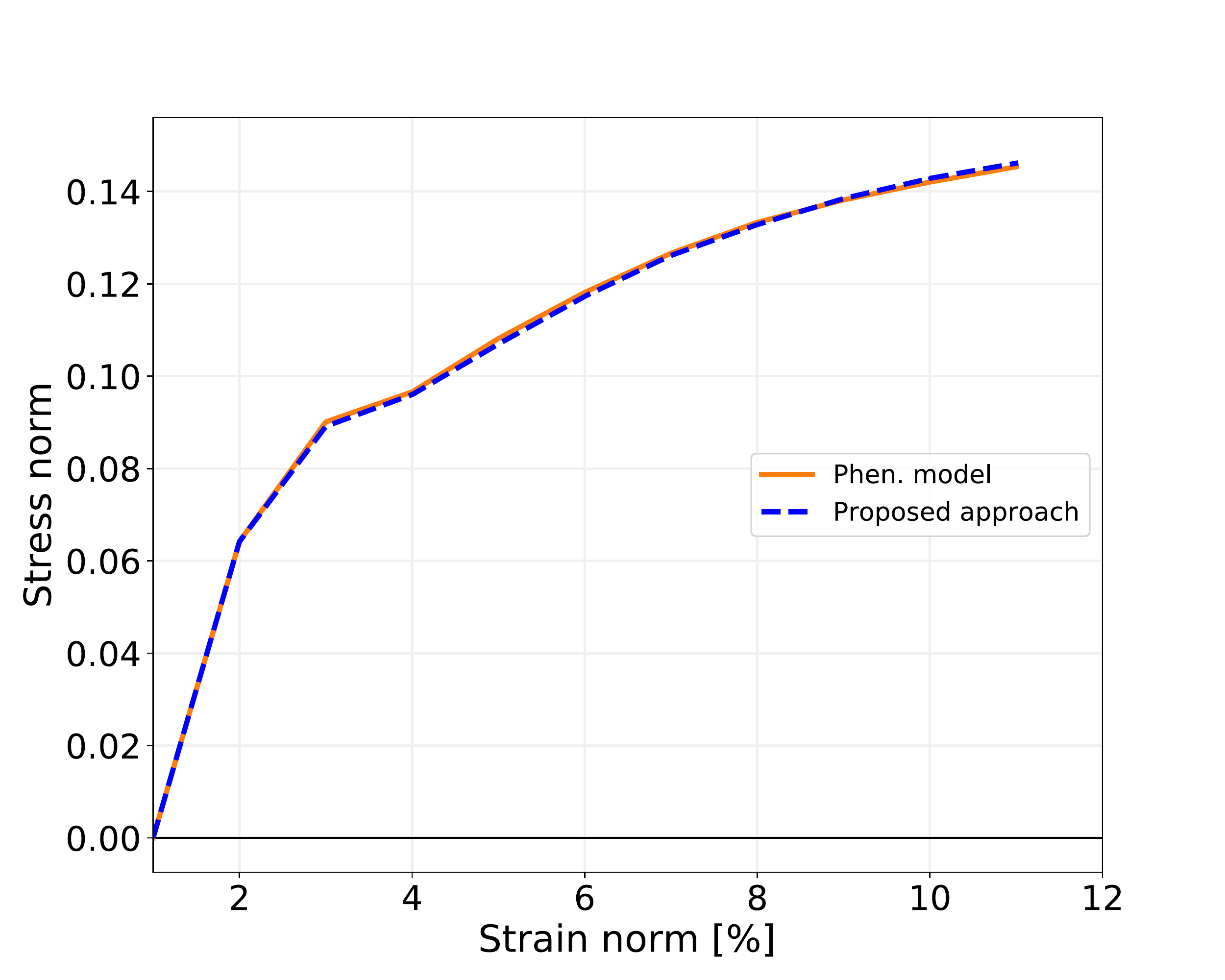}
    \caption{}\label{fig:cookMemAppLoadDisp}
\end{subfigure}
\begin{subfigure}[b]{0.45\linewidth}
        \centering
    \includegraphics[scale=0.3]{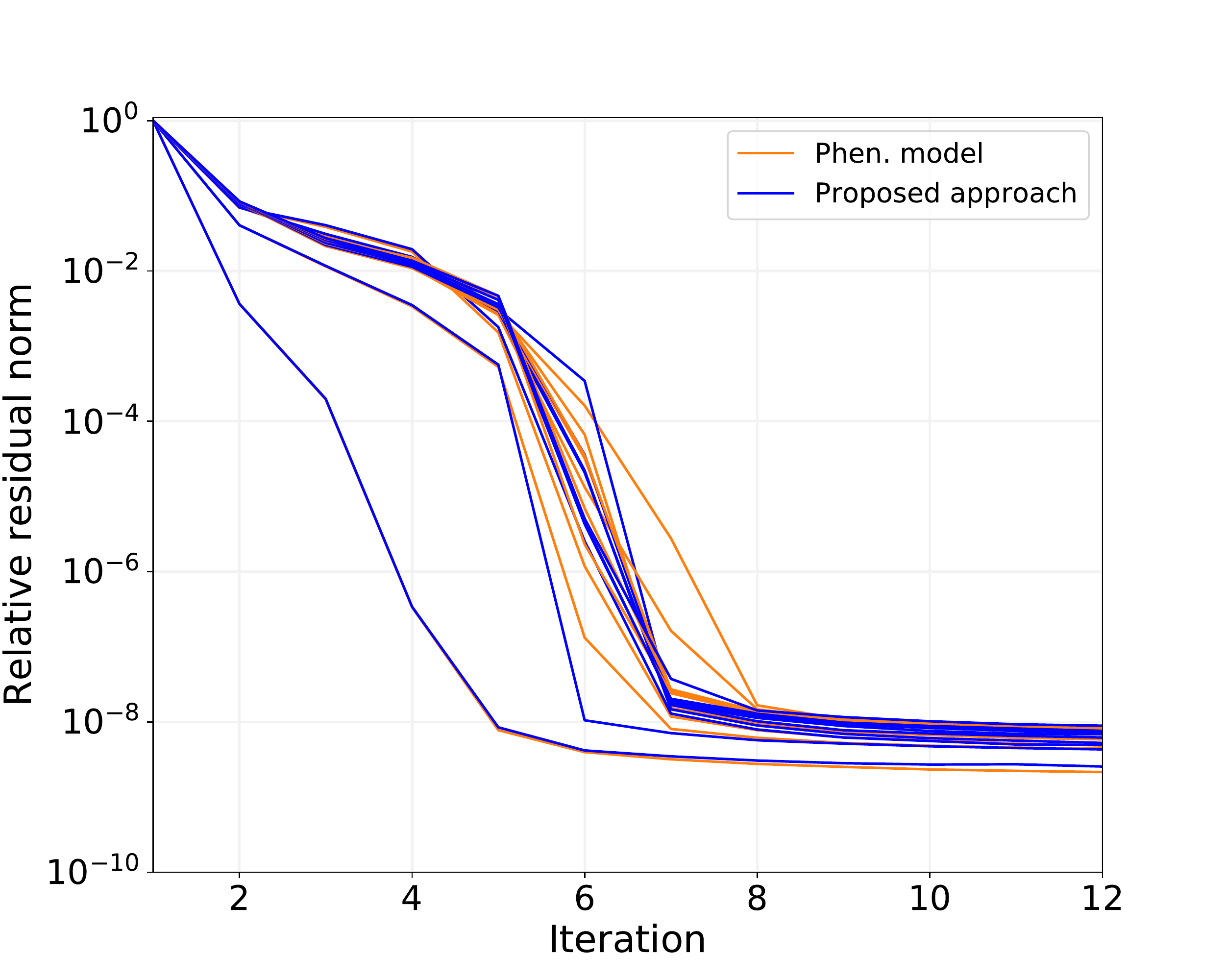}
    \caption{ }\label{fig:cookMemAppConv}
\end{subfigure}
\caption{Cook's membrane test results: (a) Averaged strain - averaged stress curve over the 10 loading increments, (b) Numerical convergence behaviors of the structural problem when using phenomenological and our proposed approach.}
\end{figure}

The second structural FEM benchmark, Cook's membrane, is
illustrated in Figure \ref{fig:cookMemApp}. A hexahedral structural member is pinned on the left and a vertical displacement of $u_{0} = 0.3 mm$ is applied on the right-hand side. 
Figures \ref{fig:cookMemAppModel} and \ref{fig:cookMemAppNN} show the displacement in $x$-direction of the ground truth and the surrogate material model.
Here again, it can be seen that the proposed framework is  able to accurately capture the true response even for complex loading conditions.
To highlight this, we compare the averaged strain-averaged stress curve over the 10-increment loading process of the structural problem, see  
Figure \ref{fig:cookMemAppLoadDisp}. We can see that the proposed framework has a similar stress response to the phenomenological material model.
Lastly, the global convergence behaviors of the two models over the number of iterations for the 10 load increments are plotted in 
Figure \ref{fig:cookMemAppConv} where no major differences can be seen.

\begin{figure}
    \centering
    \includegraphics[scale=0.3]{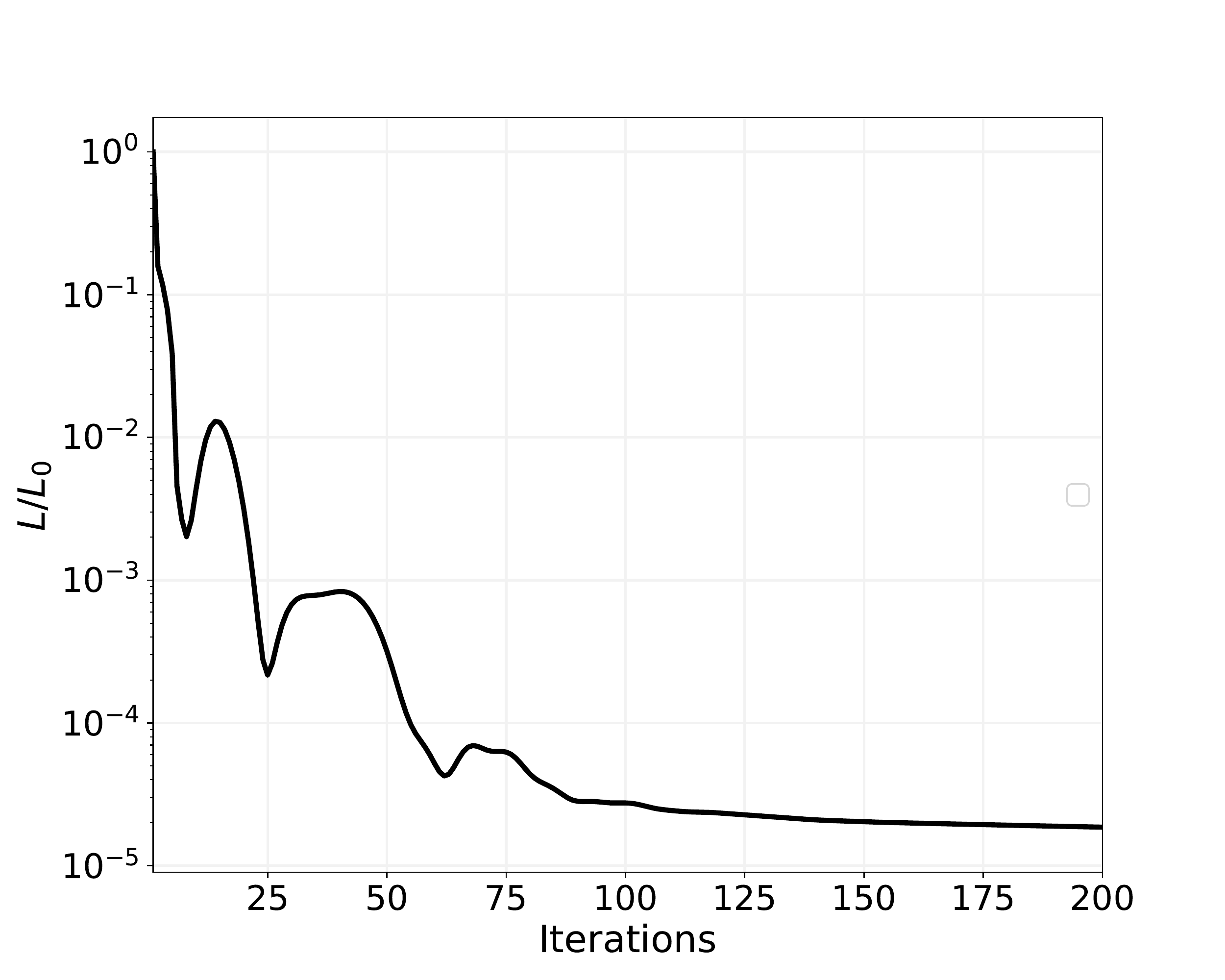}
    \caption{Fit to single NLK model with mixed hardening without constraining the neural networks. Training loss convergence over the training process}
    \label{fig:singleNLKLossnormalNN}
\end{figure}

\begin{figure}
\begin{subfigure}[b]{0.45\linewidth}
        \centering
    \includegraphics[scale=0.3]{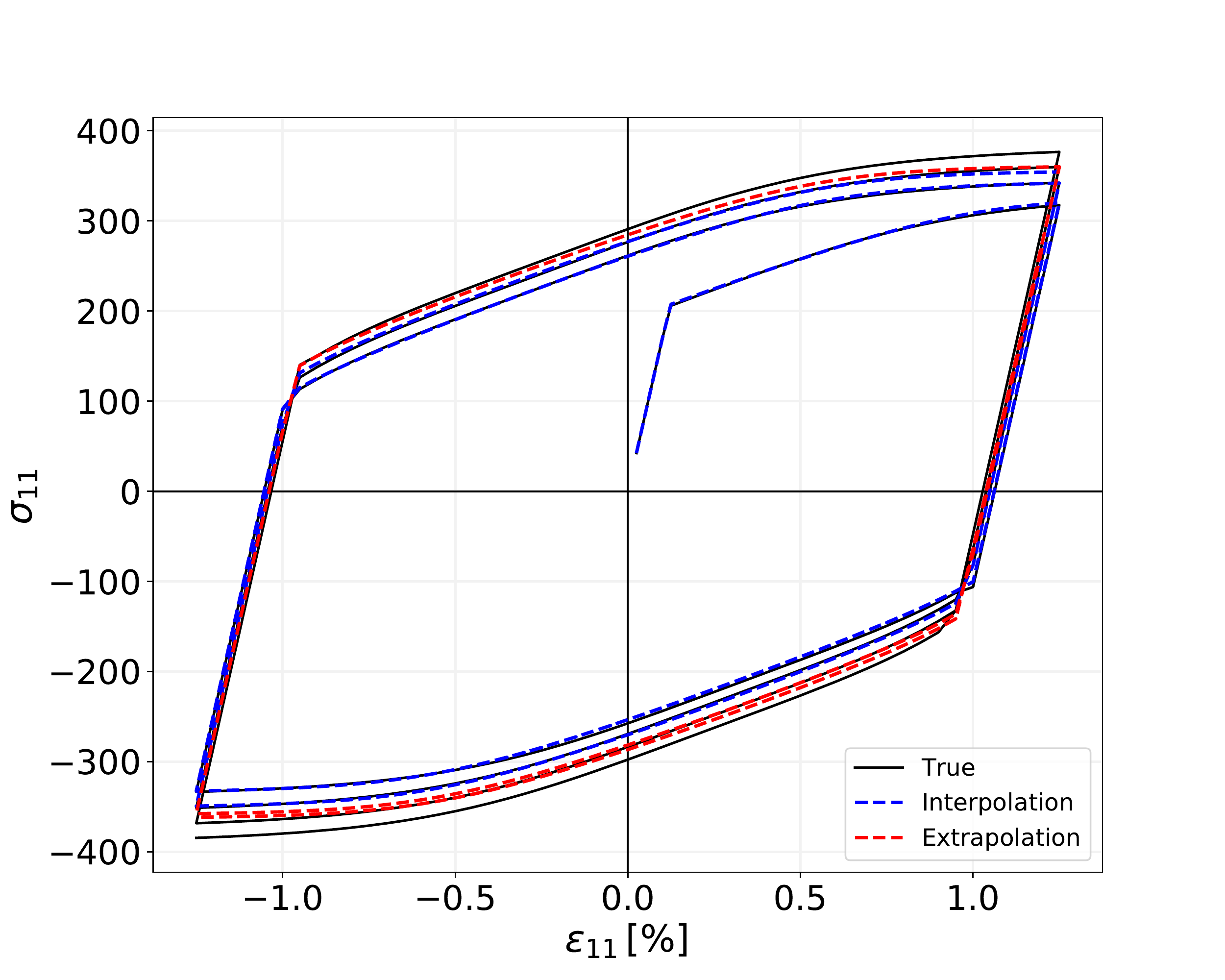}
    \caption{}\label{fig:singleNLKanormalNN}
\end{subfigure}
\begin{subfigure}[b]{0.45\linewidth}
        \centering
        \includegraphics[scale=0.3]{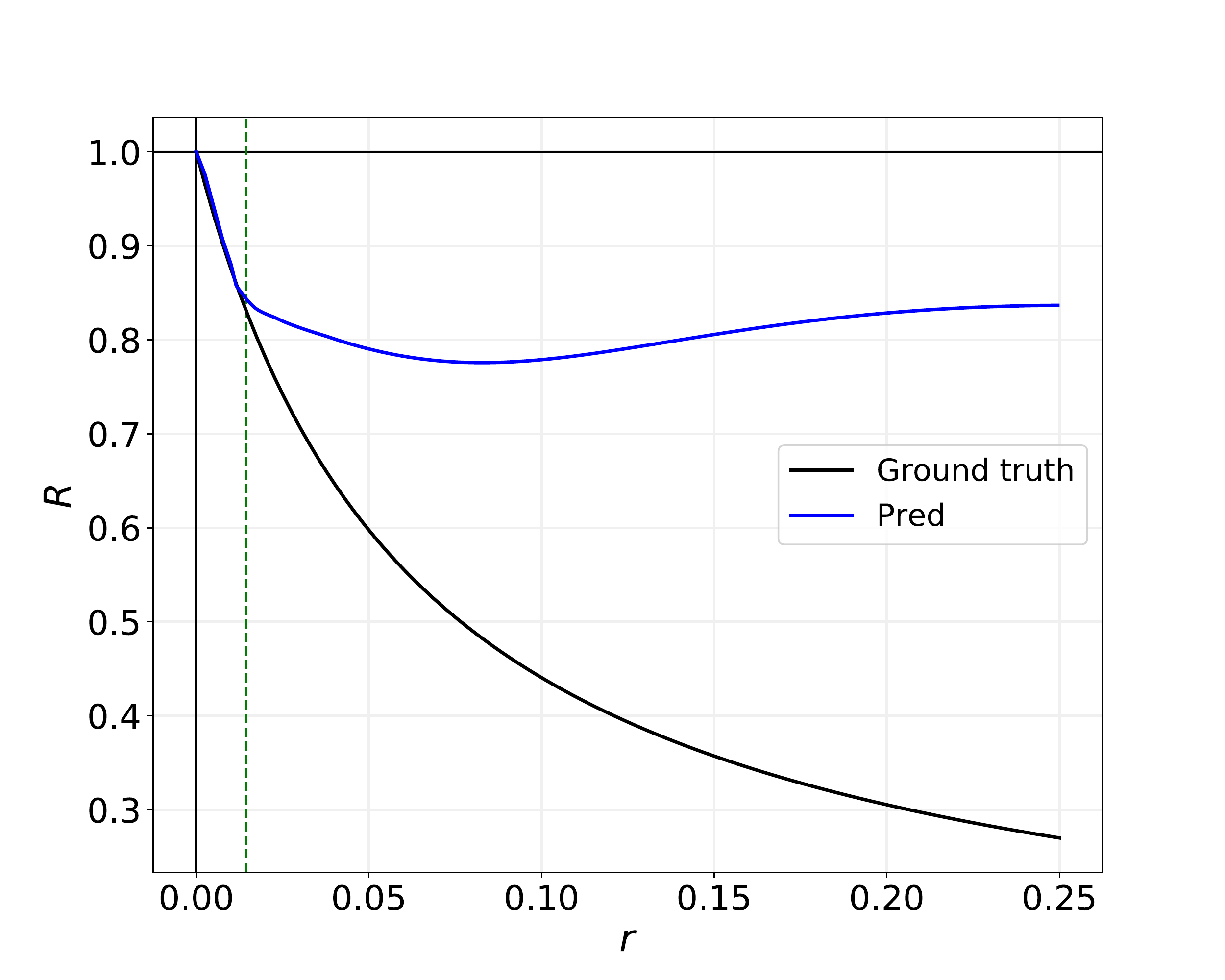}
    \caption{}\label{fig:singleNLKbnormalNN}
\end{subfigure}

\begin{subfigure}[b]{0.45\linewidth}
        \centering
    \includegraphics[scale=0.3]{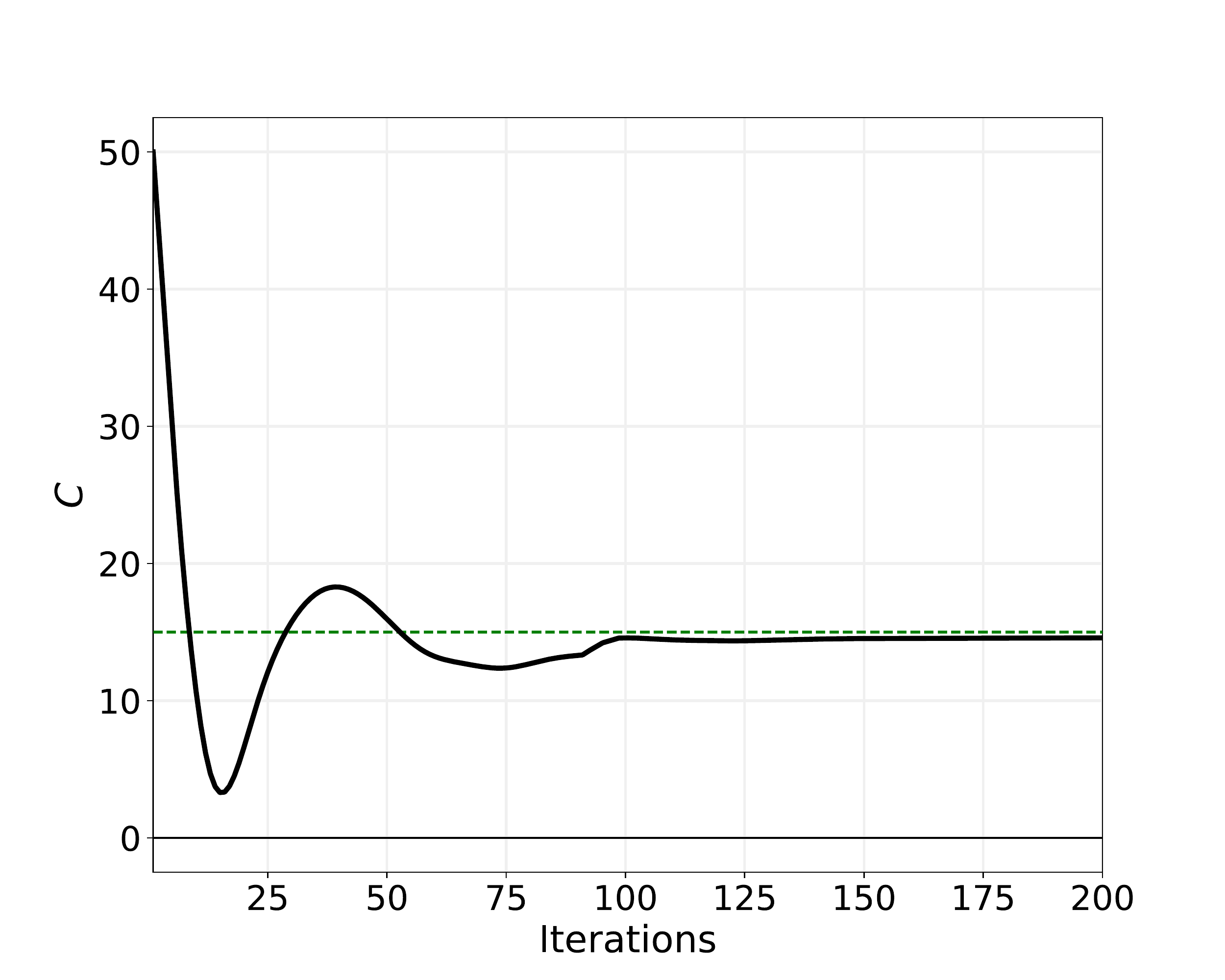}
    \caption{}\label{fig:singleNLKcnormalNN}
\end{subfigure}
\begin{subfigure}[b]{0.45\linewidth}
        \centering
    \includegraphics[scale=0.3]{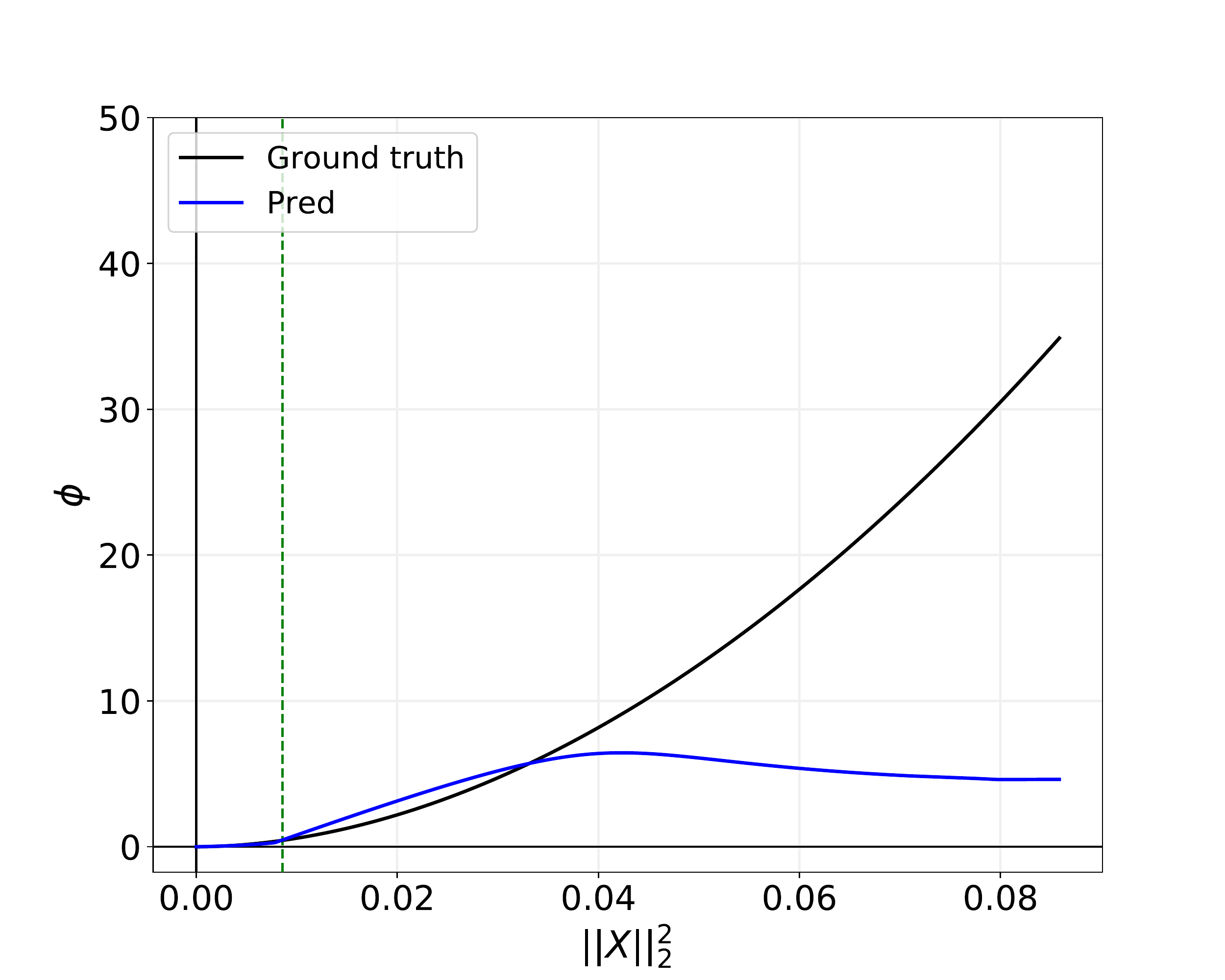}
    \caption{ }\label{fig:singleNLKdnormalNN}
\end{subfigure}
\caption{Fit to single NLK model with mixed hardening without constraining the neural networks. (a) Model fit (blue curve) as well as extrapolation to unseen data (red curve), (b) Predicted isotropic hardening function (ground truth is $R=0$). Green dotted line indicates maximum seen input data during training, (c) Evolution of the material parameter $C$ over the training process (ground truth value is $15$ as indicated by green line),  (d) Predicted nonlinear kinematic hardening function $\phi$. Green dotted line indicates maximum seen input data during training.}\label{fig:singleNLKnormalNN}
\end{figure}

\paragraph{Single NLK - without ML constraints}
In this last example involving the single NLK model, we investigate the effects that the constraints on the neural network models (convex, monotonically decreasing), c.f. page \pageref{sub:ParaConstraints}, have on the interpolation and extrapolation quality of the material model. In order to ensure initial convergence, we still assume that the neural networks conform to their respective initial conditions but we remove all other constraints. We train on the same dataset as in the previous example of Figure \ref{fig:singleNLK} where both isotropic and kinematic hardening underly the data. The training loss of this unconstrained model is shown in
Figure \ref{fig:singleNLKLossnormalNN}. It can be seen that the error reduces at a similar rate to the constrained problem (c.f. Figure \ref{fig:singleNLKLoss}) without any major differences in the final relative error value. This can also be observed in the interpolation domain of the  
uniaxial response of the trained model which is shown in Figure \ref{fig:singleNLKanormalNN}. Here, the blue line which indicates the area where training data is available is almost perfectly aligned with the ground truth. Crucially, however, the extrapolation quality of the model is severely lacking compared to the constraint version. The reason for this effect can be seen when plotting the predicted hardening functions as well as the material parameter $C$, see Figures \ref{fig:singleNLKbnormalNN}, \ref{fig:singleNLKcnormalNN}
and \ref{fig:singleNLKdnormalNN}. We can see that, even though $C$ is predicted accurately at the end of the training process (Figure \ref{fig:singleNLKcnormalNN}), both the isotropic and kinematic hardening functions appear to initially coincide with the reference model but start to strongly deviate from it after the models start extrapolating (indicated by the vertical green line). 
In particular, the nonlinear kinematic hardening function is no longer convex nor is the isotropic hardening function monotonically decreasing. This indicates that the constraints on the ML tools are necessary for good generalization capabilities of these trained models.

\subsubsection{Multi NLK-model}
Next, we check the performance of the framework on a known material model that is outside the range of the assumptions of the proposed framework. For this test we utilize the nonlinear kinematic hardening model taken from Chaboche \cite{chaboche1991some} which relies on the superposition of multiple backstresses, increasing the number of internal variables by more than a factor of two in comparison to the single NLK model that was discussed so far.
We consider a classical von Mises yield function defined by
\begin{equation}
f = J(\bm{\sigma} - \bm{X}) - \sigma_{y} - R \leq 0    
\end{equation}
where $J$ indicates the von Mises equivalent stress. The backstress $\bm{X}$ is now defined by
\begin{equation}
    \bm{X} = \sum_{i} \bm{X}_{i}
\end{equation}
with 
\begin{equation}
    \dot{\bm{X}}_{i} = \frac{2}{3} C_{i} \dot{\bm{\epsilon}}^{p} - \frac{\gamma_{i}^{2}}{C_{i}} [J(\bm{X}_{i})]^{m-1} \bm{X}_{i} \dot{r}.
\end{equation}
Here $C_{i}$ and $\gamma_{i}$ are material parameters.
The isotropic hardening evolution is defined by
\begin{equation}
\begin{aligned}
        \dot{R} &= b [Q(q) - R ] \dot{r}, \qquad \text{where}, \qquad
        Q(q) &= Q_{M} + (Q_{0}-Q_{M}) e^{\mu \Delta \epsilon_{p}}.
\end{aligned}
\end{equation}
The model is fully defined by the material parameters $Q_{M}$, $Q_{0}$ and $\mu$. Following \cite{chaboche1991some}, we normalize the nonlinear kinematic hardening parameters by
\begin{equation}
    \gamma_{i} = \frac{\gamma_{i}}{\tau(R)}, \qquad \tau(R) = 1 + \frac{R}{\sigma_{y}}.
\end{equation}
The material parameter values used here are given in Table \ref{tab:multiNLKParam}.

\begin{table}
\begin{center}
\begin{tabular}{||c c c c c c c c c c c c||} 
 \hline
 $C_{1}$ & $\gamma_{1}$ & $C_{2}$ & $\gamma_{2}$ & $C_{3}$ & $\gamma_{3}$ & $b$ & $Q_{M}$ & $Q_{0}$ & $\mu$ & $k$ & $m$\\ [0.5ex] 
 \hline\hline
 80,000 & 800 & 300,000 & 10,000 & 1,000 & 7 & 8 & 300 & 14 & 10 & 100 & 2\\ [1ex] 
 \hline
\end{tabular}
\end{center}
\caption{Non-dimensional material parameters for Multi NLK-model.}\label{tab:multiNLKParam}
\end{table}

\begin{figure}
\begin{subfigure}[b]{0.45\linewidth}
        \centering
    \includegraphics[scale=0.3]{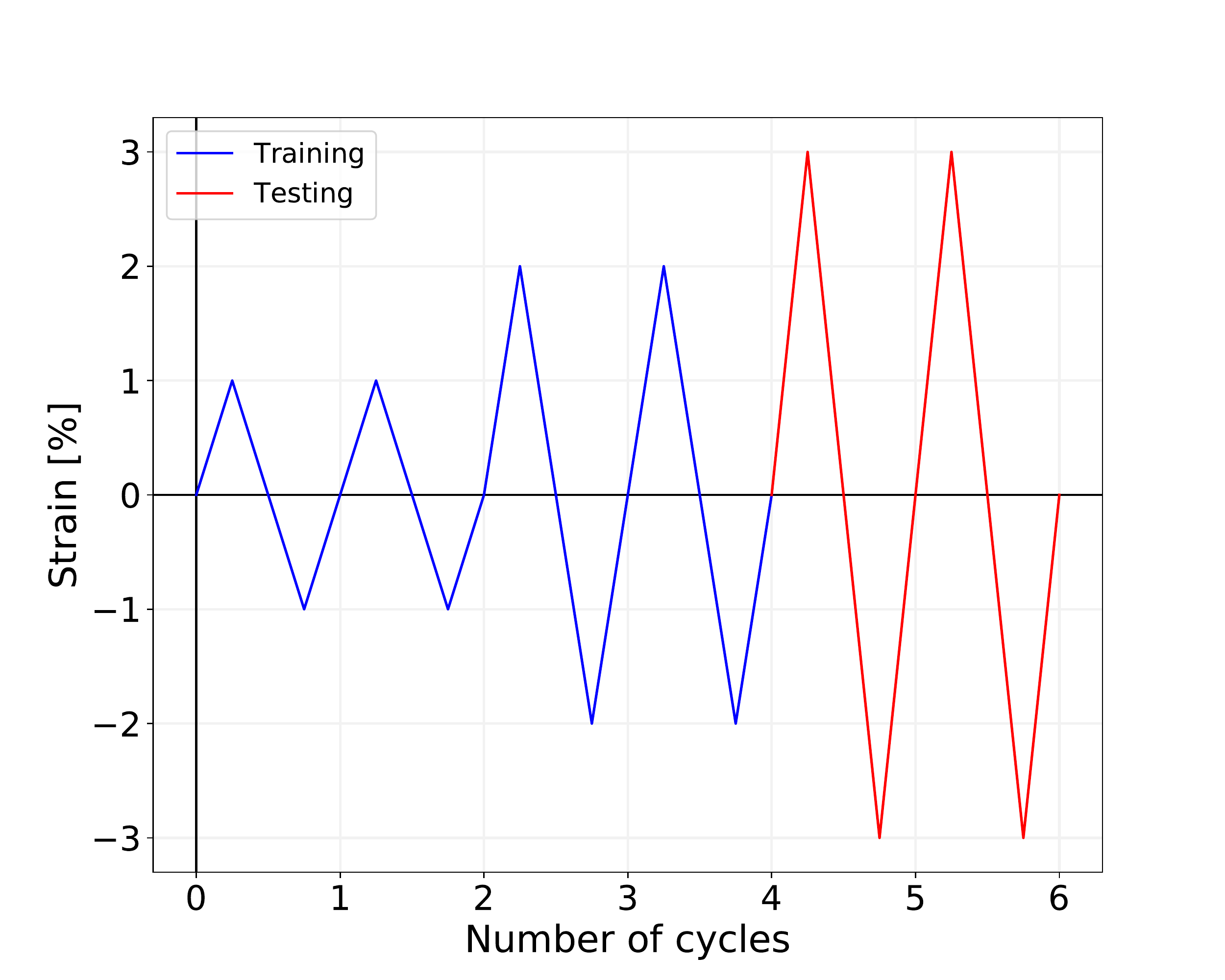}
    \caption{ }\label{fig:multiNLKa}
\end{subfigure}
\begin{subfigure}[b]{0.45\linewidth}
        \centering
    \includegraphics[scale=0.3]{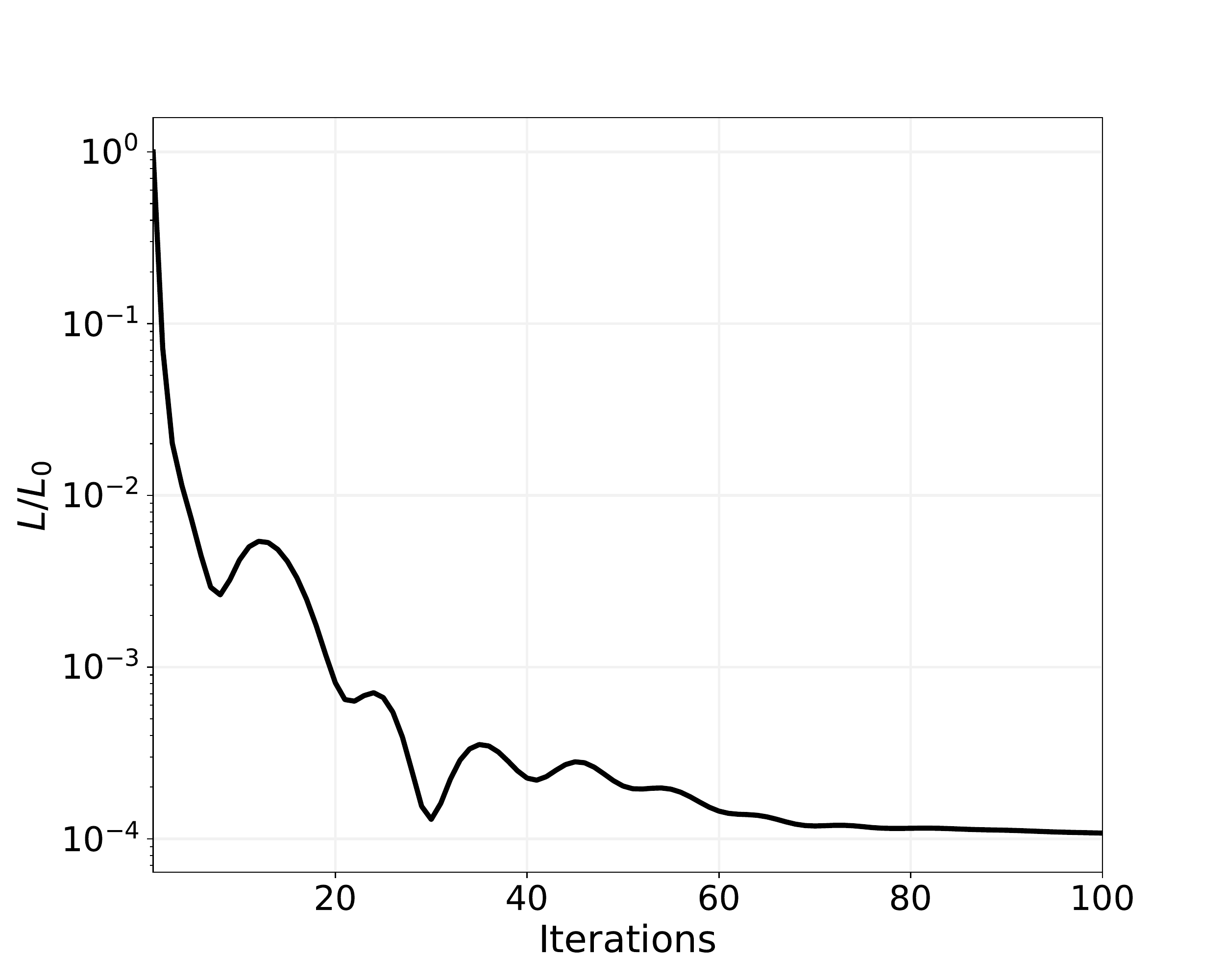}
    \caption{}\label{fig:multiNLKb}
\end{subfigure}

\begin{subfigure}[b]{1.0\linewidth}
        \centering
    \includegraphics[scale=0.3]{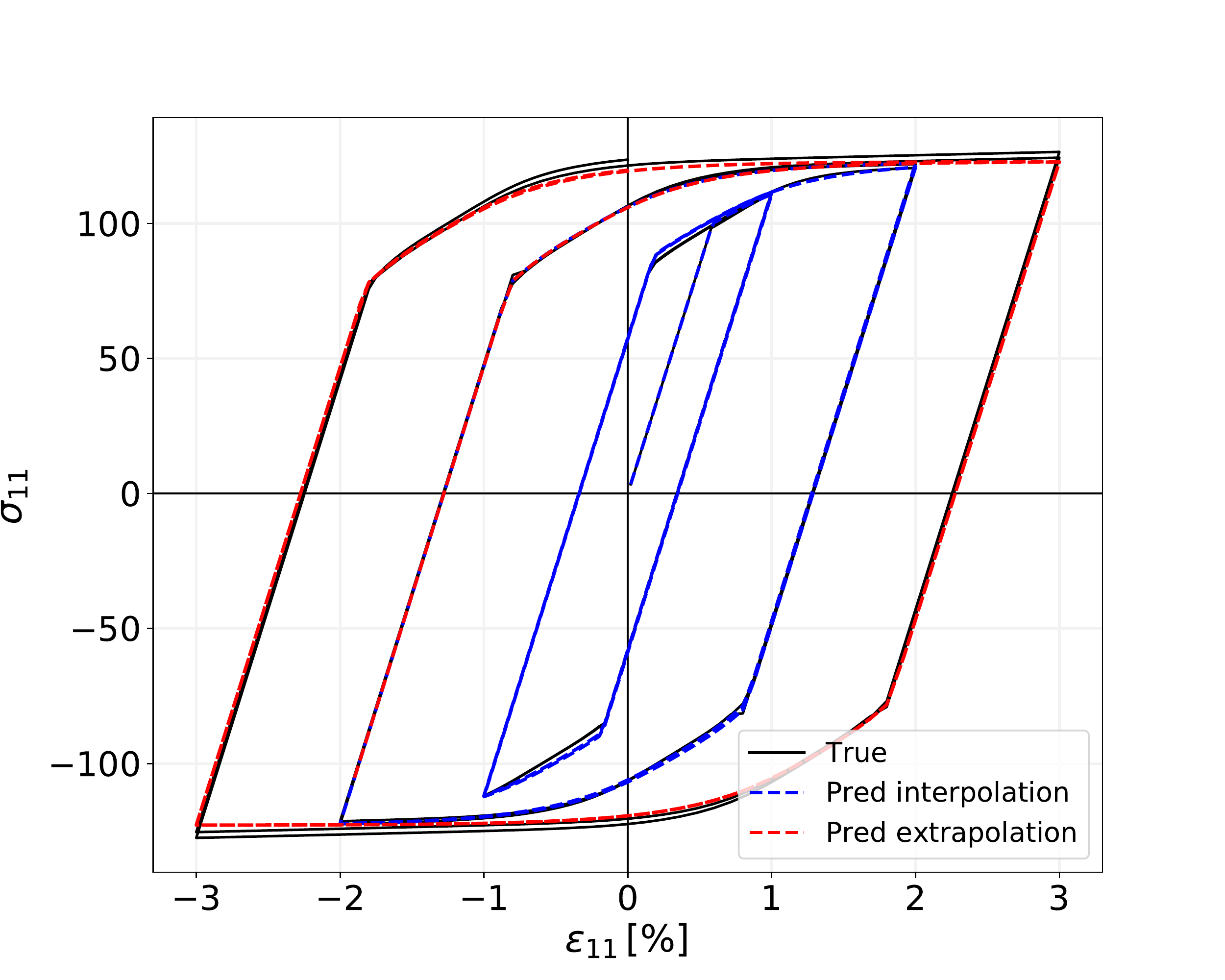}
    \caption{}\label{fig:multiNLKc}
\end{subfigure}

\caption{Fit to multi NLK model with mixed hardening. (a) Ascending applied uniaxial loading condition, blue curve indicates training data loading and red curve represents the extrapolation domain, (b) Training loss convergence over the training process, (c) Model fit (blue curve) as well as extrapolation to unseen data (red curve)}
\end{figure}

The training and testing data are generated from the uniaxial cyclic loading condition specified in
Figure \ref{fig:multiNLKa}.
Figure \ref{fig:multiNLKb} shows the convergence of the training loss over the duration of the training process ($100$ iterations) which indicates a significant reduction of the initial error.
Figure \ref{fig:multiNLKc} depicts the stress-strain response of the surrogate on the training data (blue curve) as well as its extrapolation prediction to unseen data (red curve).
We note that even though the ground truth model, with three backstresses, is not explicitly part of the modeling domain (we consider only one backstress evolution) the final fit as well as the extrapolation behavior is closely aligned with the ground truth.
This alludes to the fact that the model is expressive enough to fit hardening behaviors that are outside its explicit modeling range.

\subsection{Fitting to experimental data}
So far the proposed framework has only been trained and tested using synthetic data coming from known phenomenonlogical constitutive models. Next, we test its performance on experimental datasets. For this, we rely on data consisting of cyclic loading hysteresis loops obtained using WebPlotDigitizer \cite{rohatgi2017webplotdigitizer} directly from the plots that were published.
We explore two cases involving two different types of steel.

\begin{figure}
\begin{subfigure}[b]{0.45\linewidth}
        \centering
    \includegraphics[scale=0.3]{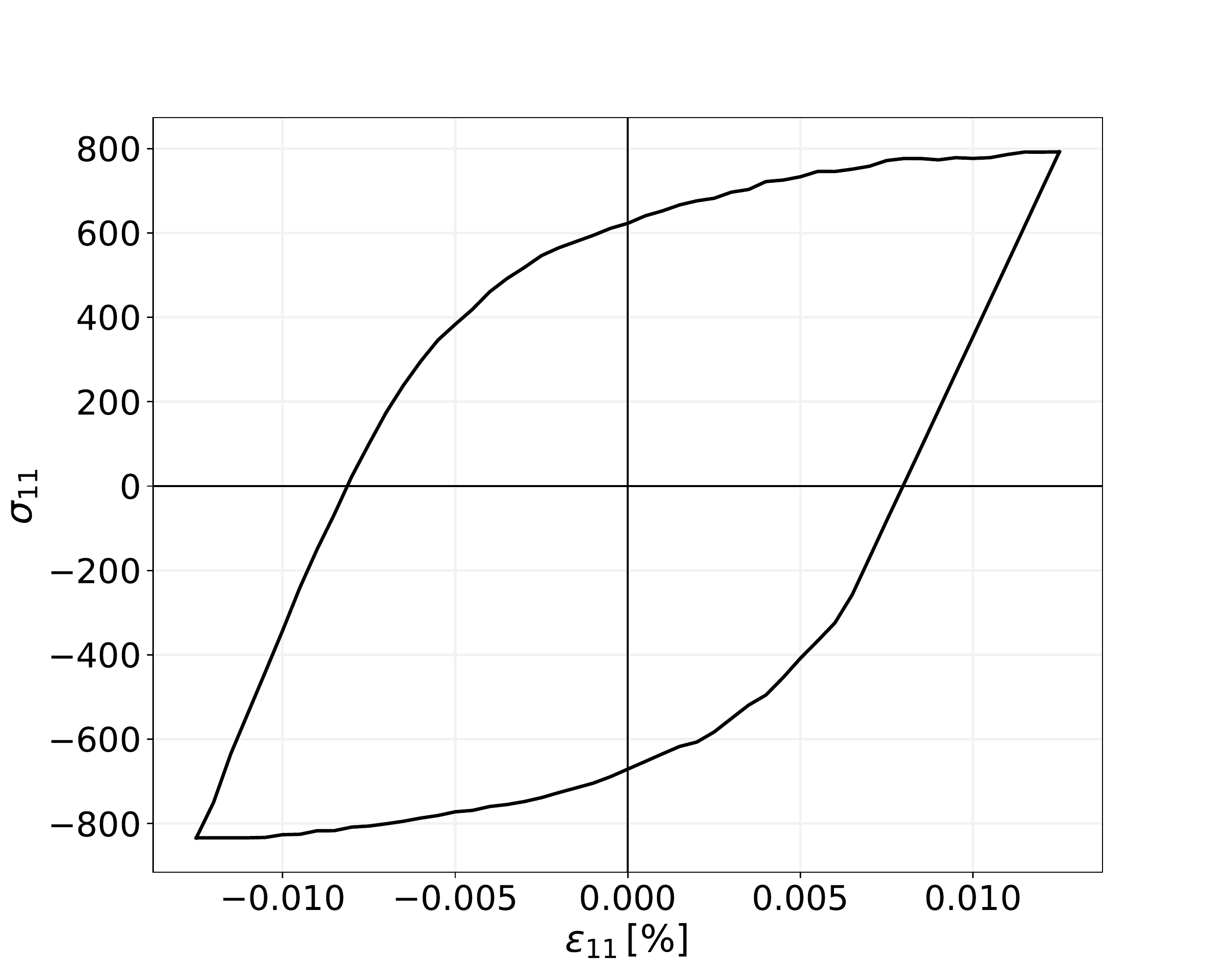}
    \caption{}\label{fig:DataStainless}
\end{subfigure}
\begin{subfigure}[b]{0.45\linewidth}
        \centering
    \includegraphics[scale=0.3]{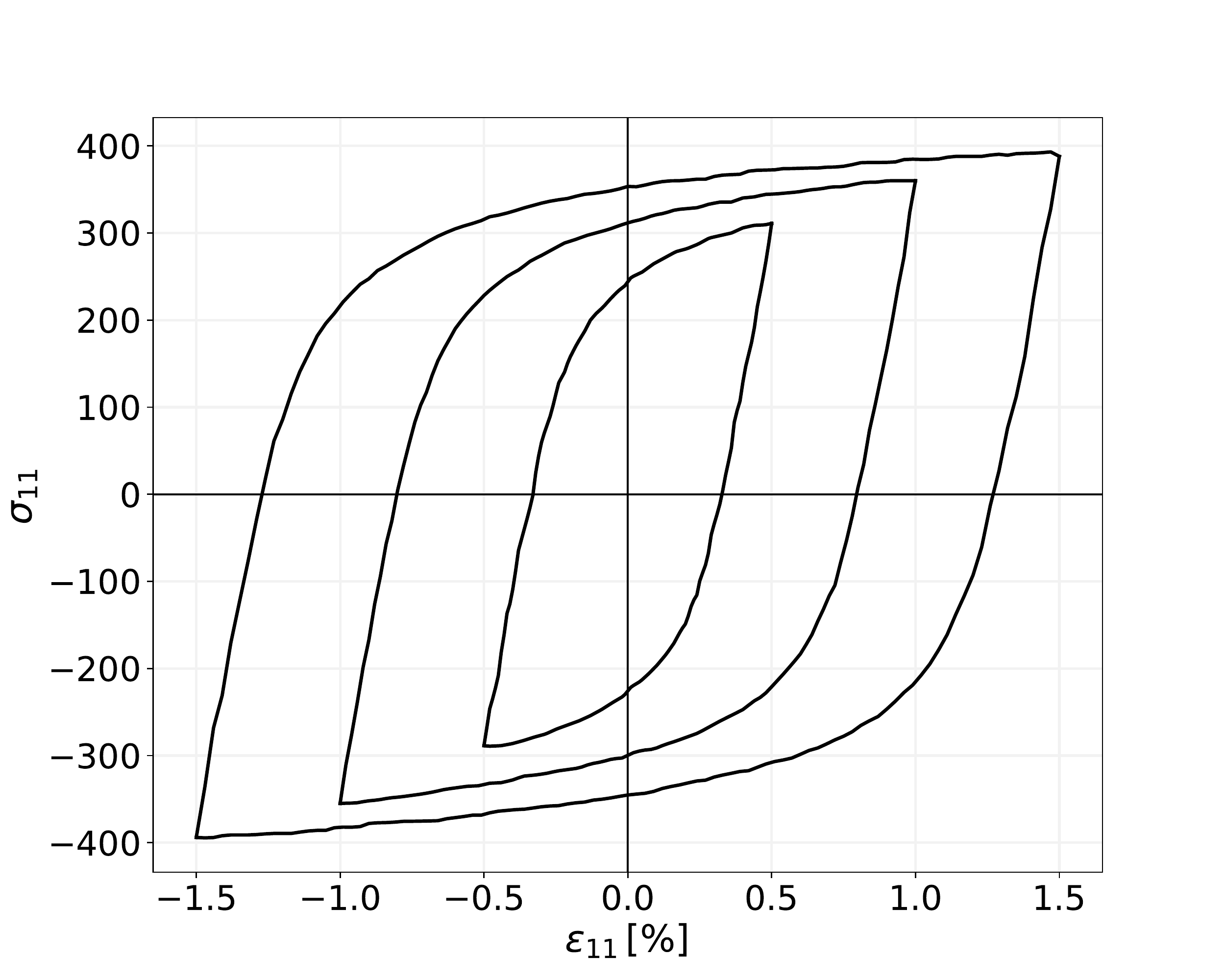}
    \caption{ }\label{fig:DataCarbon}
\end{subfigure}

\caption{Hysteresis loops from experimental data. (a) Stainless steel dataset, c.f. Fig. 6(a) of \cite{wang2021study}, (b) Structural carbon steel dataset, c.f. Fig. 2(a) of \cite{chen2014cyclic}.}
    
\end{figure}

\begin{figure}
\begin{subfigure}[b]{0.45\linewidth}
        \centering
    \includegraphics[scale=0.3]{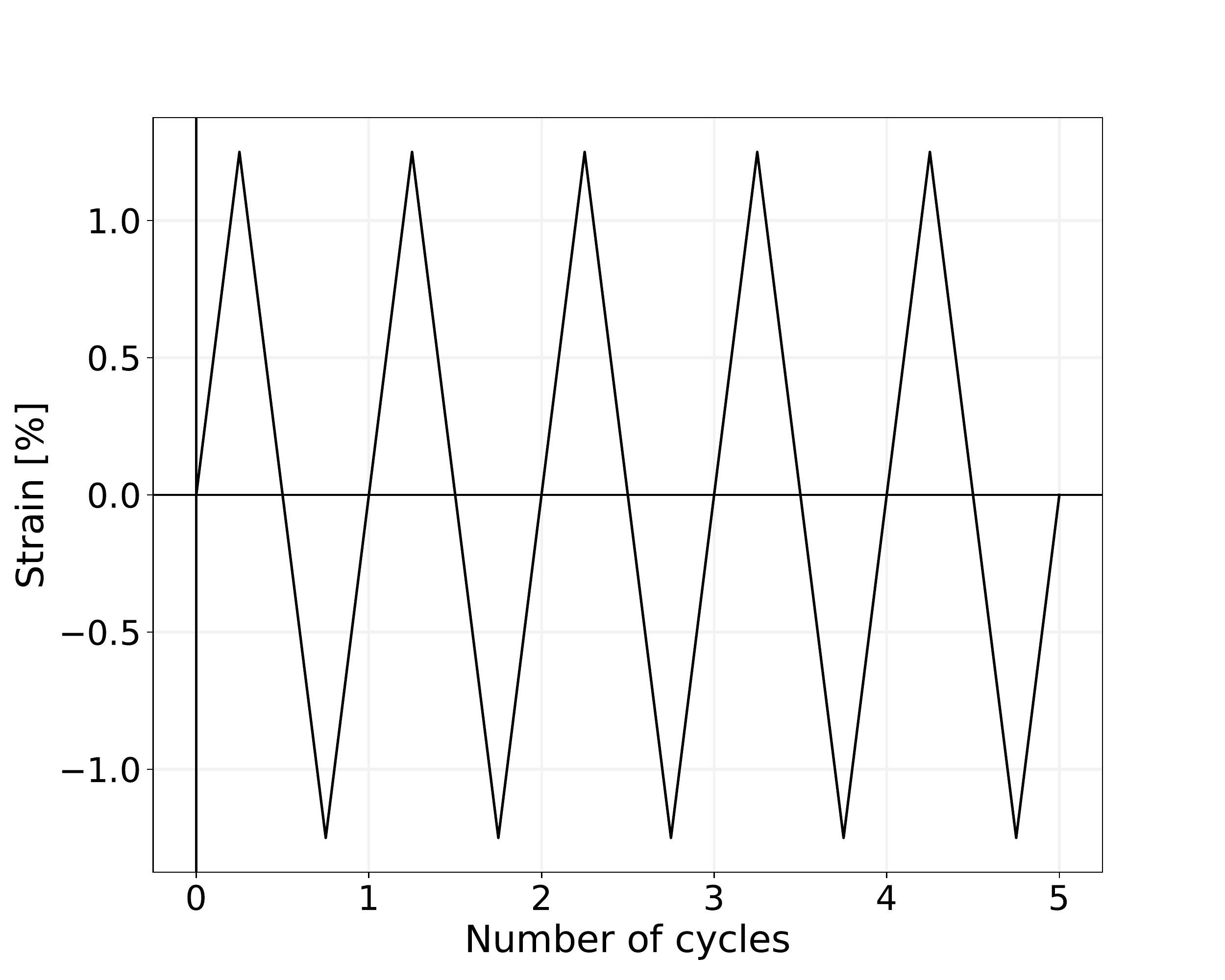}
    \caption{}\label{fig:expDataLoadStainless}
\end{subfigure}
\begin{subfigure}[b]{0.45\linewidth}
        \centering
    \includegraphics[scale=0.3]{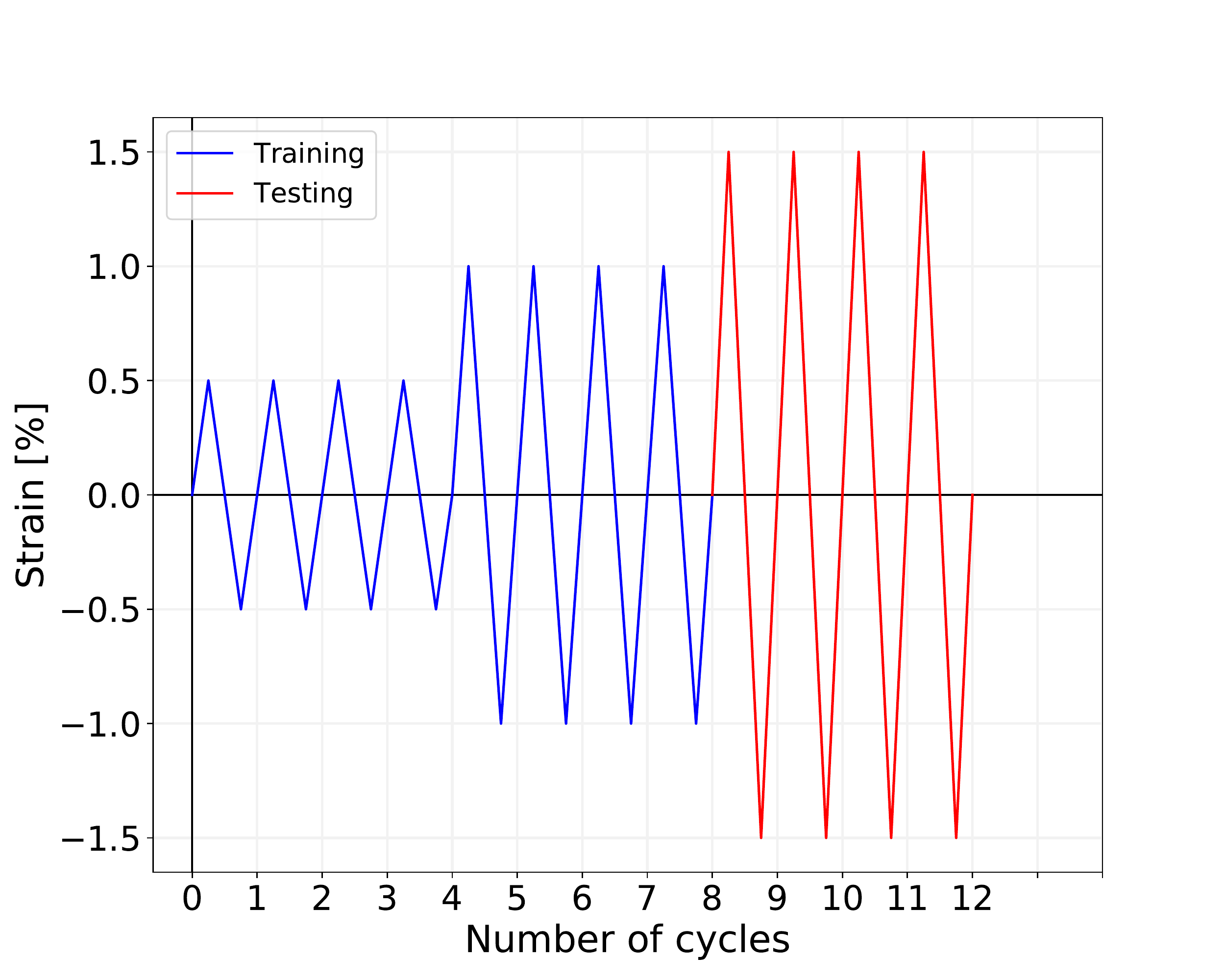}
    \caption{}\label{fig:expDataLoadCarbon}
\end{subfigure}
\caption{Loading curves to reach hysteresis loops, (a) Stainless steel dataset, (b) Carbon steel dataset}
\end{figure}

\subsubsection{Sorbite stainless steel}
\begin{figure}
\begin{subfigure}[b]{0.45\linewidth}
        \centering
   \includegraphics[scale=0.3]{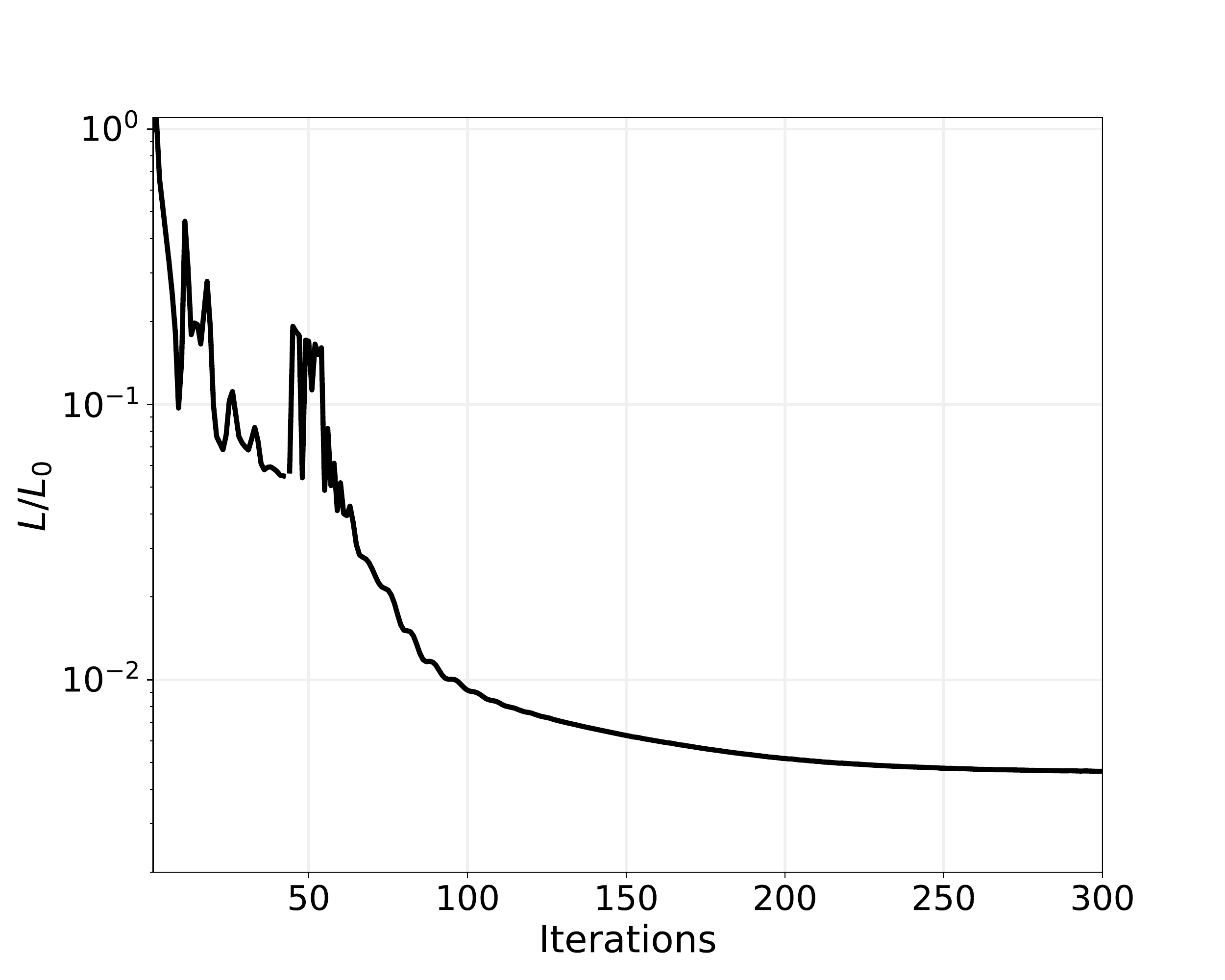}
    \caption{}\label{fig:DataStainlessLoss}
\end{subfigure}
\begin{subfigure}[b]{0.45\linewidth}
        \centering
    \includegraphics[scale=0.3]{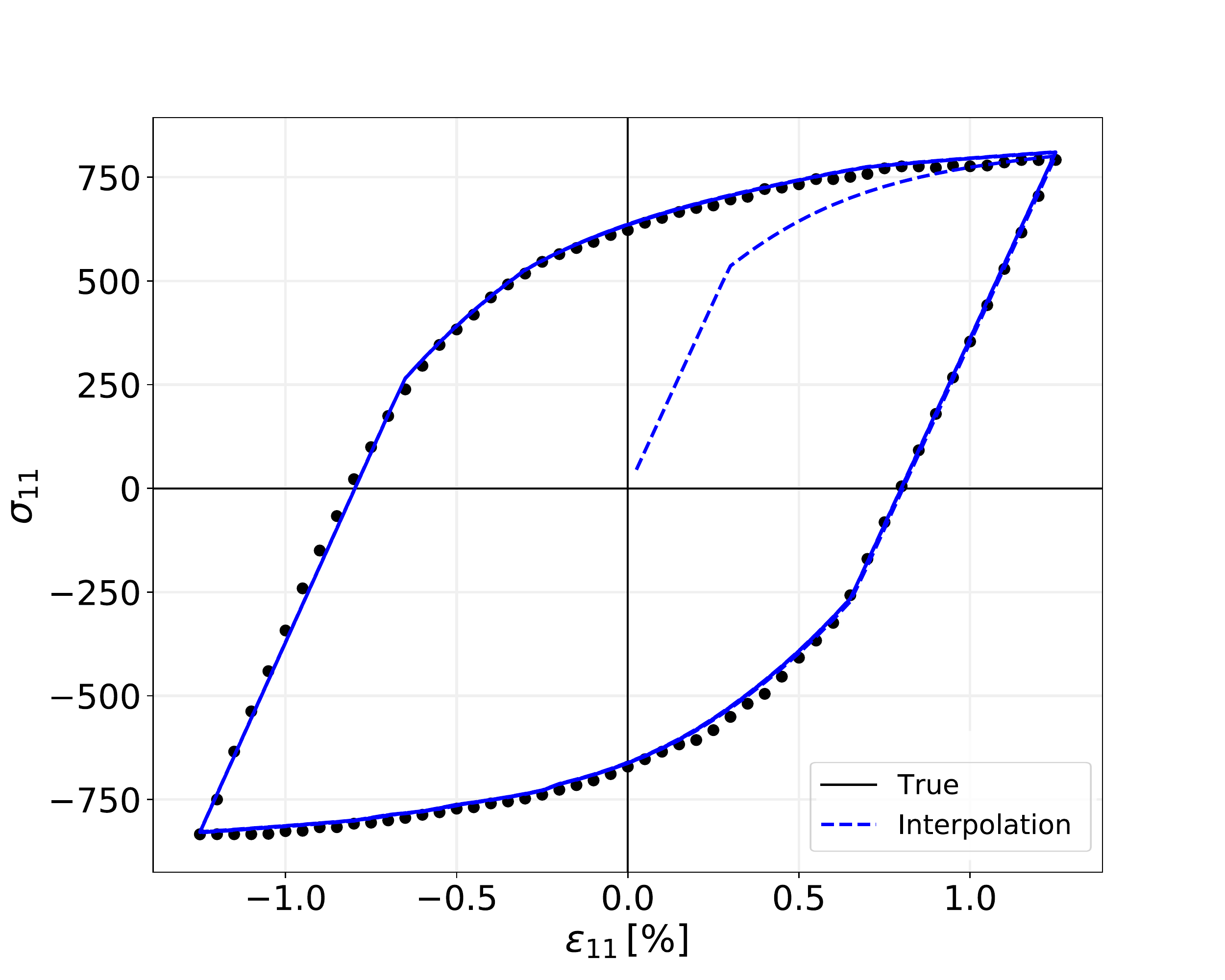}
    \caption{}\label{fig:DataStainlessFit}
\end{subfigure}
\caption{Experimental data. Sorbite stainless steel. (a) Training loss convergence over the training process, (b) Comparison of model fit (blue curve) and true response.}
\end{figure}

The first test involves an approximate hysteresis loop for 
stainless steel \cite{wang2021study} which is shown in
Figure \ref{fig:DataStainless}.
The hysteresis behavior is obtained through the  five loading cycles to a strain of $1.25\%$, see Figure \ref{fig:expDataLoadStainless}.
The material parameter for the linear elastic response ($E = 212,000$, $\nu=0.3$) as well as the von Mises yield function ($\sigma_{y} = 412$) were provided by the authors \cite{wang2021study}.
The training loss over the duration of the training process is shown in Figure \ref{fig:DataStainlessLoss}. We note that the initial training loss can be significantly reduced.
The hysteresis loop obtained with the surrogate material model is compared to the true experimental response in Figure \ref{fig:DataStainlessFit}
We can see that the proposed framework is able to accurately capture the hardening behavior that causes the hysteresis loop even without needing to a~priori specify that no isotropic hardening is involved.

\subsubsection{Structural carbon steel }
\begin{figure}
\begin{subfigure}[b]{0.45\linewidth}
        \centering
        \includegraphics[scale=0.3]{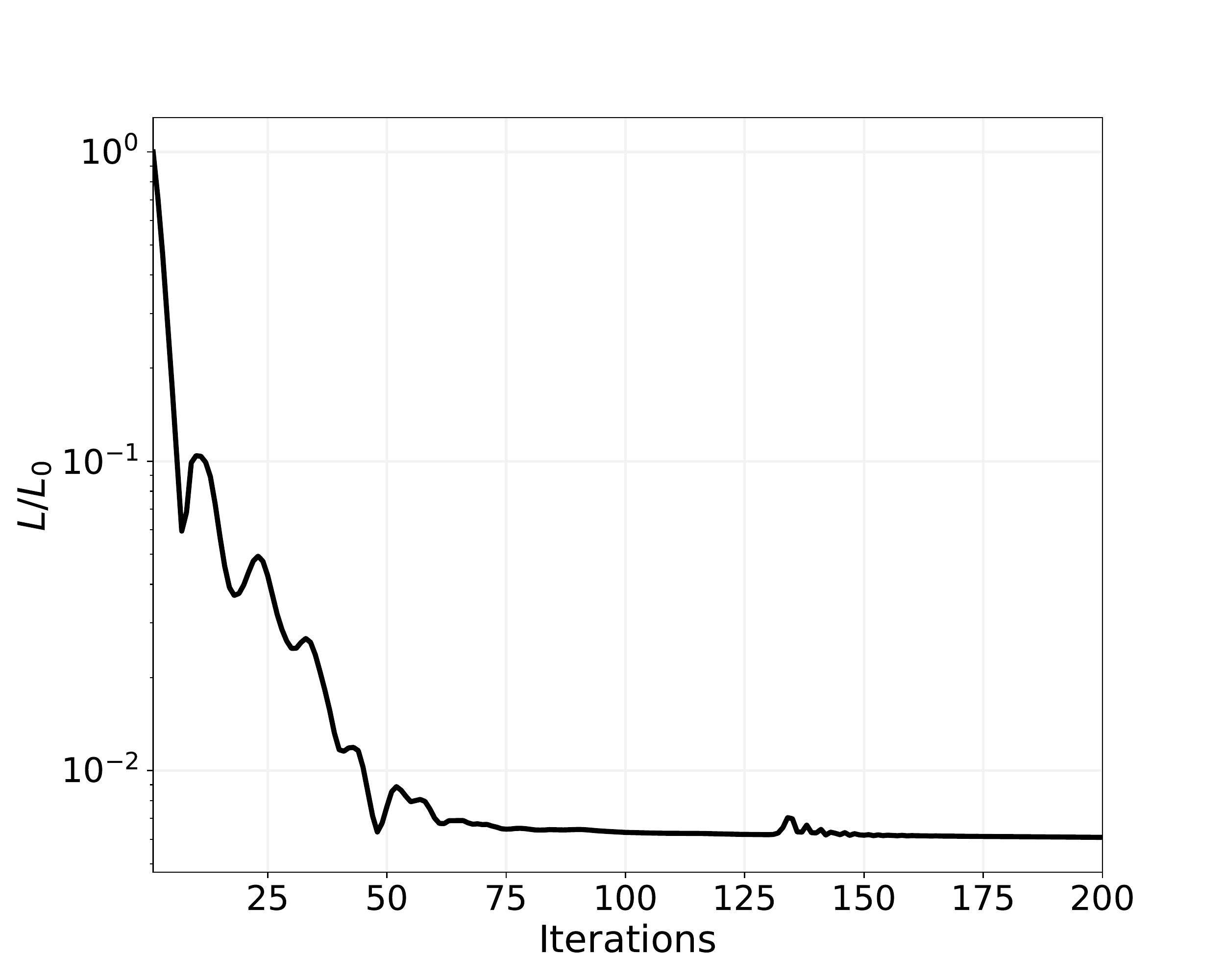}
    \caption{}\label{fig:DataCarbonLoss}
\end{subfigure}
\begin{subfigure}[b]{0.45\linewidth}
        \centering
    \includegraphics[scale=0.3]{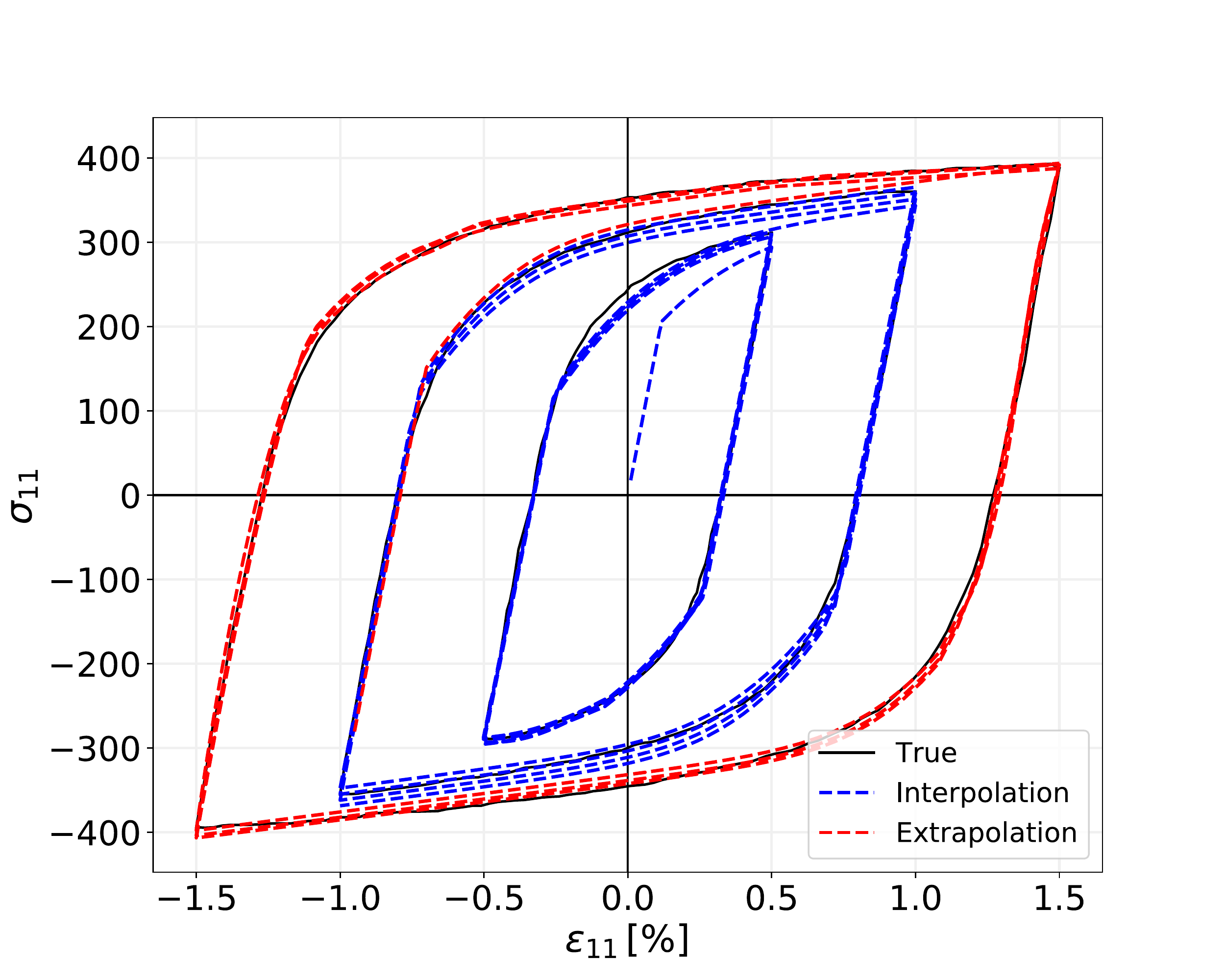}
    \caption{}\label{fig:DataCarbonFit}
\end{subfigure}
\caption{Experimental data. Structural carbon steel. (a) Training loss convergence over the training process, (b) Model fit (blue curve) as well as extrapolation to unseen data (red curve).}
\end{figure}

Second, we look at the case of ascending hysteresis loops observed in carbon steel. The training data is shown in
Figure \ref{fig:DataCarbon} which is the result of the ascending cyclic loading conditions displayed in
Figure \ref{fig:expDataLoadCarbon}.
The material parameter for the linear elastic response ($E = 200,000$, $\nu=0.27$) as well as the von Mises yield function ($\sigma_{y} = 210$) where taken from the authors \cite{wang2021study}.
Figure \ref{fig:DataCarbonLoss} plots the training loss where a sufficient loss convergence can be observed.
The model fit on the training data as well as the extrapolation performance of the surrogate model can be seen in Figure \ref{fig:DataCarbonFit}. All three loops appear to be accurately fit meaning that the proposed framework was able to reliably capture the hardening behavior that underlies the data without needing to specify a functional form.

\section{Conclusion and Outlook}\label{sec::5}
The majority of  proposed data-driven techniques for path-dependent material responses require a substantial amount of data which also needs to span the full stress space.  Hence, these approaches are in particular not suitable for limited-data regimes which are commonly encountered when data is obtained from classical experiments such as uniaxial testing.
In this work we explore a modular elastoplastic framework that is able to establish material models with a variable amount of information and data including the most limiting data cases which involve only uniaxial experimental tests. Our modular elastoplastic framework can transition between analytical and machine learning-based components due to its modularity. Additionally, we utilize physical constraints from thermodynamics and develop machine learning tools that strongly enforce these constraints. This is a feature that is uniquely responsible for the development of expressive data-driven models that can work in the limited-data regime and at the same time generalize efficiently.

Our approach is based on a two potential formulation involving isotropic and nonlinear kinematic hardening. Conditions for thermodynamic consistency are discussed and neural network formulations that model the hardening behavior and which implicitly fulfill the required thermodynamic constraints are introduced.
The proposed approach is tested both on
synthetic uniaxial data as well as experimental data from cyclic loading tests. We found that hardening representations from our framework allow for reliable extrapolation far beyond the training domain which is commonly infeasible in data-driven models for plasticity.
FEM results prove that the proposed framework is also sufficiently accurate and trustworthy, by probing complex loading conditions (beyond the uniaxial training data) in structural problems. Additionally, it is also shown that our proposed framework leads to a convergence behavior similar to that of when phenomenological models are utilized in the same FEM setting.

Next, we want to explore different data scenarios, for example, data obtained from numerical homogenization microscale simulations,  to take full advantage of the modularity of our approach where the elastic response, the yield function as well as the hardening behavior can all come either from phenomenological or from data-driven modeling frameworks.
This also involves building efficient data generation processes that allow for training reliable surrogate models that can work with the lowest amount of necessary data. Here,
adaptive sampling techniques \cite{fuhg2020state} that automatically find the next best possible data point based on the currently available information might be worth exploring.

\section*{Acknowledgements}
NB gratefully acknowledges support by the Air Force Office of Scientific Research under award number FA9550-22-1-0075. This work was partially supported by the Advanced Simulation and Computing (ASC) program at Sandia National Laboratories through investments in the Advanced Machine Learning Initiative (AMLI). Sandia National Laboratories is  a multi-mission laboratory managed and operated by National Technology \& Engineering Solutions of Sandia, LLC, a wholly owned subsidiary of Honeywell International Inc., for the U.S. Department of Energy's National Nuclear Security Administration under contract DE-NA0003525.

\clearpage
\appendix
\section{Appendix}
\subsection{Newton-Raphson vectors and matrices}\label{subsec:NewtonRaphsonVal}
For eq. \eqref{eq::NewtonRaphson} we need to define 
\begin{equation}
    \bm{x}^{v} = \begin{bmatrix}
    \epsilon_{11, n+1}^{e, v} \\
    \vdots\\
    \epsilon_{33, n+1}^{e, v} \\
    X_{11, n+1}^{v} \\
    \vdots\\
    X_{33, n+1}^{v} \\
    r_{n+1}^{v} \\
    R_{n+1}^{v} \\
    \Delta \lambda^{v}
    \end{bmatrix}, \qquad     \bm{F}(\bm{x}^{v}) = \begin{bmatrix}
\epsilon_{11, n+1}^{e, v} - \epsilon_{11, n}^{e, v} -\Delta \epsilon_{11} - \Delta \lambda^{v} \frac{\partial F}{\partial \sigma_{11, n+1}} \\
    \vdots\\
    \epsilon_{33, n+1}^{e, v} - \epsilon_{33, n}^{e, v} -\Delta \epsilon_{33} - \Delta \lambda^{v} \frac{\partial F}{\partial \sigma_{33, n+1}} \\
        X_{11, n+1}^{v} - X_{11, n}^{v} +  \Delta \lambda^{v} \left( \frac{\partial f}{\partial X_{11,n+1}^{v}} +  \frac{\partial \phi}{\partial X_{11,n+1}^{v}}  \right)\\
\vdots \\
        X_{33, n+1}^{v} - X_{33, n}^{v} +  \Delta \lambda^{v} \left( \frac{\partial f}{\partial X_{33,n+1}^{v}} + \frac{\partial \phi}{\partial X_{33,n+1}^{v}}  \right)\\
r_{n+1}^{v} - r_{n}^{v} + \Delta \lambda^{v} \frac{\partial f}{\partial R_{n+1}^{v}} \\
R_{n+1}^{v} - R_{n} + \Delta \lambda^{v}\frac{\partial^{2} \psi_{1}^{p}}{\partial r_{n+1}^{v, 2}} \frac{\partial f}{\partial R_{n+1}^{v}}  \\
f(\bm{\epsilon}^{e, v}_{n+1},  \bm{X}_{n+1}^{v}, R_{n+1}^{v} )
    \end{bmatrix}
\end{equation}
and
\begin{equation}
    \bm{J}= \begin{bmatrix}
    c_{1111} & \cdots & c_{1133}  & d_{1111} & \cdots & d_{1133} & h_{11}  & m_{11} &   \frac{\partial F}{\partial \sigma_{11, n+1}} \\
    \vdots & \cdots & \vdots & \vdots & \cdots & \vdots & \vdots & \vdots &\vdots \\
    c_{3311} & \cdots & c_{3333} & d_{3311} & \cdots & d_{3333} & h_{33}& m_{33} &  \frac{\partial F}{\partial \sigma_{33, n+1}} \\
    e_{1111} & \cdots & e_{1133} & g_{1111} & \cdots & g_{1133} & k_{11} &s_{11}&\left( \frac{\partial f}{\partial X_{11,n+1}^{v}} +  \frac{\partial \phi}{\partial X_{11,n+1}^{v}}  \right)      \\
    \vdots & \cdots & \vdots & \vdots & \cdots & \vdots & \vdots& \vdots&\vdots \\
    e_{3311} & \cdots & e_{3333} & g_{3311} & \cdots & g_{3333} &k_{33} & s_{33}&\left( \frac{\partial f}{\partial X_{33,n+1}^{v}} +  \frac{\partial \phi}{\partial X_{33,n+1}^{v}}  \right)     \\
\beta_{11} & \cdots & \beta_{33}  & \xi_{11} & \cdots & \xi_{33}  & \tau  & \Gamma &   \frac{\partial f}{\partial R_{n+1}^{v}}  \\
b_{11} & \cdots & b_{33}  & f_{11} & \cdots & f_{33} & \varphi  & \rho &    \frac{\partial^{2} \psi_{1}^{p}}{\partial r_{n+1}^{2}} \frac{\partial f}{\partial R_{n+1}}  \\
    \frac{\partial f}{\partial  \epsilon_{11, n+1}^{e, v}}  & \cdots  & \frac{\partial f}{\partial  \epsilon_{3, n+1}^{e, v}}  & \frac{\partial f}{\partial X_{11, n+1}^{v}} & \cdots & \frac{\partial f}{\partial X_{33, n+1}^{v}} & \frac{\partial f}{\partial r_{n+1}^{v}} & \frac{\partial f}{\partial R_{n+1}^{v}}&  \frac{\partial f}{\partial \Delta \lambda^{v}}
        \end{bmatrix}
\end{equation}

where
\begin{equation}
    \begin{aligned}
        c_{ijkl} &= \delta_{ik} \delta_{jl} + \Delta \lambda^{v}  \frac{\partial^{2} F}{\partial \sigma_{ij, n+1}  \partial \epsilon_{kl, n+1}^{e, v}} \\
         d_{ijkl} &= \Delta \lambda^{v}  \frac{\partial^{2} F}{\partial \sigma_{ij, n+1}  \partial X_{kl, n+1}^{e, v}} \\
         e_{ijkl} &= \Delta \lambda^{v} \left( \frac{\partial^{2} f}{\partial X_{ij,n+1}^{v} \partial \epsilon_{kl, n+1}^{e, v}} +  \frac{\partial^{2} \phi}{\partial X_{ij,n+1}^{v} \partial \epsilon_{kl, n+1}^{e, v}}  \right) 
         \\
                g_{ijkl} &= \delta_{ik} \delta_{jl} +\Delta \lambda^{v} \left( \frac{\partial^{2} f}{\partial X_{ij,n+1}^{v} \partial X_{kl, n+1}^{e, v}} +  \frac{\partial^{2} \phi}{\partial X_{ij,n+1}^{v} \partial X_{kl, n+1}^{e, v}}  \right)  \\
        h_{ij} &= \Delta \lambda^{v}  \frac{\partial^{2} F}{\partial \sigma_{ij, n+1}  \partial r_{n+1}^{v}} \\
        m_{ij} &= \Delta \lambda^{v}  \frac{\partial^{2} F}{\partial \sigma_{ij, n+1}  \partial R_{n+1}^{v}} \\
        k_{ij} &=\Delta \lambda^{v} \left( \frac{\partial^{2} f}{\partial X_{ij,n+1}^{v} \partial r_{n+1}^{v}}  \right) \\
        s_{ij} &=\Delta \lambda^{v} \left( \frac{\partial^{2} f}{\partial X_{ij,n+1}^{v} \partial R_{n+1}^{v}}  \right) \\
        b_{ij} &= \Delta \lambda^{v} \frac{\partial^{2} \psi_{1}^{p}}{\partial r_{n+1}^{v,2}}  \frac{\partial^{2} f}{\partial R_{n+1}^{v} \partial \epsilon_{ij, n+1}^{e, v}} \\
        f_{ij} &=\Delta \lambda^{v}\frac{\partial^{2} \psi_{1}^{p}}{\partial r_{n+1}^{v,2}} \frac{\partial^{2} f}{\partial R_{n+1}^{v} \partial X_{ij, n+1}^{e, v}} \\
        \beta_{ij} &=\Delta \lambda^{v} \frac{\partial^{2} f}{\partial R_{n+1}^{v} \partial \epsilon_{ij, n+1}^{e, v}} \\
        \xi_{ij} &=  \Delta \lambda^{v} \frac{\partial^{2} f}{\partial R_{n+1}^{v} \partial X_{ij, n+1}^{e, v}} \\
        \tau &= 1  + \Delta \lambda^{v} \frac{\partial^{2} f}{\partial R_{n+1}^{v}  \partial r_{n+1}^{v}} \\
        \Gamma &= \Delta \lambda^{v} \frac{\partial^{2} f}{\partial R_{n+1}^{v}  \partial R_{n+1}^{v}} \\
        \varphi &=  \Delta \lambda^{v} \left( \frac{\partial^{3} \psi_{1}^{p}}{\partial r_{n+1}^{v,3}} \frac{\partial f}{\partial R_{n+1}^{v}} + \frac{\partial^{2} \psi_{1}^{p}}{\partial r_{n+1}^{v,2}} \frac{\partial^{2} f}{\partial R_{n+1}^{v} \partial r_{n+1}^{v}}  \right) \\
        \rho &= 1 + \Delta \lambda^{v} \frac{\partial^{2} \psi_{1}^{p}}{\partial r_{n+1}^{v,2}} \frac{\partial^{2} f}{\partial R_{n+1}^{v} \partial R_{n+1}^{v}}
    \end{aligned}
\end{equation}

\clearpage
\bibliography{bib.bib}

\end{document}